\RequirePackage{fix-cm}
\documentclass[twocolumn]{svjour3}          
\smartqed  
\usepackage{graphicx}
\usepackage{enumerate}
\usepackage{epsfig}
\usepackage{graphicx}
\usepackage{multirow}
\usepackage{color}
\usepackage{xcolor}
\usepackage{caption}

\usepackage{colortbl}
\usepackage{array}
\usepackage{tikz}
\usepackage{times}
\usepackage{epsfig}
\usepackage{graphicx}
\usepackage{amsmath}
\usepackage{amssymb}
\usepackage{tabularx}
\newcommand{\abbname}[1]{P-Strip}
\usepackage{bbold}
\usepackage{comment}
\usepackage{adjustbox}
\usepackage[title]{appendix}%
\usepackage{xcolor}
\newcolumntype{Y}{>{\centering\arraybackslash}X}
\usepackage{color}
\definecolor{blueIX}{RGB}{13, 71, 161}

\makeatletter  
\newcommand{\thickhline}{%
	\noalign {\ifnum 0=`}\fi \hrule height 1pt
	\futurelet \reserved@a \@xhline
}

\usepackage{multirow}
\newcommand{\dsnameM}[1]{AnyPattern}

\usepackage{tcolorbox}
\definecolor{lightroyalblue}{HTML}{F6F8FD} 
\definecolor{royalblue}{HTML}{4169E1}
\definecolor{lighterblue}{HTML}{f2fafd}  

 \usepackage{natbib}
\usepackage{etoolbox}
\makeatletter
\patchcmd{\NAT@citex}
  {\@citea\NAT@hyper@{%
     \NAT@nmfmt{\NAT@nm}%
     \hyper@natlinkbreak{\NAT@aysep\NAT@spacechar}{\@citeb\@extra@b@citeb}%
     \NAT@date}}
  {\@citea\NAT@nmfmt{\NAT@nm}%
   \NAT@aysep\NAT@spacechar\NAT@hyper@{\NAT@date}}{}{}

\patchcmd{\NAT@citex}
  {\@citea\NAT@hyper@{%
     \NAT@nmfmt{\NAT@nm}%
     \hyper@natlinkbreak{\NAT@spacechar\NAT@@open\if*#1*\else#1\NAT@spacechar\fi}%
       {\@citeb\@extra@b@citeb}%
     \NAT@date}}
  {\@citea\NAT@nmfmt{\NAT@nm}%
   \NAT@spacechar\NAT@@open\if*#1*\else#1\NAT@spacechar\fi\NAT@hyper@{\NAT@date}}
  {}{}
\makeatother

\newcommand{\dsname}[1]{VariDet}
\newcommand{\msname}[1]{UFC}

\usepackage{bbding}
\definecolor{color1}{HTML}{C96B58}
\definecolor{color2}{HTML}{88D86B}
\definecolor{color3}{HTML}{6F5DCF}
\definecolor{color4}{HTML}{FFD966}
\definecolor{color5}{HTML}{CF5E9E}
\definecolor{color6}{HTML}{5582DB}
\definecolor{green}{HTML}{00B050}
\definecolor{mygray}{gray}{.9}
\definecolor{ggray}{RGB}{127,127,127}
\definecolor{reda}{RGB}{192,0,0}
\definecolor{redb}{RGB}{217,148,143}
\definecolor{myyellow}{RGB}{190,144,0}
\definecolor{mygreen}{RGB}{80,100,40}
\definecolor{myblue}{RGB}{30,90,100}
\definecolor{blue}{RGB}{4,20,110}
\definecolor{mygray1}{RGB}{245,245,245}

\newcommand{\dscolor}[1]{\textcolor{blueIX}{#1}}
\usepackage[colorlinks=true, citecolor=blue, linkcolor=reda, urlcolor=purple]{hyperref}
\usepackage{tcolorbox}
\definecolor{lightroyalblue}{HTML}{F6F8FD} 
\definecolor{royalblue}{HTML}{4169E1}
\definecolor{lighterblue}{HTML}{f2fafd}  
\newtcolorbox{abox}{colback=lightroyalblue,colframe=black,boxrule=0.2pt}
\usepackage{tikz}

\begin{document}

\title{\raisebox{-0.21cm}{\includegraphics[height=0.72cm]{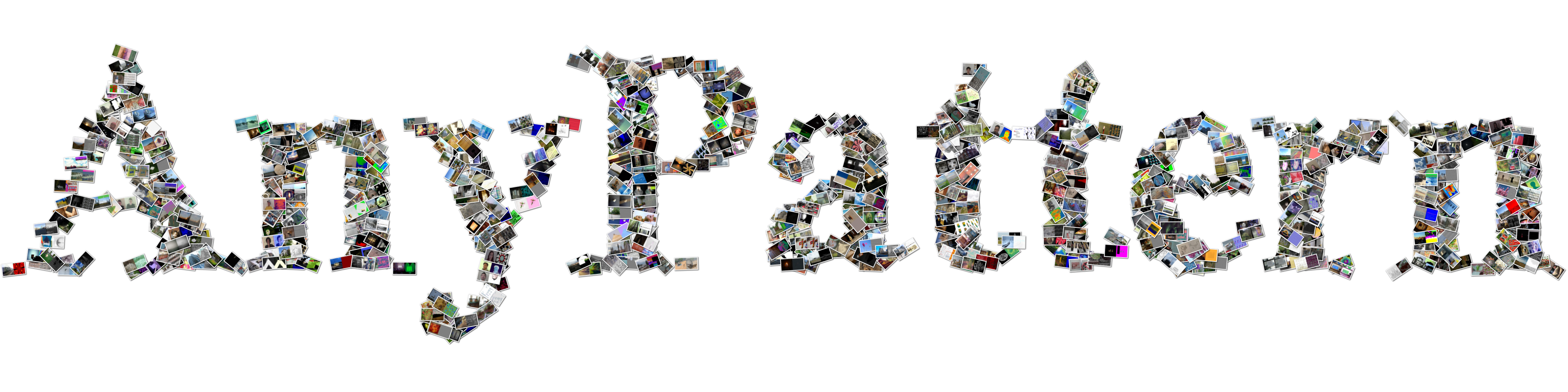}}: Towards In-context Image Copy Detection
}

\titlerunning{AnyPattern: Towards In-context Image Copy Detection}        

\author{Wenhao Wang  \and
        Yifan Sun    \and 
        Zhentao Tan \and 
        Yi Yang
}


\institute{
Wenhao Wang$^1$ \at wangwenhao0716@gmail.com
          \and
          Corresponding author: Yifan Sun$^{2}$ \at sunyf15@tsinghua.org.cn
          \and
          Zhentao Tan$^{2}$\at tanzhentao@stu.pku.edu.cn
          \and
          Yi Yang$^3$ \at
          yangyics@zju.edu.cn
\and
          $^1$University of Technology Sydney, Sydney, Austrilia.  \at
          \vspace{-3mm}
          \and
          $^2$Baidu Inc, Beijing, China. \at
          \vspace{-3mm}
          \and
          $^3$Zhejiang University, Zhejiang, China. }

\date{Received: date / Accepted: date}

\maketitle

\begin{figure*}[h]
    \centering
    \includegraphics[width=17cm]{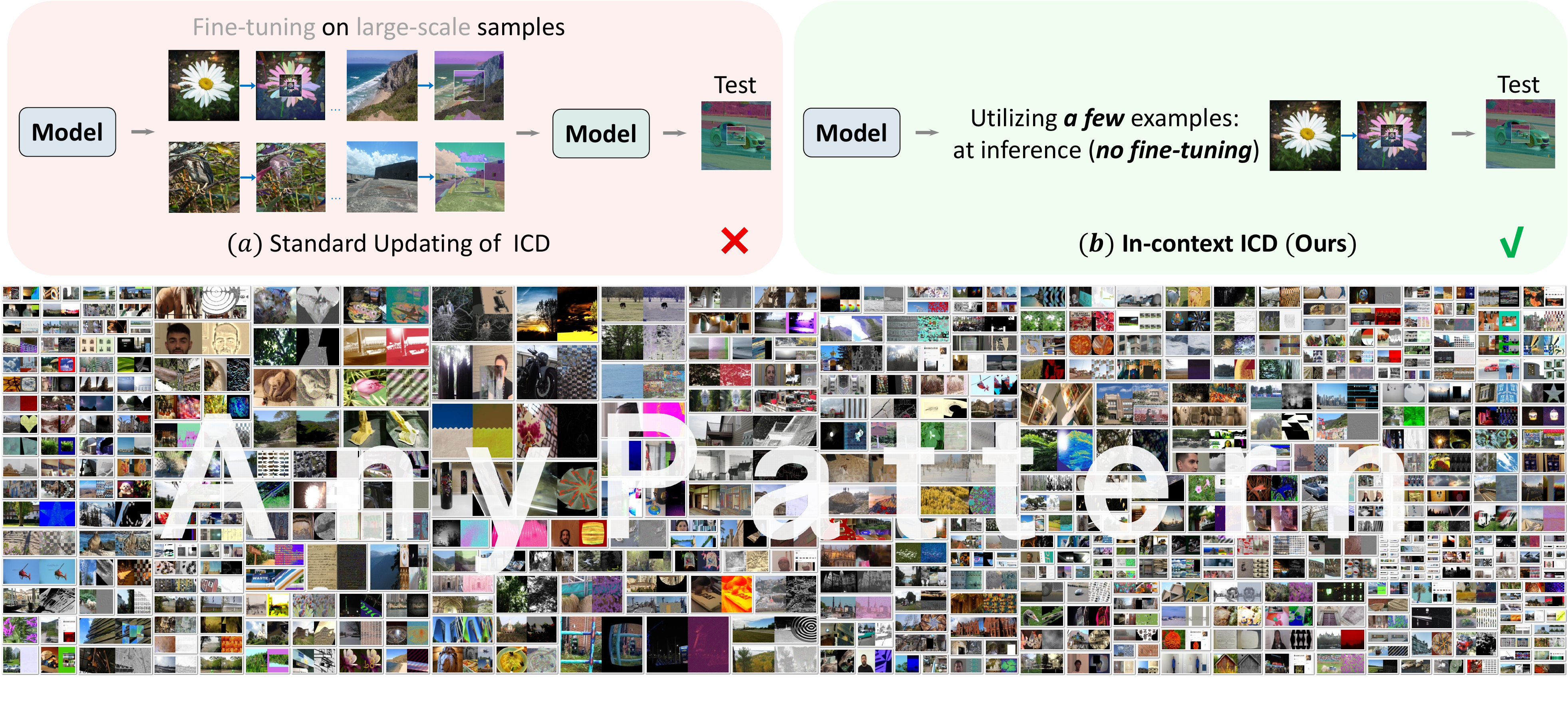}
    \vspace{-5mm} 
    \caption{\textbf{Top}: The comparison between the standard updating process of Image Copy Detection (ICD) and the proposed in-context ICD. Unlike the standard updating approach, our in-context ICD eliminates the need for fine-tuning, making it more efficient. \textbf{Bottom}: AnyPattern is the first large-scale pattern dataset, featuring 90 base and 10 novel patterns. Using 90 base patterns, we generate a training dataset containing 10 million images. Note that each pattern in this paper refers to a class of transformations that are diverse within themselves (see Appendix (Section \ref{App: demo})).}
    \label{Fig: teasor}
   \vspace{-4mm} 
\end{figure*}

\begin{abstract}

This paper explores in-context learning for image copy detection (ICD), \textit{i.e.}, prompting an ICD model to identify replicated images with new tampering patterns without the need for additional training.
The prompts (or the contexts) are from a small set of image-replica pairs that reflect the new patterns and are used at inference time. Such in-context ICD has good realistic value, because it requires no fine-tuning and thus facilitates fast reaction against the emergence of unseen patterns. To accommodate the ``seen $\rightarrow$ unseen'' generalization scenario, we construct the first large-scale pattern dataset named AnyPattern, which has the largest number of tamper patterns ($90$ for training and $10$ for testing) among all the existing ones. We benchmark AnyPattern with popular ICD methods and reveal that existing methods barely generalize to novel patterns. We further propose a simple in-context ICD method named ImageStacker. ImageStacker learns to select the most representative image-replica pairs and employs them as the pattern prompts in a stacking manner (rather than the popular concatenation manner). 
Experimental results show \textbf{(1)} training with our large-scale dataset substantially benefits pattern generalization ($+26.66  \%$ $\mu AP$), \textbf{(2)} the proposed ImageStacker facilitates effective in-context ICD (another round of $+16.75 \%$ $\mu AP$), and \textbf{(3)} AnyPattern enables in-context ICD, \textit{i.e.}, without such a large-scale dataset, in-context learning does not emerge even with our ImageStacker. 
Beyond the ICD task, we also demonstrate how AnyPattern can benefit artists, \textit{i.e.}, the pattern retrieval method trained on AnyPattern can be generalized to identify style mimicry by text-to-image models. 
The project is publicly available at \url{https://anypattern.github.io}.

\end{abstract}

\vspace{-4mm}
\section{Introduction}

Image Copy Detection (ICD) aims to identify whether a query image is replicated from a database after being tampered with. It serves critical roles in areas such as copyright enforcement, plagiarism prevention, digital forensics, and ensuring content uniqueness on the internet. 

Under the realistic scenario, the ICD models suffer from the inevitable emergence of novel tamper patterns. More concretely, the ICD models trained on some already-known patterns may fail when encountering novel patterns. 
Updating the ICD models for the novel patterns is very expensive and time-consuming. It usually requires collecting a large amount of training samples and then fine-tuning the ICD models, as illustrated in Fig.~\ref{Fig: teasor} (a). 

As a more efficient solution, this paper explores in-context learning for ICD, as illustrated in Fig.~\ref{Fig: teasor} (b). In-context learning is a relatively new machine learning paradigm that learns to solve unseen tasks by providing examples in the prompt. 
Combining this paradigm with ICD, we endow the ICD models with the ability to recognize novel patterns without fine-tuning. The resulting in-context ICD uses a few examples of image-replica pairs (that prompt the novel patterns) as the context of its input. Though the model parameters remain unchanged, the extracted features are modulated (conditioned) by the context and become competent for recognizing the novel-patterned replication. 
Consequently, in-context ICD facilitates a fast and efficient reaction against the emergence of unseen patterns.

To set up the ``seen $\rightarrow$ unseen'' pattern generalization scenario, we construct the first large-scale pattern dataset named \textit{AnyPattern}. As shown in Fig.~\ref{Fig: teasor} (bottom), AnyPattern is featured for its abundant ($100$) tamper patterns, with $90$ for training and $10$ for testing. 
Concretely, the training set consists of replicas generated from the combination of multiple training patterns (randomly chosen from the $90$ patterns), as well as the original images. 
We devote approximately one million CPU core hours to generate 10 million training images in total. 
The testing set consists of queries (replicas) generated from the combination of $10$ novel patterns and galleries (their original images and distractors). 
Another important characteristic of our AnyPattern dataset is: it provides examples for indicating novel patterns at inference time. 
Each example is an image-replica pair, within which the replica is generated from the combination of some novel patterns. 
These examples are very limited (\emph{e.g.}, 10 examples for each pattern combination) and are not to be used for fine-tuning. During inference, the in-context ICD uses these examples to gain knowledge of the novel patterns instantly. 
A thorough illustration of all the patterns is provided in the Appendix (Section \ref{App: demo}). \par 

Given the prerequisite dataset, we further propose a simple and straightforward in-context ICD method named ImageStacker. ImageStacker basically follows the standard in-context learning pipeline, \emph{i.e.}, using some examples as the context of the input. The context conditions the feature extraction of the ICD models and is also known as prompt(s). We name our in-context learning method as ImageStacker, because it has a unique prompting manner, \emph{i.e.}, stacking the examples and the input images together. 
During training, we use the ground truth to prepare the image-replica pairs that have the same patterns as the pseudo query images, yielding the in-context learning. During testing, though we are provided with a set of examples that cover the novel patterns, we still do not know which patterns are exactly the ones for generating the query (the replica). In response, we design a pattern retrieval method to select the image-replica examples that are most likely to share the same patterns with the query images. In other words, we retrieve the most representative image-replica pairs in the example set as the prompts for ImageStacker.  \par 

To further demonstrate the significance of introducing the AnyPattern dataset, we present an additional application using AnyPattern with the proposed pattern retrieval method. The text-to-image diffusion model can be used to mimic the style of artwork with little cost, and this threatens the livelihoods and creative rights of artists. To help them protect their work, we treat an artist’s `style’ as a `pattern’ and generalize the trained pattern retrieval method to identify generated images with style mimicry.

To sum up, this paper makes the following contributions:
\begin{enumerate}
\item We introduce in-context ICD, which allows the use of a few examples to prompt an already-trained ICD model to recognize novel-pattern replication, without the need of fine-tuning. To support the scenario of pattern generalization, we construct the first large-scale pattern dataset, \textit{i.e.} AnyPattern, which provides 90 training patterns and 10 base patterns.



\item We benchmark AnyPattern against (1) popular ICD methods and find that none of these methods could generalize to novel patterns well, and (2) various intuitive prompting approaches, discovering that most of these common visual prompting methods are ineffective for in-context ICD. Therefore, we propose a simple in-context ICD method, \textit{i.e.} ImageStacker. ImageStacker stacks an image-replica example (prompt) to the query image along the channel dimension, and yields a good in-context learning effect for recognizing novel tamper patterns.


\item To further highlight the generalization and importance of AnyPattern dataset, we present its another application, \textit{i.e.} the pattern retrieval method trained on AnyPattern can be generalized to identify generated images by text-to-image diffusion models that most closely matches the style of a given real artwork.

%

\end{enumerate}

\vspace{-4mm}
\section{Related Works} \label{sec:related}

\textbf{In-context Learning for Computer Vision.} In-context learning originates from large language models like GPT-3 \textcolor{blue}{\citep{brown2020language}} and Instruction GPT \textcolor{blue}{\citep{ouyang2022training}}. In the pure computer vision area, in-context learning is a relatively new concept and presents challenges for implementation. The earliest known work, MAE-VQGAN \textcolor{blue}{\citep{bar2022visual}}, implements in-context learning through image inpainting. Following MAE-VQGAN \textcolor{blue}{\citep{bar2022visual}}, (Un)SupPR \textcolor{blue}{\citep{zhang2023makes}} discusses how to find good visual in-context examples; Prompt-SelF \textcolor{blue}{\citep{sun2023exploring}} explores the factors that affect the performance of visual in-context learning; Painter \textcolor{blue}{\citep{Painter}} designs a generalist vision model to automatically generate images according to the example pairs; Prompt Diffusion \textcolor{blue}{\citep{wang2023context}} proposes diffusion-based generative models. Specific computer vision areas also see work such as SegGPT \textcolor{blue}{\citep{wang2023seggpt}} for image segmentation. These methods typically implement in-context learning in a concatenating manner,  while we propose a stacking design. 

\textbf{Existing Image Copy Detection Methods.} Existing ICD methods \textcolor{blue}{\citep{pizzi2022self,yokoo2021contrastive,papadakis2021producing,wang2021bag,wang2023benchmark,fernandez2022active}} can be broadly categorized into contrastive learning-based algorithms and deep metric learning-based algorithms. The use of contrastive learning for training an ICD model is natural due to its reliance on data augmentation. For example, SSCD \textcolor{blue}{\citep{pizzi2022self}} is based on the InfoNCE \textcolor{blue}{\citep{oord2018representation}} loss and introduces a differential entropy regularization to differentiate nearby vectors. CNNCL \textcolor{blue}{\citep{yokoo2021contrastive}} employs a large memory bank with a contrastive loss to learn from numerous positive and negative pairs. The application of deep metric learning to ICD is intuitive given that ICD is, fundamentally, a retrieval task. EfNet \textcolor{blue}{\citep{papadakis2021producing}} proposes a ``drip training'' procedure, whereby the number of classes used to train a model is incrementally increased. BoT \textcolor{blue}{\citep{wang2021bag}} sets a robust baseline for ICD and introduces descriptor stretching, which normalizes scores at the feature level. However, none of these methods can be directly applied to our in-context ICD scenario as they never consider generalizing their models to novel patterns, let alone using only image-replica pairs.

\vspace{-2mm}
\section{In-context Image Copy Detection} \label{sec: benchmark}
This section first gives a formal definition of in-context Image Copy Detection (ICD) and then introduces the constructed AnyPattern dataset.

\vspace{-2mm}
\subsection{Definition}
\begin{figure*}[t]
    \centering
    \includegraphics[width=16cm]{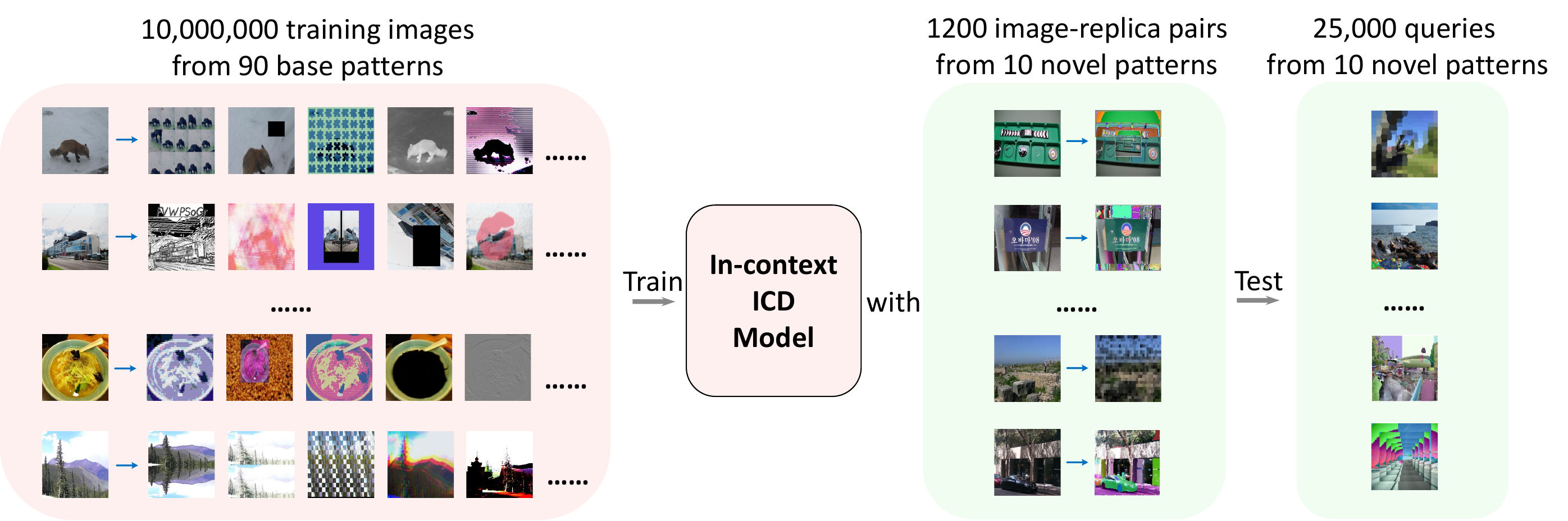}
    \vspace{-1mm} 
    \caption{The illustration for our in-context Image Copy Detection (ICD) with AnyPattern. In-context ICD necessitates a well-trained ICD model to be prompted to novel patterns with the assistance of a few image-replica pairs and without any fine-tuning process. In realistic scenarios, this setup is highly practical as it provides a feasible solution for a deployed ICD system faced with unseen patterns.} 
    \label{Fig: UniICD}
    \vspace{-4mm} 
\end{figure*}

\vspace{-2mm}
The in-context ICD requires an already-trained ICD model to recognize novel-pattern replication by using a few image-replica examples as the prompt. As illustrated in Fig. \ref{Fig: UniICD}, the base tamper patterns underlying the training data have no overlap with the novel patterns, while the prompts are generated from the random combination of novel patterns.

Formally, the objective of in-context ICD is to train a model $g$ with parameters $\tau$ using only the training (base) pattern set. 
To detect novel-pattern replication, we do not fine-tune the model to update $\tau$, but use some prompts to condition/modify the feature extraction. Given a query image $x_q$, its feature is extracted by:  
\begin{equation}
g_\tau\left(\mathcal{F}, x_q\right),
\end{equation}
where $\mathcal{F} \subset \mathcal{D}$ are the prompts chosen from the image-replica pool $\mathcal{D}=\left\{\left(A_i, A^{\prime }_{i}\right)\right\}_{i=1}^N$, and $\left(A_i, A^{\prime }_{i}\right)$ is the $i$th image-replica pair. 

In-context ICD requires the extracted feature $g_\tau\left(\mathcal{F}, x_q\right)$ to be discriminative for identifying whether $x_q$ is replicated from any gallery image. 

\vspace{-2mm}
\subsection{AnyPattern Dataset}
\vspace{-2mm}
AnyPattern has two characteristics, \emph{i.e.}, 1) having plenty of tamper patterns and 2) providing a small set of image-replica pairs as the prompts of novel patterns. 

\textbf{1) Large size of tamper patterns.} AnyPattern set encompasses a total of $100$ patterns: $90$ are designated for training in-context ICD models, while the remaining $10$ are reserved for testing. A comprehensive introduction to these patterns can be found in the Appendix (Section \ref{App: demo}). 

\textbf{2) A small pool of image-replica pairs.}
For each combination of novel patterns, we provide $10$ image-replica pairs as the prompts. Totally, there are $1,200$ prompts. We note that during inference, only partial prompts (\emph{e.g.}, $1 \sim 10$, depending on the hyper-parameters) are used for each query. \par

The source images to generate these replicated images and prompts are from the DISC21 dataset \textcolor{blue}{\citep{douze20212021,papakipos2022results}}. DISC21 has $1$ million unlabeled training images, from which we randomly select $100,000$ images as the un-edited images following \textcolor{blue}{\citep{wang2021bag, wang2023benchmark}}. Each training image is transformed $99$ times by randomly selected training patterns. Together with the original training images, we construct a training dataset containing $10$ million images. Owing to the complexity and volume of the patterns, this process is distributed across \textbf{200 CPU nodes} in a supercomputing cluster and requires about \textbf{one million} CPU core hours. We adopt the gallery dataset from DISC21 as our gallery. The query set includes $25,000$ queries, among which $5,000$ are generated by applying a randomly selected novel pattern combination to gallery images, and the remaining $20,000$ queries serve as distractors (without true matches in the gallery). 
\vspace{-2mm}
\subsection{Comparison against Existing Datasets}
\vspace{-2mm}
Currently, there are three publicly available ICD benchmarks, i.e. CopyDays \textcolor{blue}{\citep{douze2009evaluation}}, DISC21 \textcolor{blue}{\citep{douze20212021}}, and NDEC \textcolor{blue}{\citep{wang2023benchmark}}. 




\textbf{CopyDays} \textcolor{blue}{\citep{douze2009evaluation}} was launched in 2009. This dataset contains only 157 query images and 3,000 gallery images, and lacks training data. The types of tampering patterns involved are relatively straightforward, such as alterations in contrast and blurring.

\textbf{DISC21} \textcolor{blue}{\citep{douze20212021,papakipos2022results}} was established in 2021 as an extensive benchmark for ICD, notable for its massive scale, including one million training images and one million gallery images, along with complicated tampering patterns. Additionally, it includes numerous distractor queries, which do not correspond to any true matches in the gallery.

\textbf{NDEC} \textcolor{blue}{\citep{wang2023benchmark}} addresses the challenge of hard negatives in ICD, \textit{i.e.}, some images may appear very similar yet are not replications. By incorporating this aspect of hard negatives, NDEC enhances the realism of ICD evaluations.

Beyond these existing datasets, our AnyPattern provides several novel explorations:

\textbf{(1) AnyPattern has the largest number of tampering patterns.} Specifically, our AnyPattern features 90 base patterns and 10 novel patterns. For CopyDays, there are only a few very simple patterns, e.g., contrast changes and blurring. DISC21 features about 20 patterns, including complex ones. NDEC focuses on hard negative problems and inherits the patterns from DISC21.

\textbf{(2) AnyPattern is the first dataset that carefully regulates the base and novel patterns.} All of the previous datasets only define the patterns for generating queries, and none of them restrict the patterns for training. Therefore, researchers directly use the patterns for generating queries as the training patterns, which brings over-optimistic results. In contrast, the test and training patterns in our AnyPattern are well-defined and separated.

\textbf{(3) AnyPattern is the only dataset that enables in-context ICD.} As shown in Table \ref{Table: discuss}, in-context learning does not emerge when training with DISC21 and a small number of patterns. In contrast, with our AnyPattern, training ImageStacker significantly improves performance. This reaffirms the value of our proposed AnyPattern. 


\begin{figure*}[t]
   \centering
    \includegraphics[width=17cm]{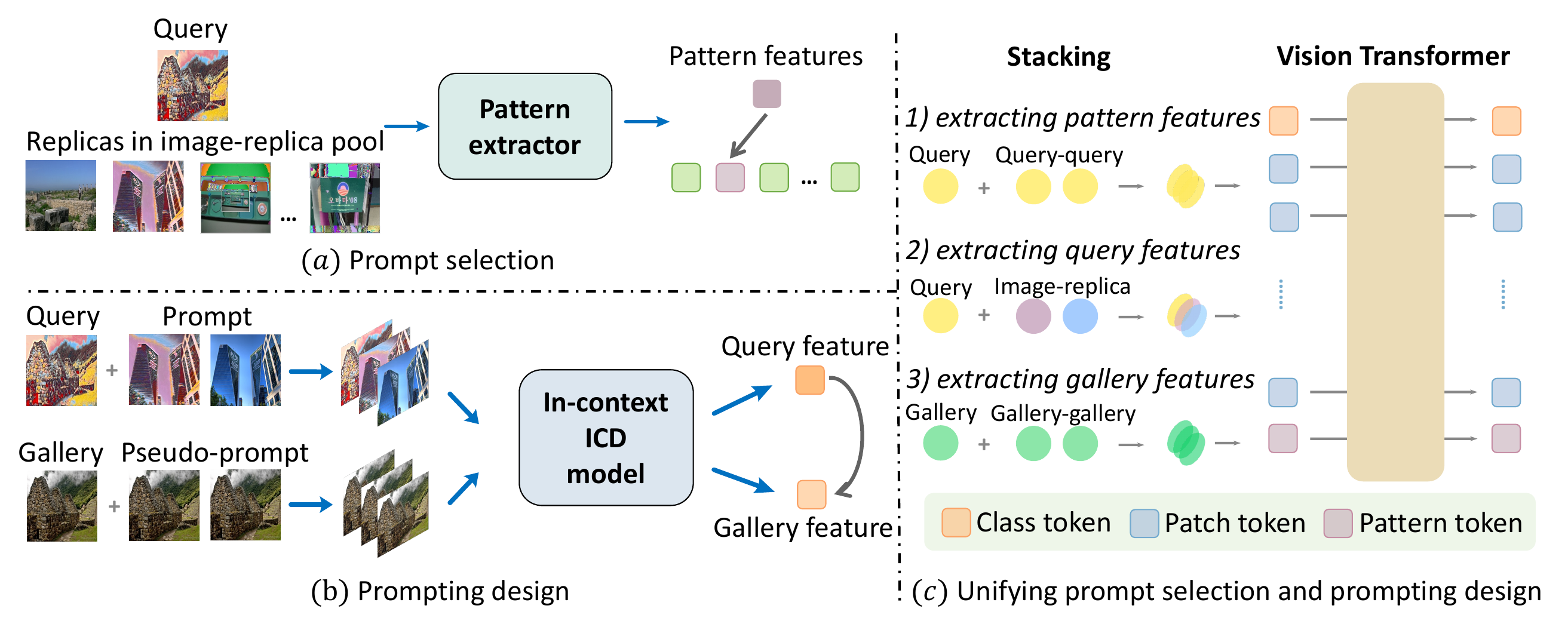}
    \vspace{-2mm} 
    \caption{The proposed \textbf{ImageStacker} includes: (a) \textbf{prompt selection} fetches the most representative image-replica pair from the whole pool for a given query, and (b) \textbf{prompting design} stacks the selected image-replica pair onto a query along the channel dimension, and thus the image-replica pair conditions the feed-forward process. In (c), we show how to \textbf{unify prompt selection and prompting design} into one vision transformer.}
    \label{Fig: main}
    \vspace{-4mm} 
\end{figure*}

\vspace{-4mm}
\section{Method} \label{sec: method}


In this section, we provide a detailed illustration of our proposed ImageStacker (Fig. \ref{Fig: main}). The deep metric learning baseline used by our ImageStacker is first briefly reviewed in Section \ref{sec: baseline}. During testing for queries with novel patterns, though we have a set of examples containing the novel patterns, we still do not know exactly which patterns generate the query (the replica). Hence, in Section \ref{sec: pr}, we propose a prompt selection method, pattern retrieval, to fetch the most representative image-replica pair from the entire image-replica pool. Subsequently, in Section \ref{sec: stack}, using the selected prompt, we introduce a unique prompt design, \textit{i.e}, stacking. Finally, we try to unify the prompt selection method and the prompting design into one ViT backbone in Section \ref{sec: unify}.

\vspace{-2mm}
\subsection{Baseline}\label{sec: baseline}
\vspace{-2mm}
This section briefly overviews the ICD baseline implemented in our ImageStacker. Following \textcolor{blue}{\citep{wang2021bag,papadakis2021producing}}, we conceptualize ICD as an image retrieval task, primarily adopting deep metric learning methods. Specifically, we treat each original image and all its replicas as a \textit{training class} and perform \textit{deep metric learning} on these classes. Pairwise training \textcolor{blue}{\citep{sohn2016improved,hermans2017defense}}, classification training \textcolor{blue}{\citep{liu2016large,wang2018cosface,sun2020circle}}, or their combination can be utilized for this purpose. In our baseline, we select classification training, specifically CosFace \textcolor{blue}{\citep{wang2018cosface}}, due to its demonstrated effectiveness and simplicity.
\vspace{-2mm}
\subsection{Pattern Retrieval}\label{sec: pr}
\vspace{-2mm}

Drawing inspiration from image retrieval techniques, we propose a pattern retrieval method for identifying the image-replica pairs corresponding to a given query (Fig. \ref{Fig: main} (a)). The training of pattern retrieval can be seen as a multi-label classification task. When using Vision Transformer (ViT) \textcolor{blue}{\citep{dosovitskiy2020image}} as the pattern extractor, we design a pattern token $x^{0}_{ptr}$ and concatenate it to the ViT input:
\begin{equation}
{}\left[ x^{L}_{cls},\mathbf{X}^{L} ,x^{L}_{ptr}\right]  =f\left( \left[ x^{0}_{cls},\mathbf{X}^{0} ,x^{0}_{ptr}\right]  \right),
\end{equation}
where $f$ represents the ViT, $\mathbf{X}^{0}$ is the patch tokens, $x^{0}_{cls}$ is the class token, and $L$ is the number of layers in a ViT. 


To use $x^{L}_{ptr}$ as the representation of patterns in an image, it is supervised by the binary cross-entropy loss, which is formulated as

\begin{equation}
\mathcal{L}_{ptr} = -\frac{1}{M} \sum^{M}_{i=1} \sum^{C}_{c=1} \left[ y_{ic}\log (p_{ic})+(1-y_{ic})\log (1-p_{ic})\right]  ,
\end{equation}
where $M$ is the number of training images, $C$ is the number of training patterns ($C=90$ here), $y_{ic}$ is the label of the $i$-th image for the $c$-th pattern class, and $p_{ic}$  is the predicted probability by the model that the $i$-th image belongs to the $c$-th pattern class. The pattern token interacts with patch tokens during the feed-forward process and can be considered as the feature of a pattern combination. During testing, the classification head is discarded, and retrieval is performed with the feature.
\par

\begin{figure*}[h]
    \centering
    \includegraphics[width=16cm]{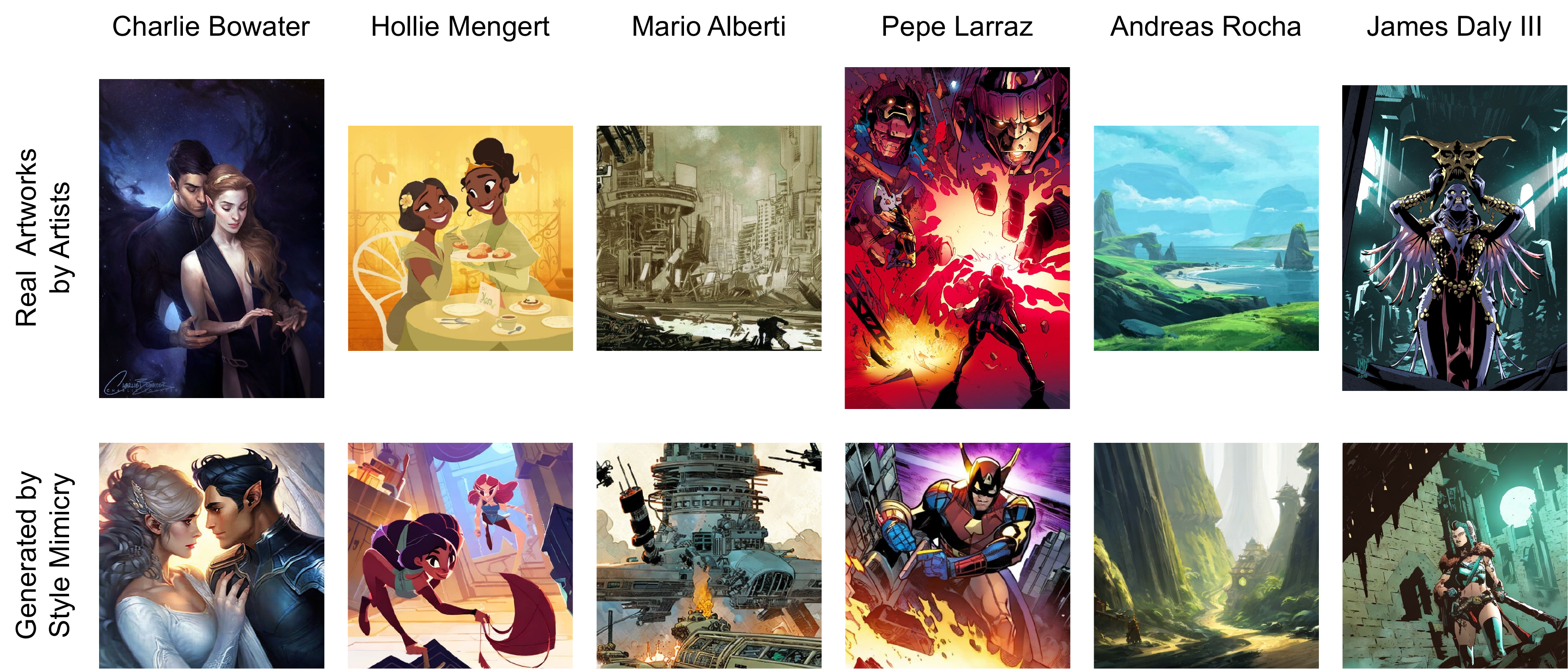}
    \vspace{-1mm} 
    \caption{The demonstration for the style mimicry by a tuned DreamBooth model.} 
    \label{Fig: mimicry}
    \vspace{-4mm} 
\end{figure*}

\vspace{-2mm}
\subsection{Stacking}\label{sec: stack}
\vspace{-2mm}
To address the in-context ICD, given a prompt, we introduce a simple yet effective prompting manner, \textit{i.e.} stacking (Fig. \ref{Fig: main} (b)). This unique prompting manner modifies the input structure of a ViT: traditionally, an image is divided into $N$ patches ($\left\{x_i \in \mathbb{R}^{\textbf{3} \times P \times P} \mid i=1,2, \ldots, N\right\}$, where $ P \times P$ is the patch size) before being passed to the embedding layer. In contrast, ImageStacker stacks an image-replica pair along the channel dimension, tripling the original channel count.  As a result, the $N$ patches are represented in a new format:
\begin{equation}
\left\{\bar{x_i}  \in \mathbb{R}^{\textbf{9} \times P \times P} \mid i=1,2, \ldots, N\right\}.
\end{equation}
To accommodate these $9$-channel image patches, we initialize a new embedding layer (while keeping the hidden size and all other details of the original ViT). 
For a query/replica, we use the retrieved image-replica pair or ground truth as the prompt; for a gallery/original image, we directly duplicate itself two times as the pseudo-prompt.
The stacked image-replica pair alters the feed-forward process, thus allowing for a conditioned input image feature.
The advantage of our stacking design lies in introducing an inductive bias to the in-context learning, which emphasizes the contrasts at the same or similar positions between the original image and the copy. Since many tampering patterns occur at the same or similar positions, this inductive bias brings more benefits to the in-context learning process compared to the traditional concatenation method without any inductive bias.

\vspace{-2mm}
\subsection{Unifying Pattern Retrieval and Stacking}\label{sec: unify}
\vspace{-2mm}
We try to unify the pattern retrieval and stacking process into one ViT backbone (Fig. \ref{Fig: main} (c)). The unification is straightforward for training because the image-replica pair is selected based on ground truths. 
The final loss is defined as

\begin{equation}
\mathcal{L}_{final} = \mathcal{L}_{cls} + \lambda \cdot \mathcal{L}_{ptr},
\end{equation}
where $\mathcal{L}_{cls}$ is the CosFace loss \textcolor{blue}{\citep{wang2018cosface}}, and $\lambda$ is the balance parameter.  \par 

\begin{figure*}[h]
    \centering
    \includegraphics[width=16cm]{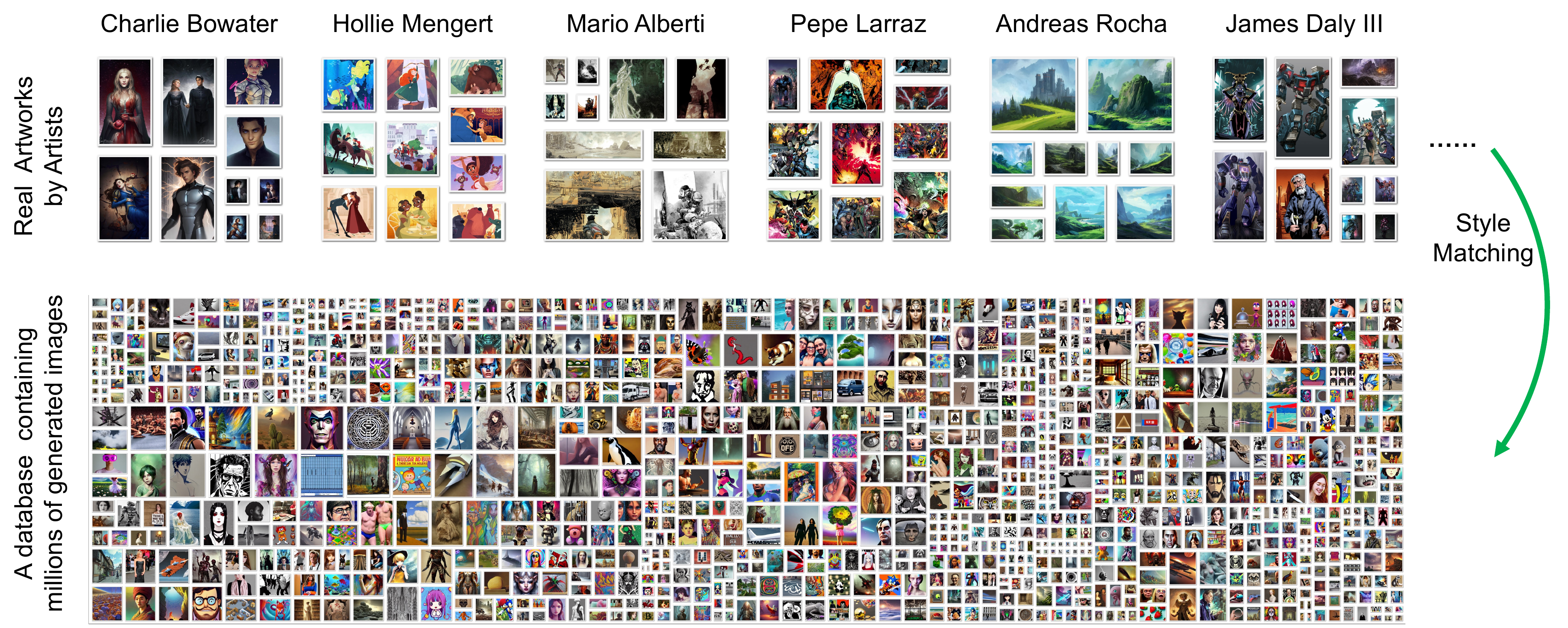}
    \vspace{-2mm} 
    \caption{The demonstrations for matching the style of given artworks against millions of generated images.} 
    \vspace{-4mm}
    \label{Fig: target}
\end{figure*}

During testing, a key challenge arises: \textit{ImageStacker requires an image-replica pair as input; yet before the feed-forward process, we do not have the pattern feature to retrieve an image-replica pair.} To overcome this, we introduce the pseudo-image-replica pairs to get the pattern feature for a query, \textit{i.e.} we duplicate one query itself two times as the pseudo-prompt (Fig. \ref{Fig: main} (c-1)). Consequently, we acquire the pattern feature and thus use it to fetch the most representative image-replica pair for each query. 
Stacking the fetched image-replica pair onto a query, we extract its (image) feature (Fig. \ref{Fig: main} (c-2)). Because the gallery does not contain patterns, we duplicate itself two times as its pseudo-prompt and then extract its (image) feature (Fig. \ref{Fig: main} (c-3)).

\vspace{-4mm}
\section{AnyPattern helps artists}\label{App: anypatternstyle}

\vspace{-2mm}
\subsection{Background}
\vspace{-2mm}
\textbf{DreamBooth} \textcolor{blue}{\citep{ruiz2023dreambooth}} presents a novel method for \textit{personalizing} text-to-image diffusion models. By using a few reference images of a subject, DreamBooth enables the model to generate new, high-quality images of that subject in various contexts specified by textual descriptions. This fine-tuning approach retains the model's general capabilities while enhancing its ability to produce detailed and contextually appropriate depictions of the specific subject. Due to its potential negative societal impact, with concerns that ‘malicious parties might try to use such images to mislead viewers’, the inventors at Google decided not to release any code or trained models. However, a third party re-implemented it and made it publicly available \textcolor{blue}{\citep{Xiao2022DreamboothStableDiffusion}}.

\textbf{Style mimicry.} After the release of DreamBooth, people discover that it can easily be used to mimic the styles of any artist. Specifically, anyone can collect as few as five artworks created by an artist and spend less than one dollar to train a DreamBooth model, which can then generate numerous images in the same style. For instance, Ogbogu Kalu released a tuned DreamBooth model on Hugging Face \textcolor{blue}{\citep{ogkalu2024comicdiffusion}}, which was tuned using the artworks of six comic artists and has gained significant popularity. In Fig. \ref{Fig: mimicry}, we contrast the real artworks created by these artists with the images generated by the tuned DreamBooth model, demonstrating that their styles are indeed very similar.

\textbf{Opinion of artists and others.} The phenomenon of style mimicry has sparked a debate about the ethics of fine-tuning AI on the artworks of living artists in the comments on Reddit \textcolor{blue}{\citep{reddit2022scarceillustration}} and other places. Supporters of AIGC consider the generated images `incredibly beautiful' and describe the released model as their `favorite custom model by far'. Meanwhile, some legal professionals argue that it is lawful because `style is not copyrightable'. However, there is more to it than that. \textbf{F}irstly, artists feel frustrating, uncomfortable and invasive for this \textcolor{blue}{\citep{Baio2022InvasiveDiffusion}}. For instance, Hollie Mengert wonders if the model’s creator simply did not think of her as a person. She also has concerns about copyright because some of the training artworks from her are created for copyright holders, such as Disney and Penguin Random House. \textbf{S}econdly, these models may end artists ability to earn a living \textcolor{blue}{\citep{shan2023glaze}}. Artists invest significant time and effort in cultivating their unique styles, which is a critical aspect of their livelihood. As a model replicates these styles without offering compensation, artists' opportunities to market their work and connect with potential buyers are significant hindered. \textbf{F}inally, the artistic creation will be a swan song \textcolor{blue}{\citep{Nguyen2023AIStudentArtists}}. This imitation by AI can be demoralizing for art students who are training to become the next generation of artists. Seeing AI models potentially replace their future roles can be discouraging and impact their career aspirations.

\vspace{-3mm}
\subsection{Our target}
\vspace{-2mm}

Glaze \textcolor{blue}{\citep{shan2023glaze}} offers `style cloaks' to artists to help mislead the mimicry of their styles by text-to-image models. However, the authors of Glaze acknowledge a limitation: this preventative approach can only protect newer artworks. More specifically, many works have already been downloaded from art repositories such as ArtStation and DeviantArt, and these artists' styles can still be mimicked using older artworks collected before Glaze was released.

Therefore, it is important for artists to be aware of any generated images that mimic the styles of their released artworks. By knowing this, they can utilize opt-out and removal options, \textit{i.e.}, requesting that providers of these text-to-image models or the owners of generated images with mimicked styles cease the style mimicry.

Here, we aim to provide artists with such a `\textbf{style retrieval}' tool. 
It is a direct application of our AnyPattern and pattern retrieval method, showing their generalizability to other datasets or real-world scenarios where patterns significantly differ. Specifically, we treat an artist's `style' as a `pattern'. Therefore, using our pattern retrieval method trained with AnyPattern, we can directly search a database containing millions of generated images to identify the image that most closely matches the style of a given real artwork. A demonstration of such a process is shown in Fig. \ref{Fig: target}.

\vspace{-2mm}
\subsection{Implementation}
\vspace{-2mm}
\begin{figure*}[h]
    \centering
    \includegraphics[width=17cm]{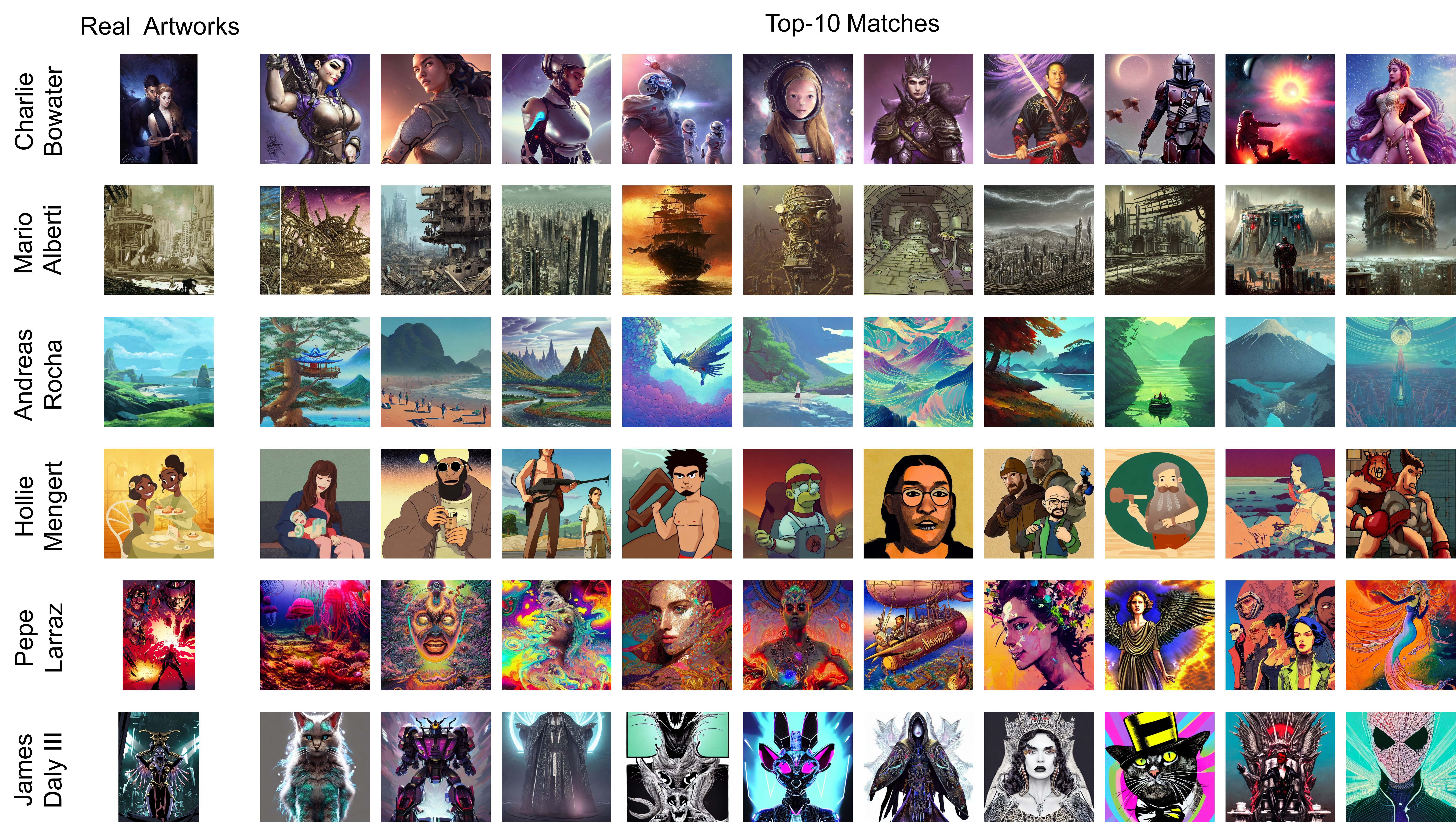}
    \vspace{-2mm} 
    \caption{Original artworks (left column) with their corresponding top-10 style matches generated by a text-to-image model (right column), showcasing our trained model's proficiency in capturing color, texture, and thematic elements.} 
    \vspace{-4mm}
    \label{Fig: style_match}
\end{figure*}

\textbf{Experimental setup.} To test the generalizability of our trained pattern retrieval method in identifying style mimicry by text-to-image models, we first construct a database of millions of generated images. Specifically, we utilize prompts from DiffusionDB \textcolor{blue}{\citep{wangDiffusionDBLargescalePrompt2023}} and Stable Diffusion V1.5 \textcolor{blue}{\citep{runwayml2022stablediffusionv15}} to generate $1,819,776$ images. Then, we collect several publicly available artworks by artists such as Charlie Bowater, Hollie Mengert, Mario Alberti, Pepe Larraz, Andreas Rocha, and James Daly III as query images. It is important to note that our use of these artworks falls under the category of fair use for research purposes \textcolor{blue}{\citep{WikipediaFairUse}}, thereby avoiding any copyright infringement. Employing the style descriptor\footnote{\url{https://github.com/WangWenhao0716/AnyPatternStyle}} we trained on AnyPattern, we extract a 768-dimensional vector from each generated image and real artwork. By computing the cosine similarities, we identify the top-10 generated images that most closely match the style of each real artwork.

\textbf{Observations.} We visualize some matching results in Fig. \ref{Fig: style_match} and conclude that our trained model successfully identifies the style mimicry by text-to-image models: \textbf{F}irstly, the model appears to accurately capture the unique color palettes and lighting of the original artworks. For instance, the vibrant purples and blues in Charlie Bowater's piece are reflected in the generated images. Similarly, the dusky, sepia tones of Mario Alberti's cityscapes are well-represented. \textbf{S}econdly, the stylistic elements, like brush strokes and texturing, seem to be well understood by the model. Andreas Rocha's landscapes with their distinct, somewhat stylized textures are matched with similar generated images.
\textbf{F}inally, the thematic elements are reflected in the generated matches. For Hollie Mengert's character-focused art, the matched generated images also focus on character-centric scenes. Pepe Larraz's dynamic compositions with a flair for the dramatic are mirrored in the matched images which capture similar energy and movement. James Daly III's work that features a blend of sci-fi and fantasy elements is matched with images that maintain this blend.

\vspace{-2mm}
\subsection{Limitations and future directions}
\vspace{-2mm}
Although our trained pattern retrieval method successfully generalizes to identify style mimicry, we acknowledge that a gap still exists between the manually-designed patterns in AnyPattern and the art styles created by artists. Therefore, to better assist artists in identifying style mimicry, future work may involve incorporating a broader range of art styles into the training set and developing corresponding quantitative evaluations.

\begin{figure*}
    \centering
    \includegraphics[width=17cm]{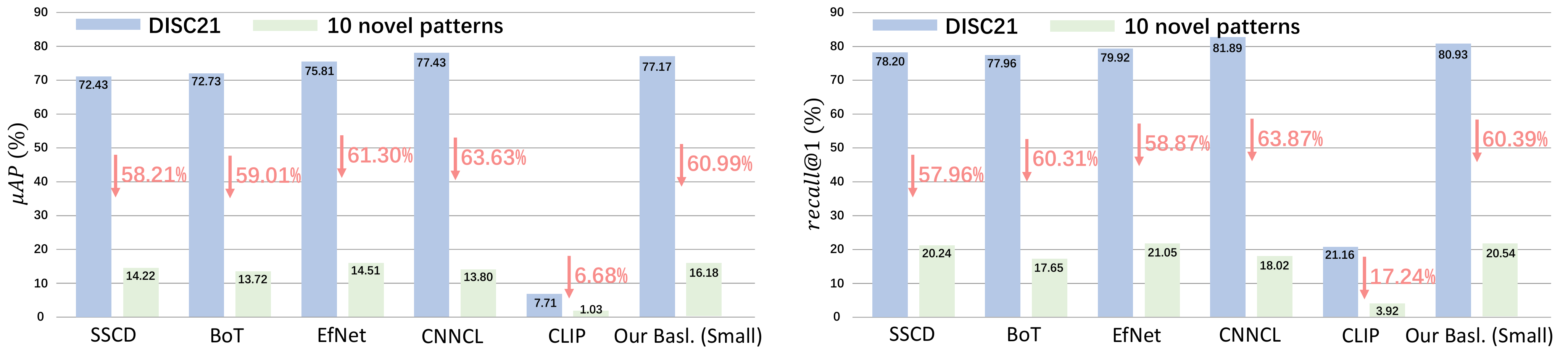}
    \vspace{-2mm} 
    \caption{The performance of ICD state-of-the-arts, CLIP \textcolor{blue}{\citep{radford2021learning}}, and our Basl. (Small) on the DISC21 dataset and novel patterns.  Our Basl. (Small) is trained with the same patterns with BoT \textcolor{blue}{\citep{wang2021bag}}. Since these algorithms (including our Basl. (Small)) are not designed to handle input of image-replica pairs, we demonstrate their performance decrease by directly testing their trained models on the $10$ novel patterns from AnyPattern.} 
   \vspace{-4mm} 
    \label{Fig: challenge}
\end{figure*}

\vspace{-4mm}
\section{Experiments} \label{sec: exp}
\vspace{-2mm}
\subsection{Evaluation Metrics and Training Details} 
\vspace{-2mm}
\textbf{Evaluation metrics.}
We employ two evaluation metrics for the in-context ICD, namely $\mu AP$ and $recall@1$/$R@1$. $\mu AP$ serves as an overall evaluation metric and is equivalent to the area under the Precision-Recall curve, providing a comprehensive measure of both the precision and recall of our model across varying thresholds. On the other hand, the $recall@1$ metric is query-specific. It checks whether the actual (correct) result appears first in the list of all returned results, offering insight into the effectiveness of our model in accurately retrieving the most relevant result at the top.
\begin{table}
        \caption{The performance improvement on the \textbf{novel} patterns from AnyPattern and in-context learning.}
\vspace*{-2mm}
\small
\begin{minipage}{0.48\textwidth}
  \begin{tabularx}{\hsize}{|>{\centering\arraybackslash}p{1.5cm}||Y|Y|Y|Y|Y|}
    \hline\thickhline
    
   \rowcolor{mygray}\scalebox{0.9}{$\mu AP$ ($\%$)} &  \scalebox{0.9}{SSCD}& \scalebox{0.9}{BoT} & \scalebox{0.9}{EfNet} & \scalebox{0.9}{CNNCL} & \scalebox{0.9}{Baseline} \\ \hline \hline

  \scalebox{0.9}{SmallPattern}& $14.22$& $13.72$& $14.51$ & $13.80$& $16.18$ \\ 
 \scalebox{0.9}{AnyPattern}&$39.79$&$39.81$&$41.33$& $40.65$&$42.84$\\
 \rowcolor{lightroyalblue}\scalebox{0.9}{ImageStacker}&$53.94$&$54.11$&$55.17$&$53.51$& $\mathbf{56.65}$\\ \hline
    \end{tabularx}
    \end{minipage}
    
    \vspace*{1mm}
    
    \begin{minipage}{0.48\textwidth}
   \begin{tabularx}{\hsize}{|>{\centering\arraybackslash}p{1.5cm}||Y|Y|Y|Y|Y|} 
    \hline\thickhline
 \rowcolor{mygray}\scalebox{0.9}{$R@1$ ($\%$)}&\scalebox{0.9}{SSCD}&\scalebox{0.9}{BoT}&\scalebox{0.9}{EfNet} &\scalebox{0.9}{CNNCL} & \scalebox{0.9}{Baseline}\\ \hline\hline
\scalebox{0.9}{SmallPattern}&$20.24$&$17.65$&$21.05$&$18.02$&$20.54$\\
\scalebox{0.9}{AnyPattern} &$42.83$&$43.97$& $47.19$& $45.68$ &$47.86$\\
\rowcolor{lightroyalblue}\scalebox{0.9}{ImageStacker}&$57.31$&$60.14$&$59.96$&$57.10$& $\mathbf{60.86}$\\ \hline
  \end{tabularx}
  \end{minipage}
  \label{Table: AnyPattern}
  \vspace*{-4mm}

\end{table}

\textbf{Training details.}
We implement our ImageStacker using PyTorch \textcolor{blue}{\citep{paszke2019pytorch}} and distribute its training across eight Nvidia A100 GPUs. We use ViT-B/16 \textcolor{blue}{\citep{dosovitskiy2020image}} as the backbone, which is pre-trained on the ImageNet dataset \textcolor{blue}{\citep{deng2009imagenet}} using DeiT \textcolor{blue}{\citep{touvron2021training}} unless otherwise specified. Before training, we resize images to a resolution of $224\times224$ pixels. We set the balance parameter, $\lambda$, at $1$ and use a batch size of $512$. Each batch adopts the standard PK sampling method, with $128$ classes and $4$ images per class. The total number of training epochs is $25$ with a cosine-decreasing learning rate. The margin $m$ and scale $s$ in CosFace loss \cite{wang2018cosface} are set to $0.35$ and $64$, respectively.
\vspace{-4mm}
\subsection{The \textbf{Challenge} from Novel Patterns}
\vspace{-2mm}

In this section, we present the performance degradation of trained ICD models (SSCD \textcolor{blue}{\citep{pizzi2022self}}, BoT \textcolor{blue}{\citep{wang2021bag}}, EfNet \textcolor{blue}{\citep{papadakis2021producing}}, and CNNCL \textcolor{blue}{\citep{yokoo2021contrastive}}) and the CLIP \textcolor{blue}{\citep{radford2021learning}} model, when encountering novel patterns. All the ICD state-of-the-arts are trained on the DISC21 dataset, and we select the most successful CLIP \textcolor{blue}{\citep{radford2021learning}} model, which is trained on the $2$ billion sample English subset of LAION-5B \textcolor{blue}{\citep{schuhmann2022laion}} and achieves a zero-shot top-1 accuracy of $80.1\%$ on ImageNet-1k \textcolor{blue}{\citep{deng2009imagenet}}. The corresponding $\mu AP$ and $recall@1$ scores are summarized in Fig. \ref{Fig: challenge}, leading us to two main observations. \textbf{F}irst, all the ICD models, despite being trained with different methods and pattern combinations, experience a significant accuracy decrease (about $60\%$ for $\mu AP$ and $recall@1$) when the evaluation dataset changes from DISC21 to novel patterns. Notably, our baseline is trained on the same pattern combination with BoT \textcolor{blue}{\citep{wang2021bag}} (Basl. (Small)) and maintains comparable performance with state-of-the-arts on both the DISC21 dataset and the $10$ novel patterns. \textbf{S}econd, while CLIP models display impressive results in zero-shot image classification and image retrieval tasks, they are not ideally suited for ICD tasks (less than $10\%$ $\mu AP$). This can be attributed to CLIP being predominantly trained on natural images. In light of these findings, we argue that in practical scenarios, the continuous emergence of novel patterns poses a significant challenge for deployed ICD models.

\begin{table}
        \caption{The performance on the \textbf{base} patterns from AnyPattern and in-context learning. Our in-context learning method (ImageStacker) not only improves performance on novel patterns significantly but also maintains (marginally improves) the performance on base patterns.}
\vspace*{-2mm}
\small
\begin{minipage}{0.48\textwidth}
  \begin{tabularx}{\hsize}{|>{\centering\arraybackslash}p{1.5cm}||Y|Y|Y|Y|Y|}
    \hline\thickhline
    
   \rowcolor{mygray}\scalebox{0.9}{$\mu AP$ ($\%$)} &  \scalebox{0.9}{SSCD}& \scalebox{0.9}{BoT} & \scalebox{0.9}{EfNet} & \scalebox{0.9}{CNNCL} & \scalebox{0.9}{Baseline} \\ \hline \hline

 \scalebox{0.9}{AnyPattern}&$77.54$&$76.12$&$77.56$& $79.84$&$79.37$\\
 \rowcolor{lightroyalblue}\scalebox{0.9}{ImageStacker}&$81.39$&$79.98$&$80.11$&$84.13$& $83.56$\\ \hline
    \end{tabularx}
    \end{minipage}
    
    \vspace*{1mm}
    
    \begin{minipage}{0.48\textwidth}
   \begin{tabularx}{\hsize}{|>{\centering\arraybackslash}p{1.5cm}||Y|Y|Y|Y|Y|} 
    \hline\thickhline
 \rowcolor{mygray}\scalebox{0.9}{$R@1$ ($\%$)}&\scalebox{0.9}{SSCD}&\scalebox{0.9}{BoT}&\scalebox{0.9}{EfNet} &\scalebox{0.9}{CNNCL} & \scalebox{0.9}{Baseline}\\ \hline\hline
\scalebox{0.9}{AnyPattern} &$81.76$&$80.37$& $81.13$&$83.41$&$83.05$\\
\rowcolor{lightroyalblue}\scalebox{0.9}{ImageStacker}& $84.52$& $83.96$ &$83.78$&$86.07$&$85.85$\\ \hline
  \end{tabularx}
  \end{minipage}
  \label{Table: base}
  \vspace*{-4mm}

\end{table}


\begin{figure*}
    \centering
    \includegraphics[width=16cm]{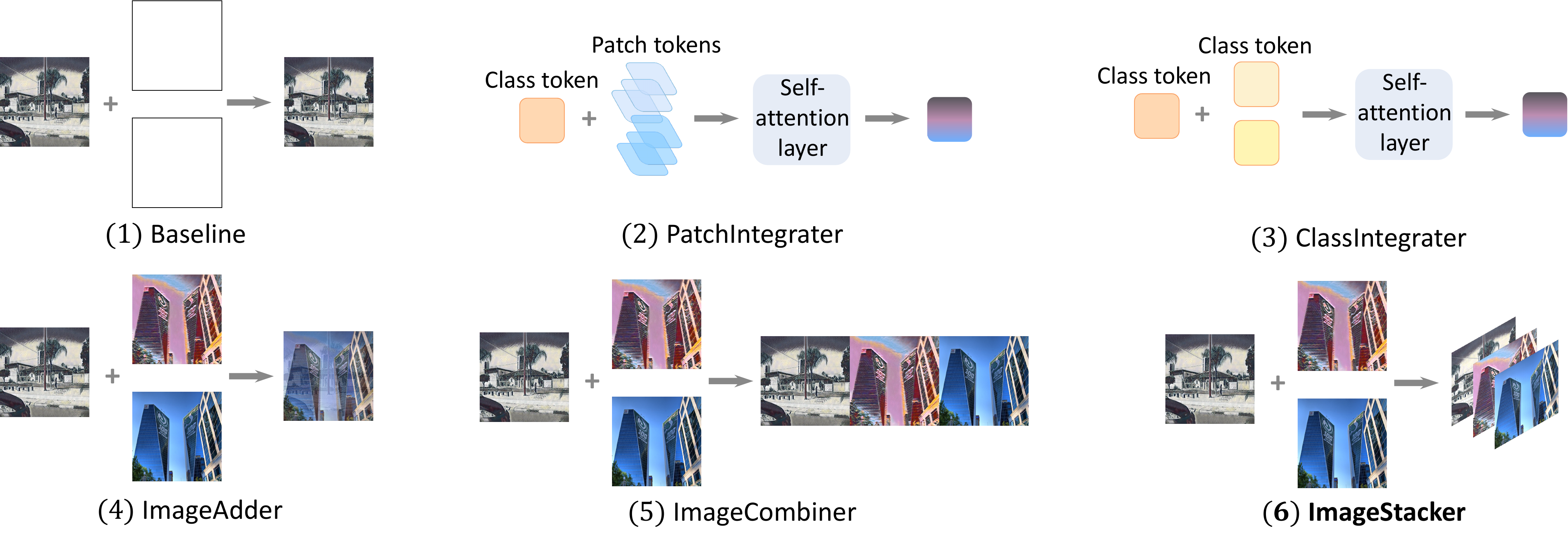}
   \vspace{-2mm} 
    \caption{The illustration for different visual prompting methods for incorporating image-replica pairs. Designs (2) and (3) operate on the feature level, while designs (4) through (6) function at the image level.} 
    \label{Fig: demo}
    \vspace{-4mm}
\end{figure*}

\vspace{-4mm}
\subsection{The \textbf{Benefits} from AnyPattern and In-context Learning}
\vspace{-2mm}
To improve performance on novel patterns, we show that both directly training models on larger pattern sets and conducting in-context learning are beneficial. As demonstrated in Table \ref{Table: AnyPattern}, we initially train models on the base patterns of our AnyPattern. This expansion significantly enhances performance on novel patterns: for example, resulting in a gain of $26.66\%$ in $\mu AP$ and $27.32\%$ in $recall@1$ for our baseline. However, a noticeable performance gap persists compared to scenarios where training and testing occur on the same pattern sets. This emphasizes the necessity of introducing in-context learning methods to further enhance performance. Our in-context learning method (ImageStacker) further improves performance on novel patterns significantly, achieving gains of $+16.75\%$ in $\mu AP$ and $+15.30\%$ in $recall@1$ for our baseline. Furthermore, it is also crucial to maintain performance on base patterns while enhancing performance on novel patterns through in-context learning because the edited copies generated by the base patterns may still appear in the future. As illustrated in Table \ref{Table: base}, our proposed ImageStacker achieves this objective. For instance, with our baseline, the $\mu AP$ has been increased from $79.37\%$ to $83.56\%$, and the $recall@1$ has improved from $83.05\%$ to $85.85\%$.

\begin{table}
        \caption{The performance of the baseline and our ImageStacker with different (pre-)training data. Without large-scale AnyPattern, \textbf{in-context learning} does not emerge.}
\vspace*{-2mm}
\small
\begin{minipage}{0.48\textwidth}
  \begin{tabularx}{\hsize}{|>{\centering\arraybackslash}p{1cm}||Y|Y|Y|>{\centering\arraybackslash}p{0.7cm}|>{\centering\arraybackslash}p{0.7cm}|}
    \hline\thickhline
    
   \rowcolor{mygray}\scalebox{0.9}{Method} &  \scalebox{0.9}{\hspace{-1mm}AnyPattern}& \scalebox{0.9}{\hspace{-1.5mm}SmallPattern} & \scalebox{0.9}{ImageNet} & \scalebox{0.9}{$\mu AP$} & \scalebox{0.9}{$R@1$} \\ \hline \hline

 \cellcolor{white} & $\checkmark$ &{} & $\checkmark$ & $42.84$ & $47.86$ \\
\cellcolor{white}Baseline & \cellcolor{gray!15}{}& \cellcolor{gray!15}$\checkmark$ & \cellcolor{gray!15}$\checkmark$ & \cellcolor{gray!15}$16.18$ & \cellcolor{gray!15}$20.54$ \\
\cellcolor{white} & $\checkmark$ &{} &{} & $40.52$ & $46.16$ \\ \hline\hline

& \cellcolor{lightroyalblue} $\checkmark$ &\cellcolor{lightroyalblue} {} & \cellcolor{lightroyalblue} $\checkmark$ & \cellcolor{lightroyalblue} $\hspace{-0.5mm}\mathbf{56.65}$ & \cellcolor{lightroyalblue} $\hspace{-0.5mm}\mathbf{60.86}$ \\
\cellcolor{white}\multirow{-2 }{*}{Image-} &\cellcolor{gray!15} {}& \cellcolor{gray!15}$\checkmark$ & \cellcolor{gray!15}$\checkmark$ & \cellcolor{gray!15}$15.28$ & \cellcolor{gray!15}$20.10$ \\
\multirow{-2 }{*}{Stacker} & $\checkmark$ & {}&{} & $53.53$ & $58.51$ \\ \hline\hline

\cellcolor{white} & $\checkmark$ & {}& $\checkmark$ & $99.25$ & $99.84$ \\
\cellcolor{white}\multirow{-2 }{*}{Upper} &\cellcolor{gray!15}{} & \cellcolor{gray!15}$\checkmark$ & \cellcolor{gray!15}$\checkmark$ & \cellcolor{gray!15}$18.11$ & \cellcolor{gray!15}$22.83$ \\
\cellcolor{white}\multirow{-2 }{*}{Bound} & $\checkmark$ &{} & {}& $99.26$ & $99.86$ \\ \hline
    \end{tabularx}
    \end{minipage}
    
  \label{Table: discuss}
  \vspace*{-4mm}

\end{table}

\vspace{-4mm}
\subsection{AnyPattern \textbf{Enables} In-context ICD}
\vspace{-2mm}
In this section, \textbf{from a data perspective}, we explore factors beyond our proposed ImageStacker that contribute to the emergence of in-context learning. We first adjust the training patterns of ImageStacker from AnyPattern to SmallPattern (the one used in BoT \textcolor{blue}{\citep{wang2021bag}}), and subsequently discard the ImageNet-pre-trained models. The performance of the baseline and ImageStacker is presented in Table \ref{Table: discuss}. The Upper Bound is achieved by using the query and its original image as the image-replica (example) during the inference.  Our analysis reveals that:
\textbf{data plays a crucial role alongside the model:} (1) In-context learning does not emerge when using SmallPattern and ImageNet-pretrained models: comparing against baseline with $16.18\%$ in $\mu AP$, ImageStacker only achieves $15.28\%$ in $\mu AP$. (2) The use of AnyPattern leads to the emergence of in-context learning: even without ImageNet, training ImageStacker on AnyPattern already significantly improves performance ($+13.01\%$ in $\mu AP$ and $+12.35\%$ in $recall@1$). \textbf{This reaffirms the value of our proposed AnyPattern dataset.}

\vspace{-4mm}
\subsection{In-context Learning \textbf{Surpasses} Fine-tuning}
\vspace{-2mm}
Beyond the \textbf{efficiency} advantage, in-context learning also offers two \textbf{performance} advantages, as show in Table \ref{Table: tune}: \textbf{(1)} Fine-tuning fails when the amount of training data is limited. For instance, when using the $1,200$ image-replica pairs ($2,400$ images) in AnyPattern for fine-tuning instead of in-context learning, the $\mu AP$ on novel patterns is $41.15\%$, which is $-15.50\%$ compared to our in-context solution; and \textbf{(2)} Fine-tuning on large-scale data generated by novel patterns can lead to catastrophic forgetting of the base patterns. For instance, when using $200,000$ images generated by novel patterns for fine-tuning, compared to the in-context solution, although there is a performance ($+10.87\%$ $\mu AP$) superiority on novel patterns, the performance on base patterns drops significantly ($-63.5\%$ $\mu AP$).

\begin{table}
        \caption{The performance comparison between the in-context solution against the \textbf{fine-tuning} one. $2,400$, $20,000$, and $200,000$ represent the number of images used for in-context learning and fine-tuning, respectively.}
\vspace*{-2mm}
\small
\begin{minipage}{0.48\textwidth}
  \begin{tabularx}{\hsize}{|>{\centering\arraybackslash}p{2cm}||>{\centering\arraybackslash}p{1.1cm}|Y|Y|Y|}
    \hline\thickhline
    
   \rowcolor{mygray}&\multicolumn{1}{c|}{\scalebox{0.9}{In-context}}&\multicolumn{3}{c|}{\scalebox{0.9}{Fine-tuning}}\\
   \rowcolor{mygray}\multirow{-2 }{*}{\scalebox{0.9}{$\mu AP$}}&  \scalebox{0.9}{$2,400$} & \scalebox{0.9}{$2,400$} &\scalebox{0.9}{$20,000 $}& \scalebox{0.9}{$200,000$}   \\ \hline\hline
Novel patterns&\cellcolor{lightroyalblue}$56.65$&$41.15$&$60.01$&$67.52$\\ 
Base patterns&\cellcolor{lightroyalblue}$83.56$&$73.44$&$73.25$&$20.06$\\ \hline 
 
    \end{tabularx}
    
\vspace*{1mm}    
    
    \end{minipage}
    
    \begin{minipage}{0.48\textwidth}
  \begin{tabularx}{\hsize}{|>{\centering\arraybackslash}p{2cm}||>{\centering\arraybackslash}p{1.1cm}|Y|Y|Y|}
    \hline\thickhline
    
   \rowcolor{mygray}&\multicolumn{1}{c|}{\scalebox{0.9}{In-context}}&\multicolumn{3}{c|}{\scalebox{0.9}{Fine-tuning}}\\
   \rowcolor{mygray}\multirow{-2 }{*}{\scalebox{0.9}{$R@1$}}&  \scalebox{0.9}{$2,400$} & \scalebox{0.9}{$2,400$} &\scalebox{0.9}{$20,000 $}& \scalebox{0.9}{$200,000$}   \\\hline\hline
Novel patterns&\cellcolor{lightroyalblue}$60.86$&$ 46.70$&$ 65.73$&$ 77.44$\\ 
Base patterns&\cellcolor{lightroyalblue}$85.85$&$ 78.04$&$ 77.77$&$ 27.99$\\ \hline 
 
    \end{tabularx}
    \end{minipage}
    
  \label{Table: tune}
  \vspace*{-4mm}

\end{table}

\vspace{-4mm}
\subsection{\textbf{Ablation} Studies}
\vspace{-2mm}
\begin{table}
        \caption{The comparison between different visual prompting methods for incorporating image-replica pairs. Our ImageStacker not only achieves the highest \textbf{performance} but also maintains \textbf{efficient} training and inference. `Infer' and `Train' are in `$10^{-3} s/img$' and `$s/iter$', respectively.}
\vspace*{-2mm}
\small
\begin{minipage}{0.48\textwidth}
  \begin{tabularx}{\hsize}{|>{\centering\arraybackslash}p{2cm}||Y|Y|Y|Y|}
    \hline\thickhline
    
   \rowcolor{mygray}\scalebox{0.9}{Method} &  \scalebox{0.9}{$\mu AP$} & \scalebox{0.9}{$R@1$} &\scalebox{0.9}{Infer}& \scalebox{0.9}{Train}   \\ \hline \hline
Basline (Any)&$42.84$&$47.86$&$\textbf{1.28}$&$\textbf{0.184}$\\  \hline\hline
PatchIntegrater&$37.42$&$42.58$&$4.13$&$0.474$\\ 
ClassIntegrater&$38.91$&$43.75$&$2.54$&$0.462$\\ \hline \hline
ImageAdder&$25.07$&$33.89$&$1.29$&$0.194$\\
ImageCombiner&$48.47$&$53.34$&$5.39$&$0.538$\\ 
\rowcolor{lightroyalblue}\textbf{ImageStacker}&$\textbf{56.65}$&$\textbf{60.86}$&$1.31$&$0.199$\\ \hline
 
    \end{tabularx}
    \end{minipage}
    
  \label{Table: demo}
  \vspace*{-4mm}

\end{table}

\begin{figure*}[h]
    \centering
    \includegraphics[width=17cm]{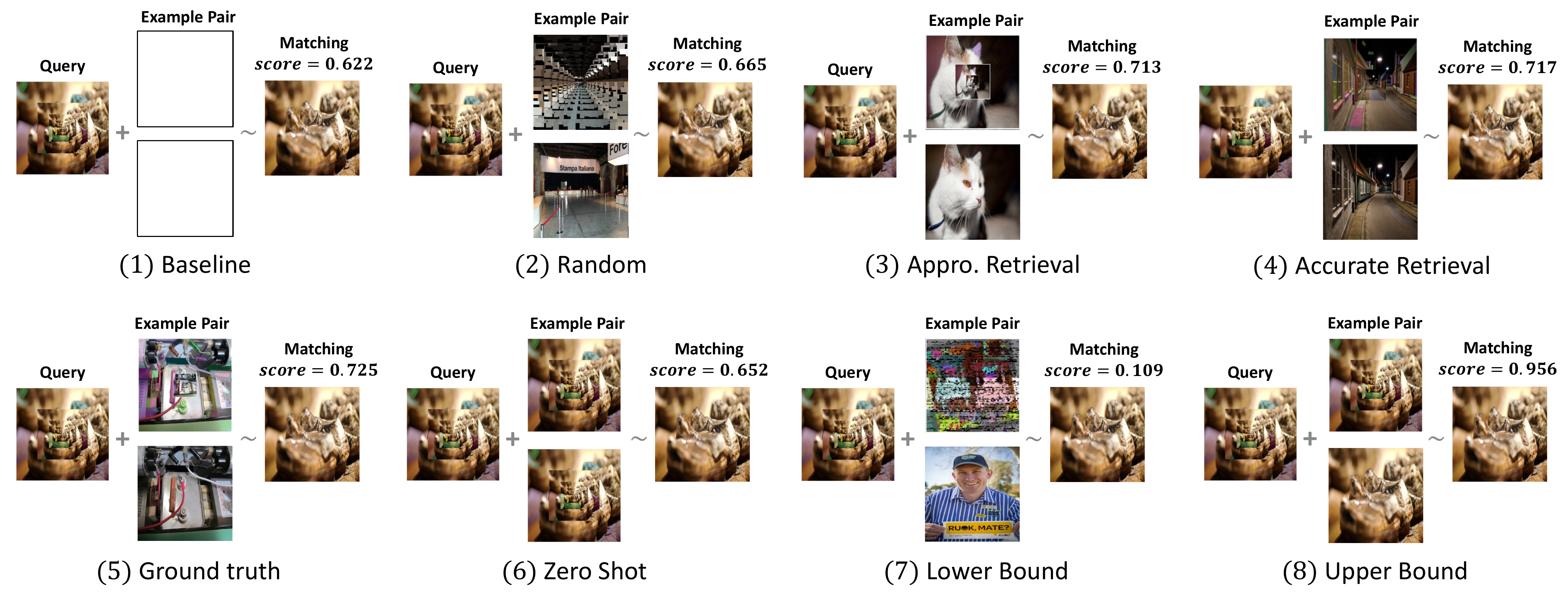}
    \vspace{-2mm} 
    \caption{The demonstration of different methods for selecting an image-replica pair. A larger score means a better matching. } 
    \label{Fig: re}
    \vspace*{-4mm}
\end{figure*}

\textbf{ImageStacker outperforms common visual prompting methods.} Models can integrate image-replica pairs at both feature and image levels. At the feature level, we append an extra self-attention layer to the last block of ViT to enable interaction between the class token of the replica and the patch tokens or class token of the image-replica pairs (see Fig. \ref{Fig: demo} (2) and (3)). At the image level, besides our proposed ImageStacker, we can directly add the image-replica pairs to the replica (See Fig. \ref{Fig: demo} (4)) or concatenate the images along the height (width) dimension (see Fig. \ref{Fig: demo} (5)). 
We compare these different designs in Table \ref{Table: demo}, drawing three main observations: \textbf{(1)} Incorporating image-replica pairs at the feature level is ineffective, resulting in about a $4\sim 6 \%$ performance drop in $\mu AP$. This is attributable to insufficient interaction between the replica and its image-replica pair. \textbf{(2)} Directly adding image-replica pairs to the replica significantly degrades the performance. This is because imposing a strong priori restriction without specific meaning can be detrimental. \textbf{(3)} Both ImageCombiner and ImageStacker significantly improve performance. Compared to ImageCombiner, ImageStacker achieves a greater performance gain ($+8.18\%$ in $\mu AP$ and $+7.52\%$ in $recall@1$), while requiring only about $1/4$ inference workload and $1/3$ training workload. Also, ImageStacker enhances the baseline by $13.81\%$ in $\mu AP$ and $13.00\%$ in $recall@1$, with nearly the same efficiency.

\begin{table}
        \caption{Different methods to \textbf{retrieve} the image-replica pair of a query.}
\vspace*{-2mm}
\small
\begin{minipage}{0.48\textwidth}
  \begin{tabularx}{\hsize}{|>{\centering\arraybackslash}p{2cm}||Y|Y|Y|}
    \hline\thickhline
    
  \rowcolor{mygray}Method&$\mu AP$ & $R@1$&Pattern Acc.\\ \hline\hline
  Basline (Any) &$42.84$&$47.86$&-\\ \hline\hline
  Random &$50.42$&$54.83$&$30.11$\\ 
  \rowcolor{lightroyalblue}Approximate &$56.65$&$60.86$&$62.90$\\
  Accurate &$56.68$&$60.78$&$64.60$\\
  Ground truth &$56.99$&$60.96$&$100.00$\\ 
 \hline\hline
  Zero Shot &$49.71$&$53.76$&-\\\hline\hline
  Lower Bound &$28.55$&$32.25$&-\\ 
 Upper Bound&$99.25$&$99.84$&-\\ \hline
 
    \end{tabularx}
    \end{minipage}
    
  \label{Table: re}
  \vspace*{-4mm}

\end{table}

\textbf{The proposed pattern retrieval method is effective.} Table \ref{Table: re} outlines several methods for obtaining the image-replica pair for a given query: (1) Random - selecting an image-replica pair randomly from the entire image-replica pool. (2) Approximate retrieval - using the ImageStacker model. (3) Accurate retrieval - employing another model. (4) Ground truth - using pattern ground truth of queries (not available in practice). ImageStacker can also be applied in a zero-shot setting by duplicating the query two times as the pseudo-prompt (Zero Shot). The lower bound is obtained by using an incorrect image-replica pair, while the upper bound is achieved by using the query and its original image as the image-replica. 
Their pattern retrieval accuracy is shown in the next section.
The demonstration of these methods is visualized in the Fig. \ref{Fig: re}. We observe that: \textbf{(1)} our approximate retrieval method achieves performance comparable to that of the accurate method ($-0.03 \%$ $\mu AP$), and even comparable to using the pattern ground truth ($-0.34 \%$ $\mu AP$). It significantly outperforms the random method ($+6.23 \%$ $\mu AP$). \textbf{(2)} The zero-shot setting surpasses the baseline by $6.87\%$ in $\mu AP$ and $5.90\%$ in $recall@1$, demonstrating the generalizability of our method. \textbf{(3)} It is non-trivial that using the query itself and its original image as the image-replica nearly achieves $100\%$ performance (see Table \ref{Table: discuss}: without our AnyPattern, using the query itself and its original image can only improve the $\mu AP$ from $16.18\%$ to $18.11\%$), providing evidence of the emergence of in-context learning again. \par 

\begin{figure}[t]
    \centering
    \includegraphics[width=8.5cm]{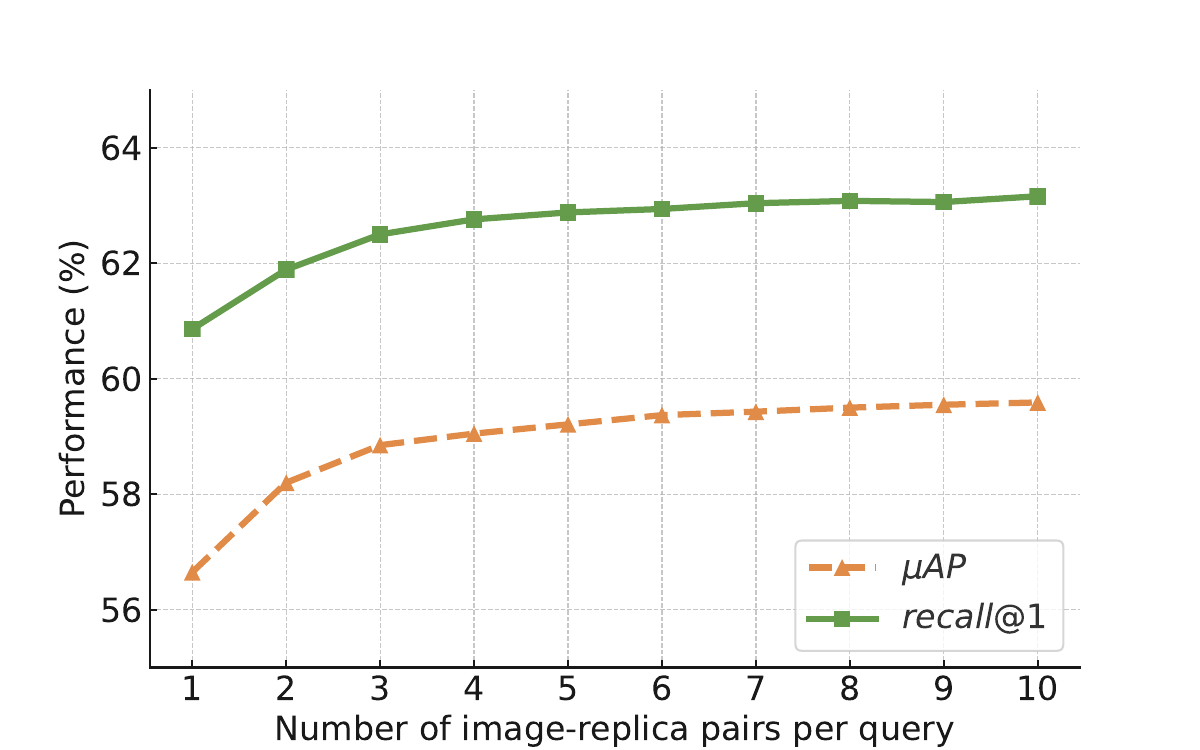}
    \vspace{-3mm} 
    \caption{Leveraging multiple image-replica pairs per query: the change of $\mu AP$ and $recall@1$ in relation to the number of image-replica pairs per query.} 
    \label{Fig: example}
    \vspace*{-4mm}
\end{figure}

\begin{figure}[t]
    \centering
    \includegraphics[width=8.5cm]{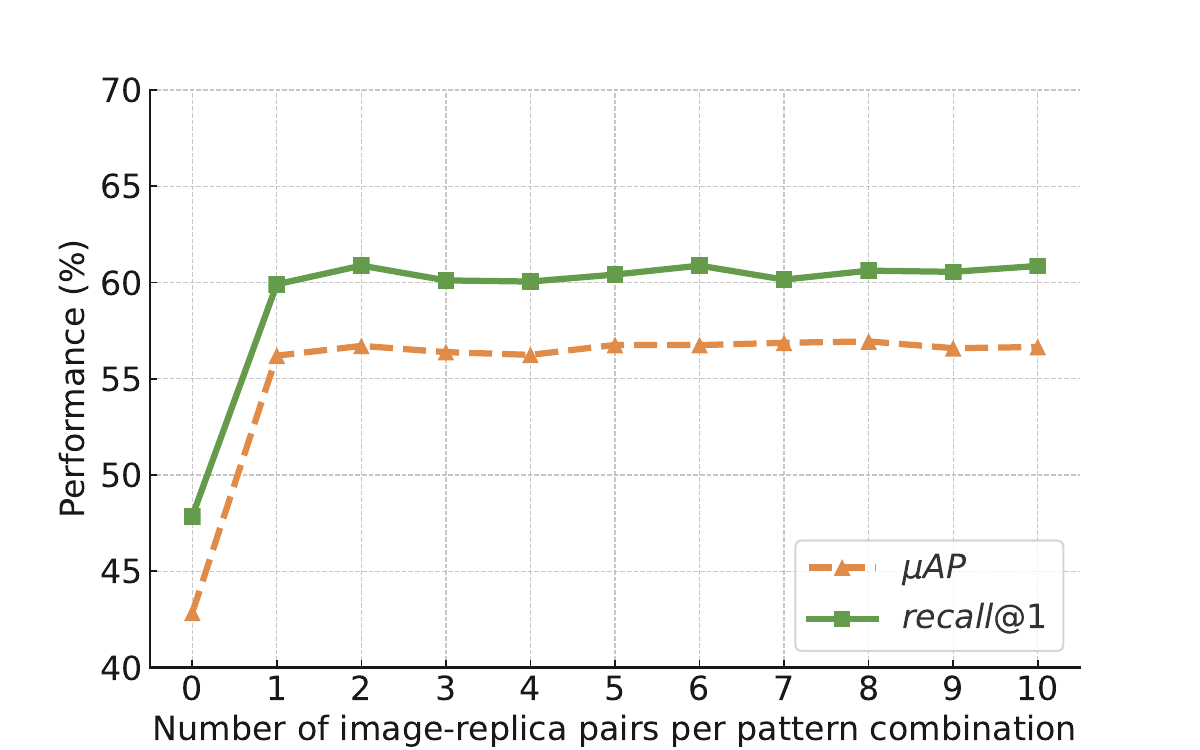}
    \vspace{-3mm} 
    \caption{The change of $\mu AP$ and $recall@1$ in relation to the number of image-replica pairs per pattern combination in the pool.} 
    \label{Fig: pool}
    \vspace*{-4mm}
\end{figure}

\textbf{Our pattern retrieval method achieves high accuracy.} This section shows the accuracy of our pattern retrieval methods achieve. The accuracy is defined as
\begin{equation}
 acc\  =\  \frac{1}{N} \sum_{i=1}^{N} acc_{i}=\frac{1}{N} \sum_{i=1}^{N} \left( \frac{\# \left( P_{q_{i}}\cap P_{s_{i}} \right)}{{} \# \left( P_{q_{i}} \right)} \right),
\end{equation}
where $N$ is the number of queries; $P_{q_{i}}$ and $ P_{s_{i}}$ represent the sets of patterns contained in the query $q_{i}$ and its example image $s_{i}$, respectively; $\# \left( \cdot \right)  $ denotes a counting function; and $\cap$ represents the intersection of sets. We use the top retrieved example image of each model. \par 
The pattern retrieval accuracy and the corresponding performance for the four retrieval methods are displayed in Table \ref{Table: re}. Our observations are as follows: (1) compared to random selection, our method of approximate pattern retrieval shows an improvement of $+32.79\%$ in $acc$, leading to a $+6.23\%$ improvement in $\mu AP$; (2) while using ground truths can achieve $100 \%$ accuracy in pattern retrieval, there is little to no room for further improvement in $\mu AP$. These findings indicate that, although our pattern retrieval is not perfect, it is relatively sufficient to retrieve an example pair to condition the feed-forward process.

\begin{figure}[t]
    \centering
    \includegraphics[width=8.5cm]{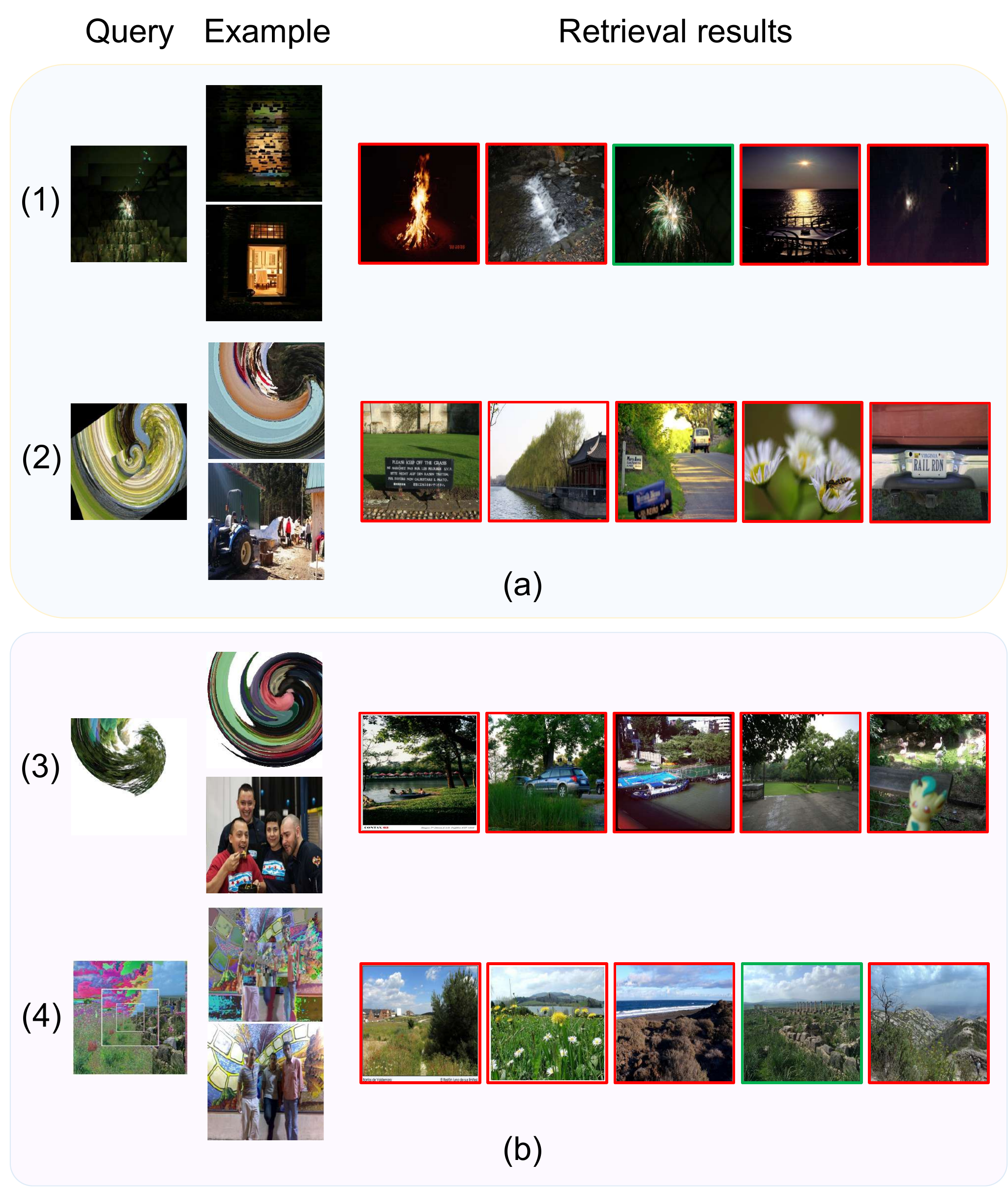}
    \vspace{-3mm} 
    \caption{The failure cases of our method include: (a) the \textbf{example pair} is \textbf{incorrect}, as in case (1), where the query contains the \texttt{Pyramid} pattern, while the example contains \texttt{Mosaic} and \texttt{AutoSeg} patterns; and (b) the presence of numerous \textbf{visual distractors}, as in case (4), where some references depict scenarios that are inherently similar to the original image. The original image is highlighted in \textbf{\textcolor{green}{green}}, while incorrect matches are indicated in \textbf{\textcolor{red}{red}}.} 
    \label{Fig: fail}
    \vspace*{-4mm}
\end{figure}

\textbf{Utilizing multiple image-replica pairs per query further improves the performance.} We also discover that leveraging multiple image-replica pairs for a given query can further enhance performance. For each query, we retrieve its top-$N$ image-replica pairs and repeat the $9$-channel ImageStacker $N$ times. The final similarity between a query and a gallery is then calculated as the maximum value among $N$ similarities. The performance change in relation to the number of image-replica pairs is depicted in Fig. \ref{Fig: example}. Compared to using only one image-replica pair, employing $10$ image-replica pairs boosts the $\mu AP$ to $59.59\%$ ($+2.94\%$) and $recall@1$ to $63.16\%$ ($+2.30\%$). 

\textbf{Our method remains effective even with fewer image-replica pairs in the pool.} Currently, the AnyPattern pool contains 10 image-replica pairs for each pattern combination. As shown in Fig. \ref{Fig: pool}, we find that for our ImageStacker, one image-replica pair per pattern combination is sufficient to improve performance. This implies that when we utilize only one image-replica pair per query for inference, 10 image-replica pairs per pattern combination in the pool achieves similar performance to just one pair. This further demonstrates the practicality of our method: when operators of a deployed ICD system identify novel patterns, they need only manually add one example pair to the pool.

\vspace{-4mm}
\subsection{\textbf{Failure} Cases}
\vspace{-2mm}

As shown in Fig. \ref{Fig: fail}, we conclude two failure cases: \textbf{(1)} Our method may fail when it retrieves incorrect example pairs. This is reasonable because in-context learning fundamentally relies on the example pairs to ``learn''; and \textbf{(2)} Our method may fail when there are many visual distractors. The hard negative problem is indeed a longstanding issue for ICD. To further enhance performance, future work could focus on addressing these two types of failure cases.

\vspace{-4mm}
\section{Conclusion} \label{sec: con}
This paper considers a practical scenario, \textit{i.e.} in-context Image Copy Detection (ICD). Unlike the standard updating of ICD, in-context ICD aims to prompt a trained model to recognize novel-patterned replication using a few example pairs, without requiring re-training. To advance research on in-context ICD, we present AnyPattern, a dataset featuring $100$ tampering patterns. We further propose ImageStacker, a method that directly stacks an example pair onto a query along the channel dimension. The stacking design conditions (modifies) the query feature and thus enables a better matching for the novel patterns. Experimental results highlight the substantial performance improvement gained by AnyPattern and ImageStacker. We hope our work draws research attention to the critical real-world problem in ICD systems, \textit{i.e.}, the fast reaction against novel patterns. Beyond the problem of ICD, we also explore the value of AnyPattern in identifying style mimicry by text-to-image diffusion models.

\textbf{Limitations and future work.} Although training ImageStacker with AnyPattern significantly improves performance on novel patterns, it still falls short compared to base patterns. Based on AnyPattern, future work may focus on developing more effective and efficient in-context learning methods to reduce overfitting on base patterns and close this performance gap.

\newpage
\clearpage
\begin{appendices}
\begin{center}
\section*{\large{Appendix}}
\end{center}
\section{Demonstration of the AnyPattern dataset} \label{App: demo}
\begin{figure}[h]
    \centering
    \vspace{10mm}
   \includegraphics[width=7cm]{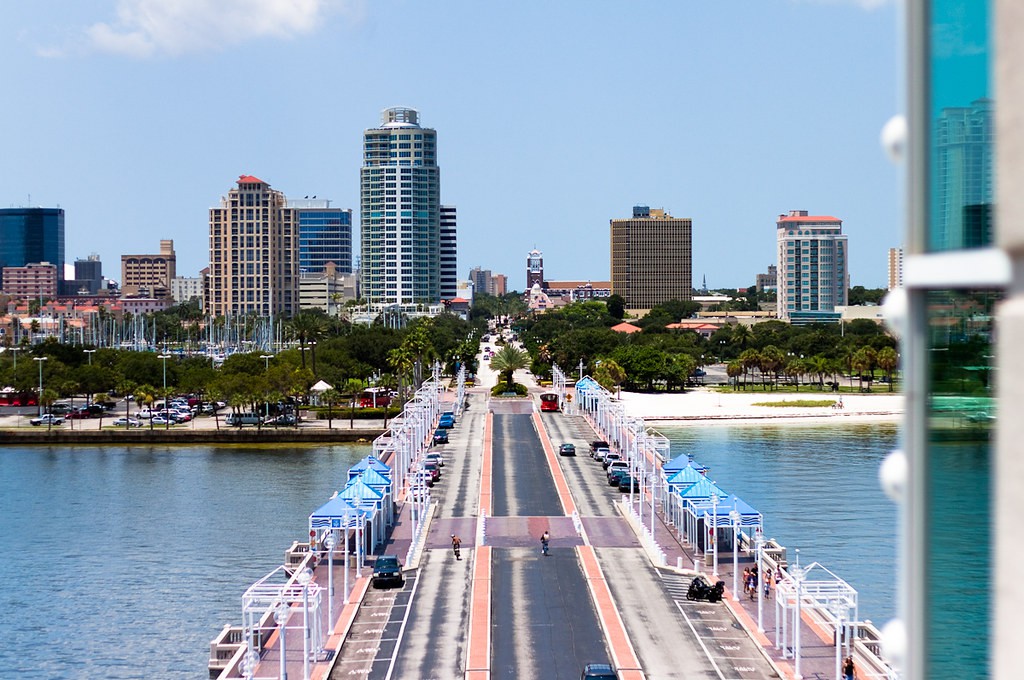}
    \vspace{2mm} 
   \caption{The original image that used to add $100$ different patterns} 
   \label{Fig: origin}
\end{figure}

This section shows the $100$ patterns utilized in the creation of the AnyPattern dataset. The original image is presented in Fig. \ref{Fig: origin}; the generated replicas are in the following tables, with the names of the training patterns indicated in \textit{black}, and the names of the test patterns depicted in \textit{\textcolor{blue}{blue}}. The majority of the constructed patterns includes a degree of randomness while a minority of the patterns remain constant. To demonstrate the variability, each pattern is replicated four times for a single image. To our knowledge, the assembled AnyPattern is the most extensive pattern set currently available. The online version of this demonstration is available at \url{https://huggingface.co/datasets/WenhaoWang/AnyPattern/viewer}.
\vspace{10mm}

\resizebox{0.97\textwidth}{!}{
\normalsize
\begin{tabularx}{\textwidth}{|m{0.9cm}||m{2.5cm}|m{2.8cm}|m{2.8cm}|m{2.8cm}|m{2.8cm}|}
  \hline\thickhline
    
   \rowcolor{mygray}
  \multicolumn{1}{|c||}{Pattern} & \multicolumn{1}{c|}{Elaboration} & \multicolumn{1}{c|}{Demo 1} & \multicolumn{1}{c|}{Demo 2} & \multicolumn{1}{c|}{Demo 3} & \multicolumn{1}{c|}{Demo 4} \\
\hline\hline
    \centering \adjustbox{angle=90}{\texttt{ResizeCrop}} & Randomly crop and resize an image to a specified size. & 
  \vspace{1mm} \includegraphics[width=0.157\textwidth]{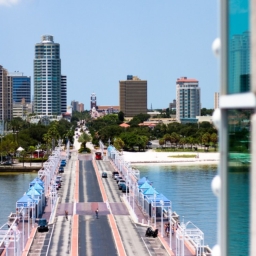} & 
  \vspace{1mm} \includegraphics[width=0.157\textwidth]{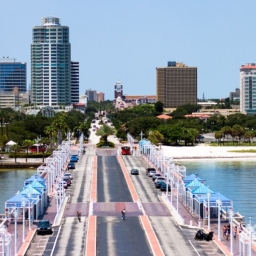} &
  \vspace{1mm} \includegraphics[width=0.157\textwidth]{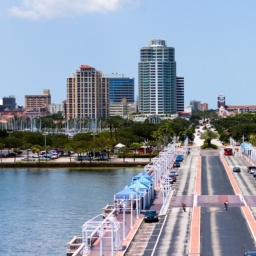} &
  \vspace{1mm} \includegraphics[width=0.157\textwidth]{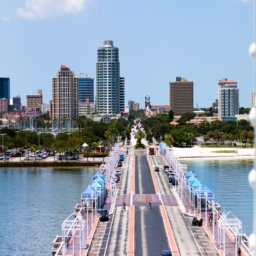} \\[20pt]
  \hline 
  \centering \adjustbox{angle=90}{\texttt{Blend}} & Blend two images with random transparency.  & 
  \vspace{1mm} \includegraphics[width=0.157\textwidth]{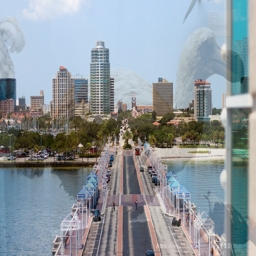} & 
  \vspace{1mm} \includegraphics[width=0.157\textwidth]{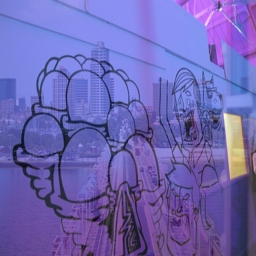} &
  \vspace{1mm} \includegraphics[width=0.157\textwidth]{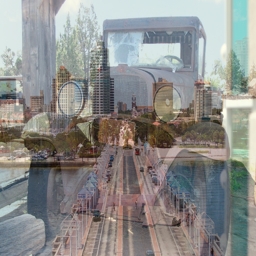} &
  \vspace{1mm} \includegraphics[width=0.157\textwidth]{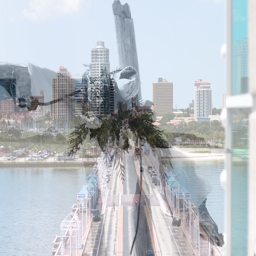} \\
  \hline 
  \centering \adjustbox{angle=90}{\texttt{GrayScale}} & Convert an image into grayscale.  & 
  \vspace{1mm} \includegraphics[width=0.157\textwidth]{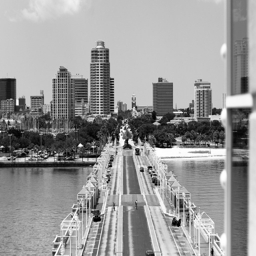} & 
  \vspace{1mm} \includegraphics[width=0.157\textwidth]{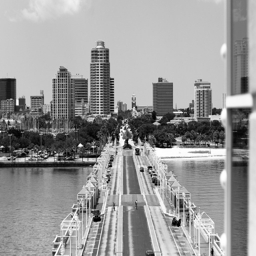} &
  \vspace{1mm} \includegraphics[width=0.157\textwidth]{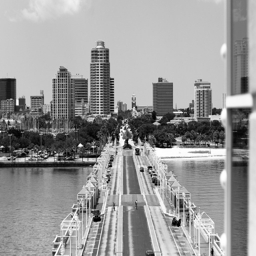} &
  \vspace{1mm} \includegraphics[width=0.157\textwidth]{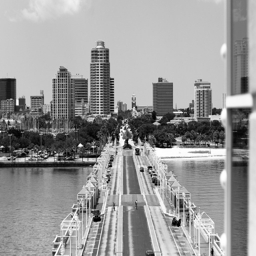} \\
  \hline 
  \centering \adjustbox{angle=90}{\texttt{ColorJitter}}  & Randomly change the brightness, contrast, saturation, and hue of an image.  & 
  \vspace{1mm} \includegraphics[width=0.157\textwidth]{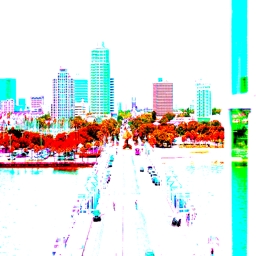} & 
  \vspace{1mm} \includegraphics[width=0.157\textwidth]{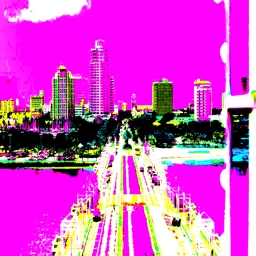} &
  \vspace{1mm} \includegraphics[width=0.157\textwidth]{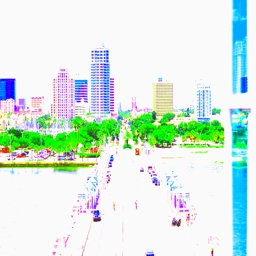} &
  \vspace{1mm} \includegraphics[width=0.157\textwidth]{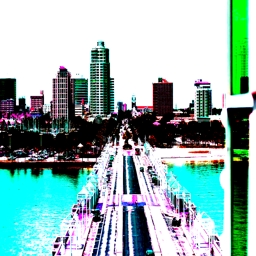}\\
  \hline
\end{tabularx}}

\begin{table*}[t]
\centering
\resizebox{0.97\textwidth}{!}{
\normalsize
 \begin{tabularx}{\textwidth}{|m{0.9cm}||m{2.5cm}|m{2.8cm}|m{2.8cm}|m{2.8cm}|m{2.8cm}|}
  \hline\thickhline
    
   \rowcolor{mygray}
  \multicolumn{1}{|c||}{Pattern} & \multicolumn{1}{c|}{Elaboration} & \multicolumn{1}{c|}{Demo 1} & \multicolumn{1}{c|}{Demo 2} & \multicolumn{1}{c|}{Demo 3} & \multicolumn{1}{c|}{Demo 4} \\
\hline\hline
     
\centering \adjustbox{angle=90}{ \texttt{Blur}} & Randomly apply a blur filter to an image.  & 
\vspace{1mm} \includegraphics[width=0.157\textwidth]{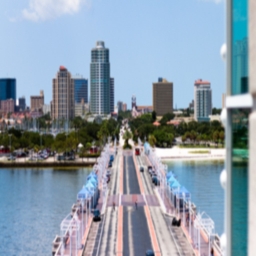} & 
\vspace{1mm} \includegraphics[width=0.157\textwidth]{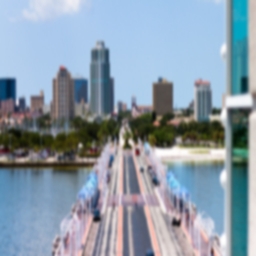} &
\vspace{1mm} \includegraphics[width=0.157\textwidth]{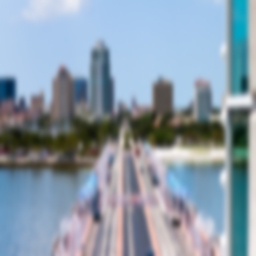} &
\vspace{1mm} \includegraphics[width=0.157\textwidth]{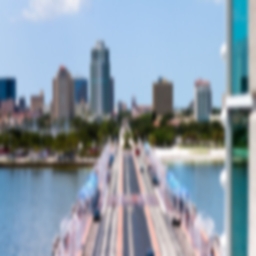}\\
\hline
\centering \adjustbox{angle=90}{ \texttt{Pixelate}} &  Pixelate random portions of an image.  & 
\vspace{1mm} \includegraphics[width=0.157\textwidth]{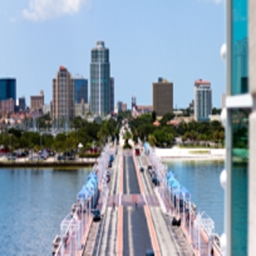} & 
\vspace{1mm} \includegraphics[width=0.157\textwidth]{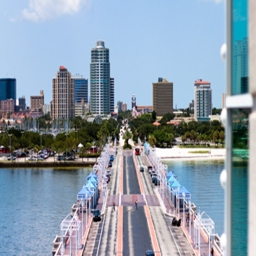} &
\vspace{1mm} \includegraphics[width=0.157\textwidth]{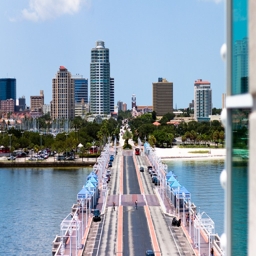} &
\vspace{1mm} \includegraphics[width=0.157\textwidth]{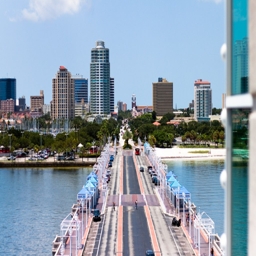}\\
\hline
\centering \adjustbox{angle=90}{ \texttt{Rotate}}  & Randomly rotate an image within a given range of degrees.  & 
\vspace{1mm} \includegraphics[width=0.157\textwidth]{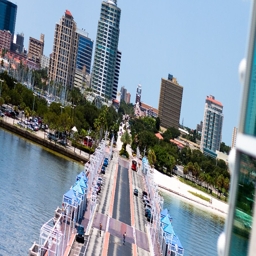} & 
\vspace{1mm} \includegraphics[width=0.157\textwidth]{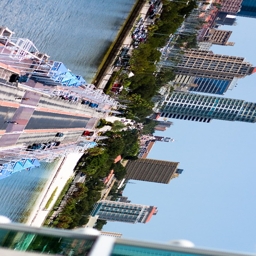} &
\vspace{1mm} \includegraphics[width=0.157\textwidth]{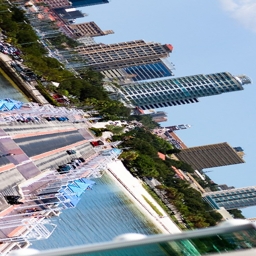} &
\vspace{1mm} \includegraphics[width=0.157\textwidth]{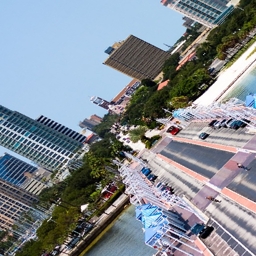}\\
\hline
\centering \adjustbox{angle=90}{ \texttt{Padding}}  & Pad an image with random colors, height and width.  & 
\vspace{1mm} \includegraphics[width=0.157\textwidth]{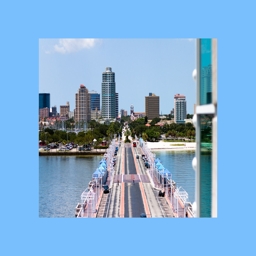} & 
\vspace{1mm} \includegraphics[width=0.157\textwidth]{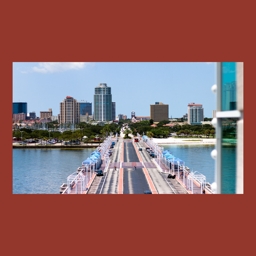} &
\vspace{1mm} \includegraphics[width=0.157\textwidth]{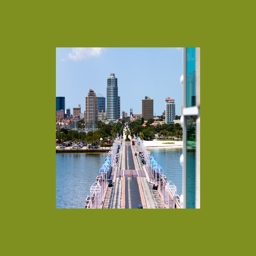} &
\vspace{1mm} \includegraphics[width=0.157\textwidth]{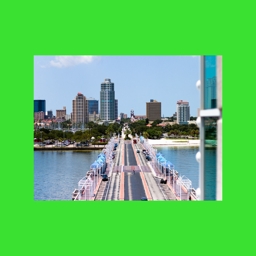}\\
\hline
\centering \adjustbox{angle=90}{ \texttt{AddNoise}}  &  Add random noise to an image.  & 
\vspace{1mm} \includegraphics[width=0.157\textwidth]{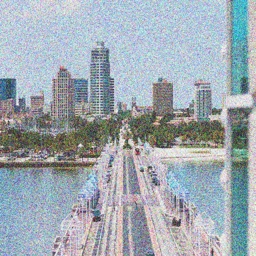} & 
\vspace{1mm} \includegraphics[width=0.157\textwidth]{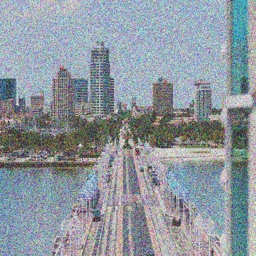} &
\vspace{1mm} \includegraphics[width=0.157\textwidth]{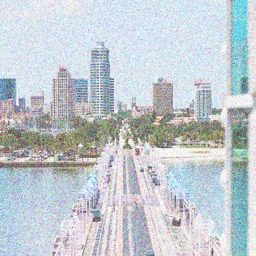} &
\vspace{1mm} \includegraphics[width=0.157\textwidth]{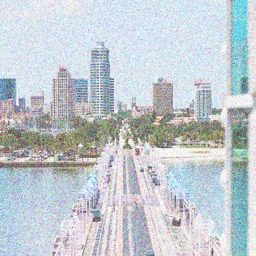}\\
\hline
\centering \adjustbox{angle=90}{ \texttt{VertFlip}}  &  Flip an image vertically.  & 
\vspace{1mm} \includegraphics[width=0.157\textwidth]{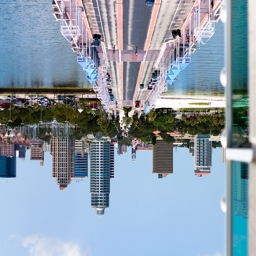} & 
\vspace{1mm} \includegraphics[width=0.157\textwidth]{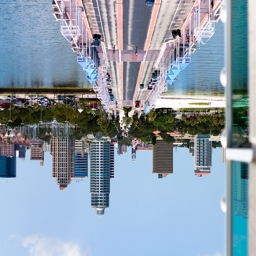} &
\vspace{1mm} \includegraphics[width=0.157\textwidth]{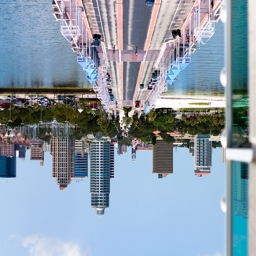} &
\vspace{1mm} \includegraphics[width=0.157\textwidth]{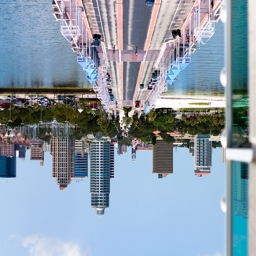}\\
\hline
\centering \adjustbox{angle=90}{ \texttt{HoriFlip}}  &  Flip an image horizontally.  & 
\vspace{1mm} \includegraphics[width=0.157\textwidth]{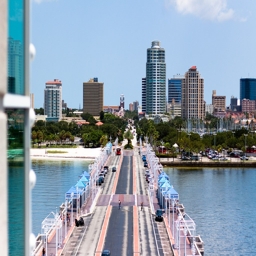} & 
\vspace{1mm} \includegraphics[width=0.157\textwidth]{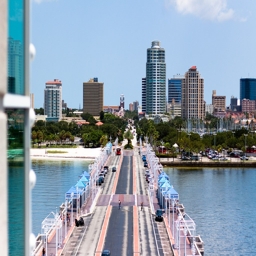} &
\vspace{1mm} \includegraphics[width=0.157\textwidth]{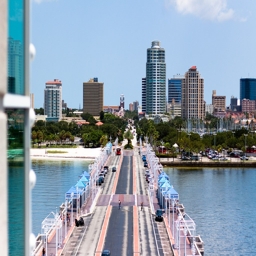} &
\vspace{1mm} \includegraphics[width=0.157\textwidth]{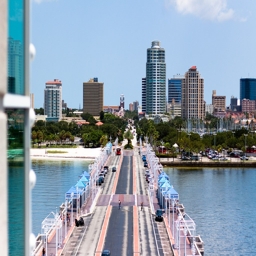}\\
\hline
\centering \adjustbox{angle=90}{ \texttt{MemeFormat}}  &  Add a random meme onto an image.  & 
\vspace{1mm} \includegraphics[width=0.157\textwidth]{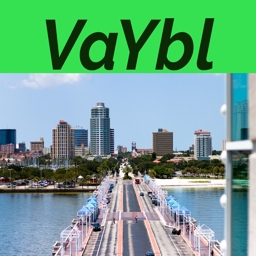} & 
\vspace{1mm} \includegraphics[width=0.157\textwidth]{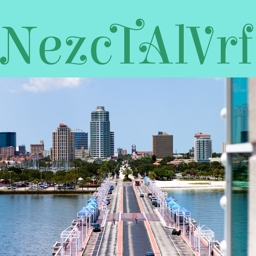} &
\vspace{1mm} \includegraphics[width=0.157\textwidth]{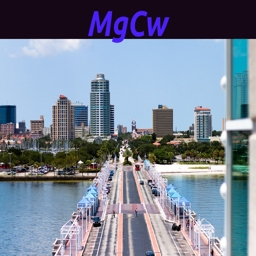} &
\vspace{1mm} \includegraphics[width=0.157\textwidth]{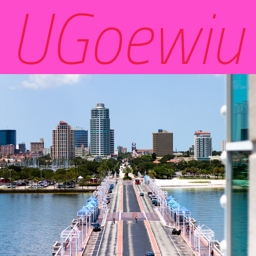}\\
\hline
  \end{tabularx}}
\end{table*}

\begin{table*}[t]
\centering
\resizebox{0.97\textwidth}{!}{
\normalsize
  \begin{tabularx}{\textwidth}{|m{0.9cm}||m{2.5cm}|m{2.8cm}|m{2.8cm}|m{2.8cm}|m{2.8cm}|}
  \hline\thickhline
    
   \rowcolor{mygray}
  \multicolumn{1}{|c||}{Pattern} & \multicolumn{1}{c|}{Elaboration} & \multicolumn{1}{c|}{Demo 1} & \multicolumn{1}{c|}{Demo 2} & \multicolumn{1}{c|}{Demo 3} & \multicolumn{1}{c|}{Demo 4} \\
\hline\hline

\centering \adjustbox{angle=90}{ \texttt{OverlayEmoji}}  &  Overlay a random Emoji onto an image.  & 
\vspace{1mm} \includegraphics[width=0.157\textwidth]{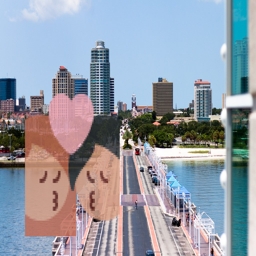} & 
\vspace{1mm} \includegraphics[width=0.157\textwidth]{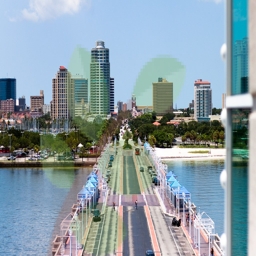} &
\vspace{1mm} \includegraphics[width=0.157\textwidth]{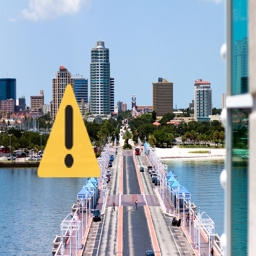} &
\vspace{1mm} \includegraphics[width=0.157\textwidth]{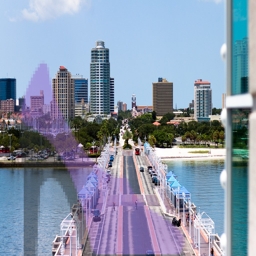}\\
\hline

\centering \adjustbox{angle=90}{ \texttt{OverlayText}}  &  Overlay some random texts onto an image.  & 
\vspace{1mm} \includegraphics[width=0.157\textwidth]{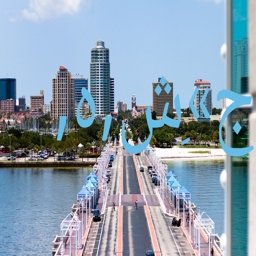} & 
\vspace{1mm} \includegraphics[width=0.157\textwidth]{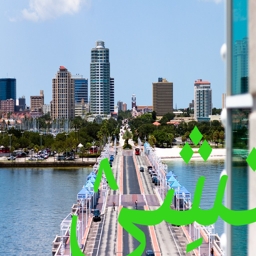} &
\vspace{1mm} \includegraphics[width=0.157\textwidth]{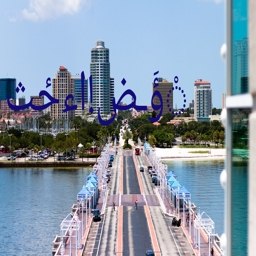} &
\vspace{1mm} \includegraphics[width=0.157\textwidth]{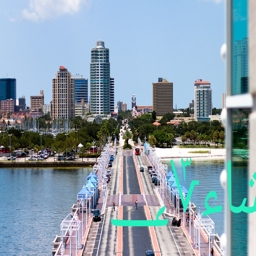}\\
\hline

\centering \adjustbox{angle=90}{ \texttt{PerspChange}}  &  Randomly transform the perspective of an image.  & 
\vspace{1mm} \includegraphics[width=0.157\textwidth]{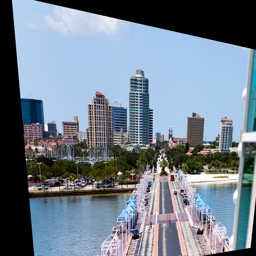} & 
\vspace{1mm} \includegraphics[width=0.157\textwidth]{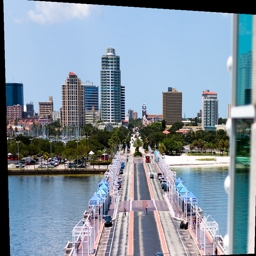} &
\vspace{1mm} \includegraphics[width=0.157\textwidth]{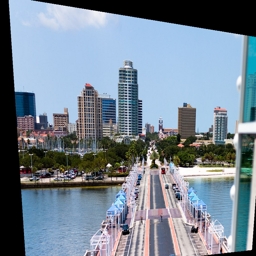} &
\vspace{1mm} \includegraphics[width=0.157\textwidth]{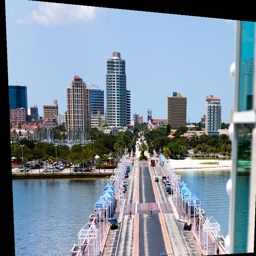}\\
\hline

\centering \adjustbox{angle=90}{ \texttt{OverlayImage}}  &  Randomly overlay an image onto another one.  & 
\vspace{1mm} \includegraphics[width=0.157\textwidth]{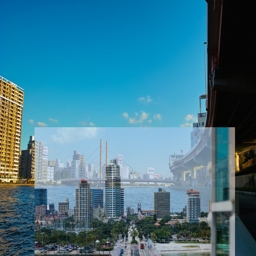} & 
\vspace{1mm} \includegraphics[width=0.157\textwidth]{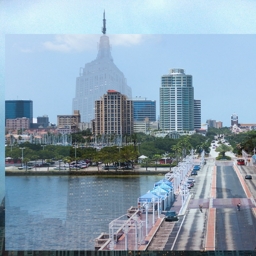} &
\vspace{1mm} \includegraphics[width=0.157\textwidth]{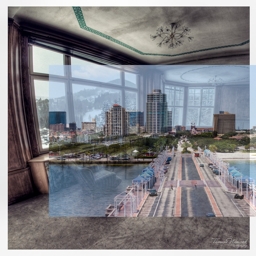} &
\vspace{1mm} \includegraphics[width=0.157\textwidth]{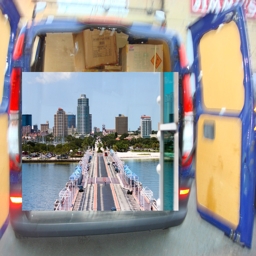}\\
\hline

\centering \adjustbox{angle=90}{ \texttt{StackImage}}  &  Randomly stack two images along the height or width dimension.  & 
\vspace{1mm} \includegraphics[width=0.157\textwidth]{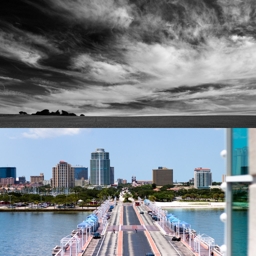} & 
\vspace{1mm} \includegraphics[width=0.157\textwidth]{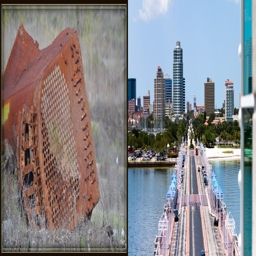} &
\vspace{1mm} \includegraphics[width=0.157\textwidth]{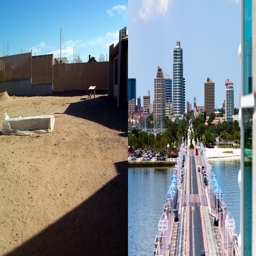} &
\vspace{1mm} \includegraphics[width=0.157\textwidth]{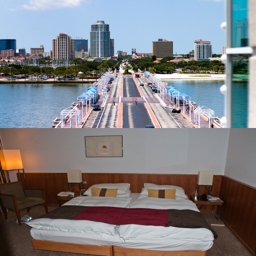}\\
\hline

\centering \adjustbox{angle=90}{ \texttt{ChangeChan}}  &  Randomly shift, swap, or invert the channel of an image \textcolor{blue}{\citep{papadakis2021producing}}.  & 
\vspace{1mm} \includegraphics[width=0.157\textwidth]{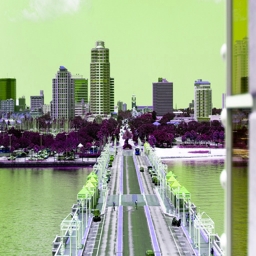} & 
\vspace{1mm} \includegraphics[width=0.157\textwidth]{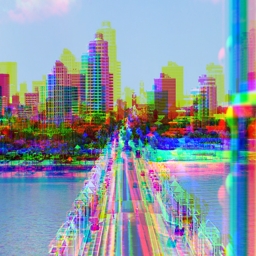} &
\vspace{1mm} \includegraphics[width=0.157\textwidth]{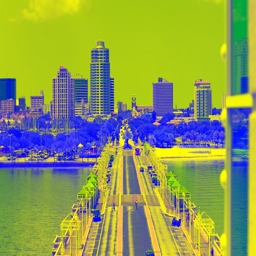} &
\vspace{1mm} \includegraphics[width=0.157\textwidth]{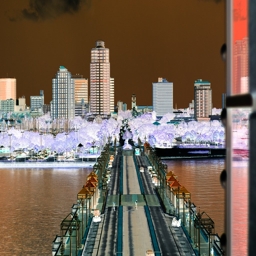}\\
\hline

\centering \adjustbox{angle=90}{ \texttt{EncQuality}}  &  Randomly encode (reduce) the quality of an image.  & 
\vspace{1mm} \includegraphics[width=0.157\textwidth]{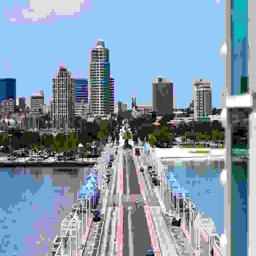} & 
\vspace{1mm} \includegraphics[width=0.157\textwidth]{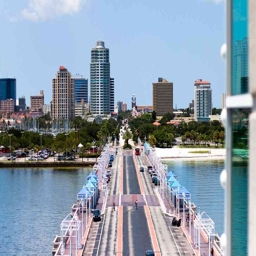} &
\vspace{1mm} \includegraphics[width=0.157\textwidth]{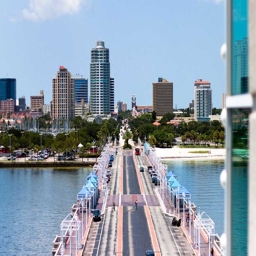} &
\vspace{1mm} \includegraphics[width=0.157\textwidth]{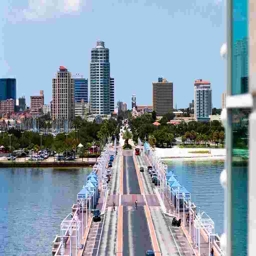}\\
\hline

\centering \adjustbox{angle=90}{ \texttt{AddStripes}}  &  Randomly add different colors of stripes onto an image.  & 
\vspace{1mm} \includegraphics[width=0.157\textwidth]{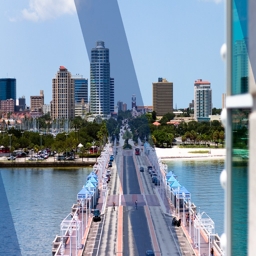} & 
\vspace{1mm} \includegraphics[width=0.157\textwidth]{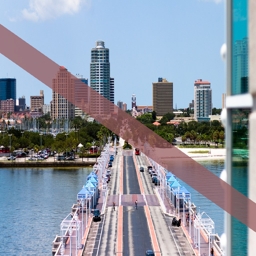} &
\vspace{1mm} \includegraphics[width=0.157\textwidth]{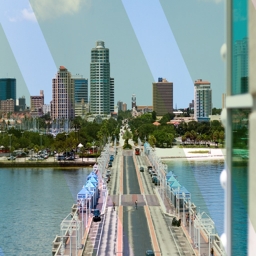} &
\vspace{1mm} \includegraphics[width=0.157\textwidth]{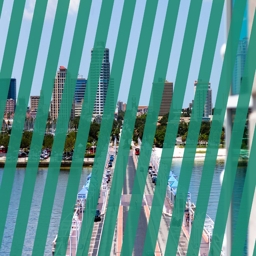}\\
\hline

  \end{tabularx}}
\end{table*}

\begin{table*}[t]
\centering
\resizebox{0.97\textwidth}{!}{
\normalsize
  \begin{tabularx}{\textwidth}{|m{0.9cm}||m{2.5cm}|m{2.8cm}|m{2.8cm}|m{2.8cm}|m{2.8cm}|}
  \hline\thickhline
    
   \rowcolor{mygray}
  \multicolumn{1}{|c||}{Pattern} & \multicolumn{1}{c|}{Elaboration} & \multicolumn{1}{c|}{Demo 1} & \multicolumn{1}{c|}{Demo 2} & \multicolumn{1}{c|}{Demo 3} & \multicolumn{1}{c|}{Demo 4} \\
\hline\hline
\centering \adjustbox{angle=90}{ \texttt{Sharpen}}  &  Randomly enhance the edge contrast of an image.  & 
\vspace{1mm} \includegraphics[width=0.157\textwidth]{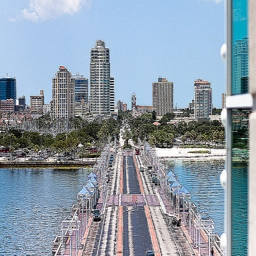} & 
\vspace{1mm} \includegraphics[width=0.157\textwidth]{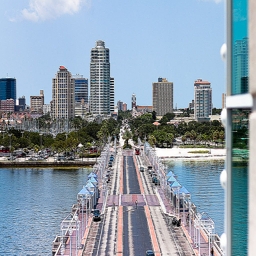} &
\vspace{1mm} \includegraphics[width=0.157\textwidth]{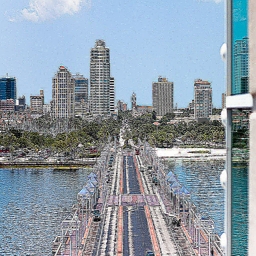} &
\vspace{1mm} \includegraphics[width=0.157\textwidth]{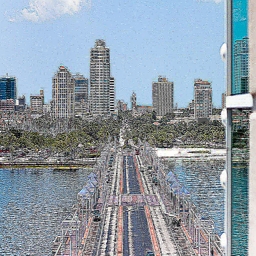}\\
\hline

\centering \adjustbox{angle=90}{ \texttt{Skew}}  &  Randomly skew an image by a certain angle.  & 
\vspace{1mm} \includegraphics[width=0.157\textwidth]{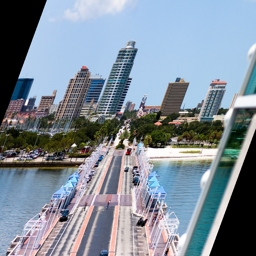} & 
\vspace{1mm} \includegraphics[width=0.157\textwidth]{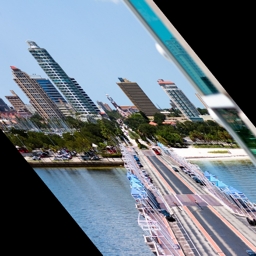} &
\vspace{1mm} \includegraphics[width=0.157\textwidth]{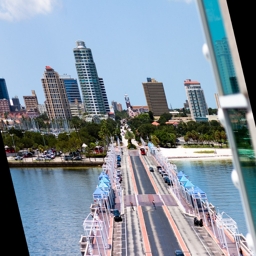} &
\vspace{1mm} \includegraphics[width=0.157\textwidth]{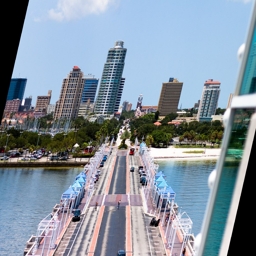}\\
\hline

\centering \adjustbox{angle=90}{ \texttt{ShufPixels}}  &  Randomly rearrange (shuffle) the pixels within an image.  & 
\vspace{1mm} \includegraphics[width=0.157\textwidth]{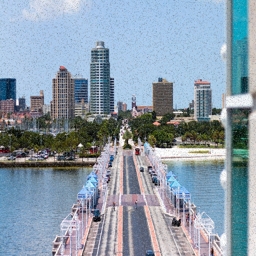} & 
\vspace{1mm} \includegraphics[width=0.157\textwidth]{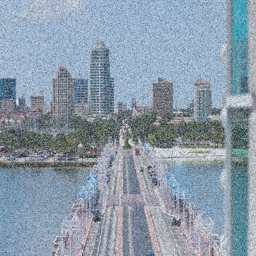} &
\vspace{1mm} \includegraphics[width=0.157\textwidth]{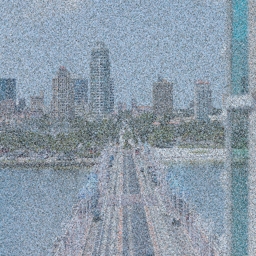} &
\vspace{1mm} \includegraphics[width=0.157\textwidth]{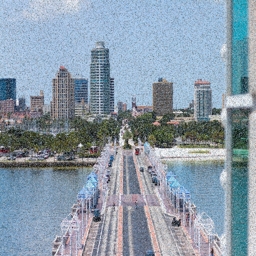}\\
\hline

\centering \adjustbox{angle=90}{ \texttt{AddShapes}}  &  Randomly add shapes (such as ellipse, rectangle, triangle, star, and pentagon) onto an image.  & 
\vspace{1mm} \includegraphics[width=0.157\textwidth]{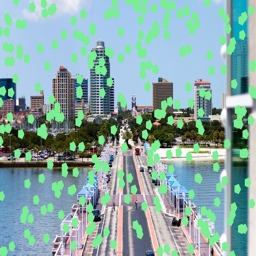} & 
\vspace{1mm} \includegraphics[width=0.157\textwidth]{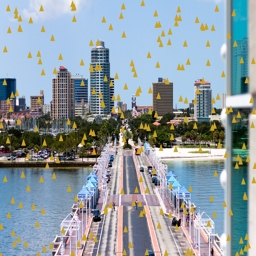} &
\vspace{1mm} \includegraphics[width=0.157\textwidth]{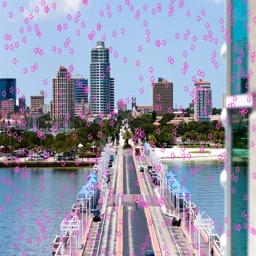} &
\vspace{1mm} \includegraphics[width=0.157\textwidth]{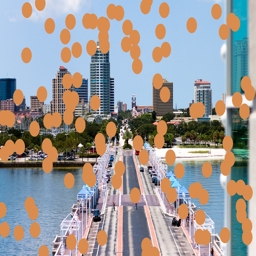}\\
\hline

\centering \adjustbox{angle=90}{ \texttt{Repeat}}  &  Randomly repeat an image for several times.  & 
\vspace{1mm} \includegraphics[width=0.157\textwidth]{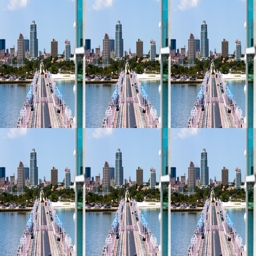} & 
\vspace{1mm} \includegraphics[width=0.157\textwidth]{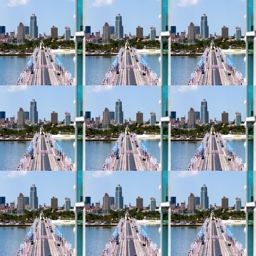} &
\vspace{1mm} \includegraphics[width=0.157\textwidth]{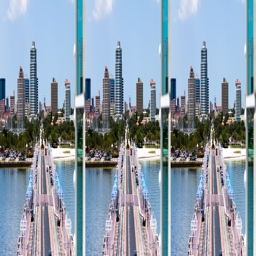} &
\vspace{1mm} \includegraphics[width=0.157\textwidth]{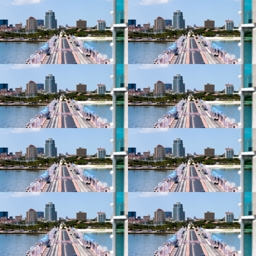}\\
\hline

\centering \adjustbox{angle=90}{ \texttt{CutAssemble}}  &  Randomly cut an image into several cells and shuffle the cells.  & 
\vspace{1mm} \includegraphics[width=0.157\textwidth]{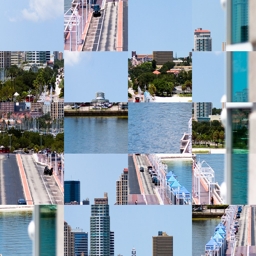} & 
\vspace{1mm} \includegraphics[width=0.157\textwidth]{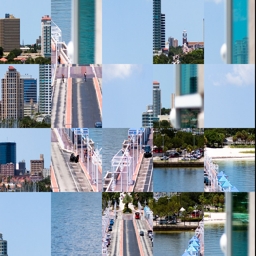} &
\vspace{1mm} \includegraphics[width=0.157\textwidth]{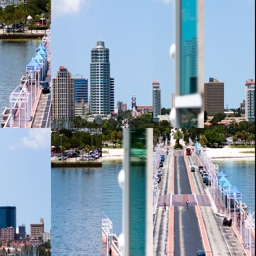} &
\vspace{1mm} \includegraphics[width=0.157\textwidth]{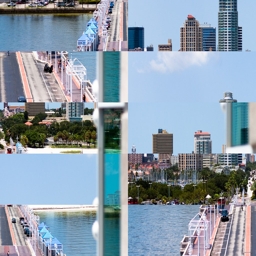}\\
\hline

\centering \adjustbox{angle=90}{ \texttt{BGChange}}  &  Randomly change the background of an image.  & 
\vspace{1mm} \includegraphics[width=0.157\textwidth]{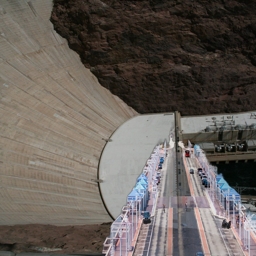} & 
\vspace{1mm} \includegraphics[width=0.157\textwidth]{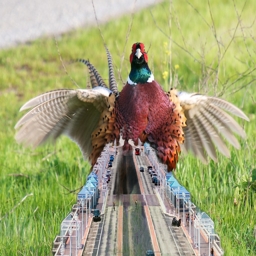} &
\vspace{1mm} \includegraphics[width=0.157\textwidth]{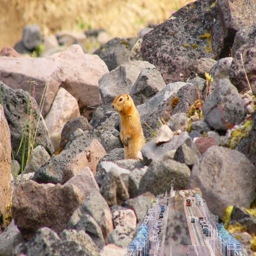} &
\vspace{1mm} \includegraphics[width=0.157\textwidth]{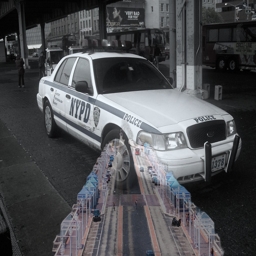}\\
\hline

\centering \adjustbox{angle=90}{ \texttt{Xraylize}}  &  Simulate the effect of an X-ray on an image.  & 
\vspace{1mm} \includegraphics[width=0.157\textwidth]{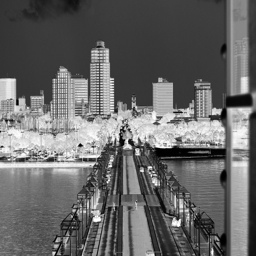} & 
\vspace{1mm} \includegraphics[width=0.157\textwidth]{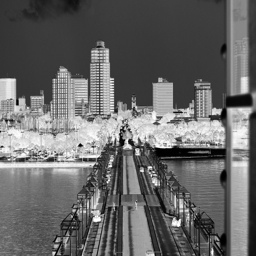} &
\vspace{1mm} \includegraphics[width=0.157\textwidth]{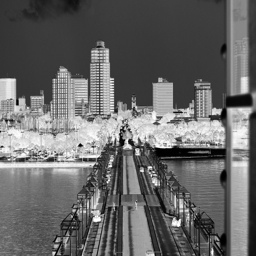} &
\vspace{1mm} \includegraphics[width=0.157\textwidth]{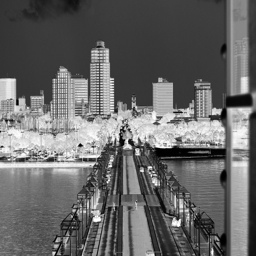}\\
\hline

  \end{tabularx}}
\end{table*}

\begin{table*}[t]
\centering
\resizebox{0.97\textwidth}{!}{
\normalsize
  \begin{tabularx}{\textwidth}{|m{0.9cm}||m{2.5cm}|m{2.8cm}|m{2.8cm}|m{2.8cm}|m{2.8cm}|}
  \hline\thickhline
    
   \rowcolor{mygray}
  \multicolumn{1}{|c||}{Pattern} & \multicolumn{1}{c|}{Elaboration} & \multicolumn{1}{c|}{Demo 1} & \multicolumn{1}{c|}{Demo 2} & \multicolumn{1}{c|}{Demo 3} & \multicolumn{1}{c|}{Demo 4} \\
\hline\hline
\centering \adjustbox{angle=90}{ \texttt{Mirrorize}}  &  Reflect an image along its vertical or horizontal axis.  & 
\vspace{1mm} \includegraphics[width=0.157\textwidth]{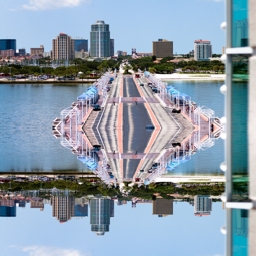} & 
\vspace{1mm} \includegraphics[width=0.157\textwidth]{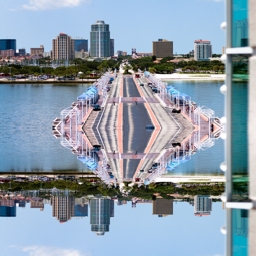} &
\vspace{1mm} \includegraphics[width=0.157\textwidth]{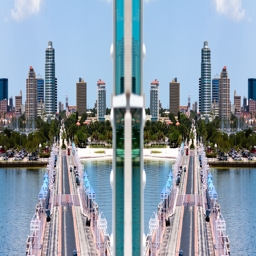} &
\vspace{1mm} \includegraphics[width=0.157\textwidth]{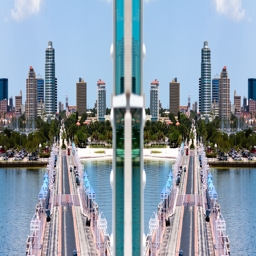}\\
\hline

\centering \adjustbox{angle=90}{ \texttt{Kaleidoscope}}  &  Replicate the effect of a kaleidoscope on an image.  & 
\vspace{1mm} \includegraphics[width=0.157\textwidth]{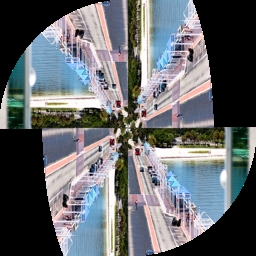} & 
\vspace{1mm} \includegraphics[width=0.157\textwidth]{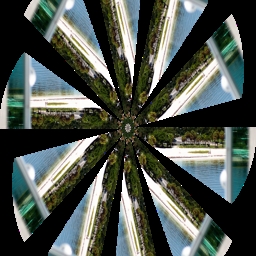} &
\vspace{1mm} \includegraphics[width=0.157\textwidth]{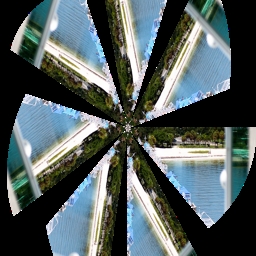} &
\vspace{1mm} \includegraphics[width=0.157\textwidth]{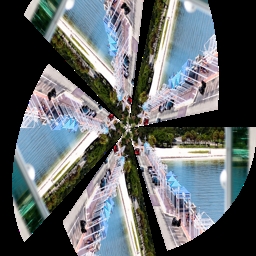}\\
\hline

\centering \adjustbox{angle=90}{ \texttt{EdgeDetect}}  &  Randomly perform edge detection by different filters.  & 
\vspace{1mm} \includegraphics[width=0.157\textwidth]{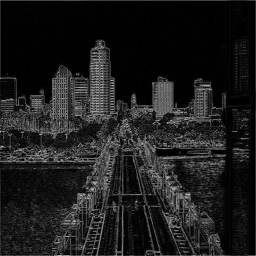} & 
\vspace{1mm} \includegraphics[width=0.157\textwidth]{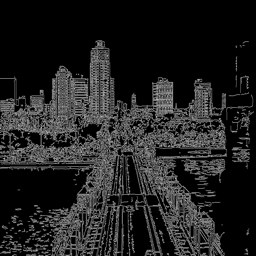} &
\vspace{1mm} \includegraphics[width=0.157\textwidth]{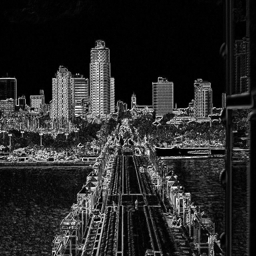} &
\vspace{1mm} \includegraphics[width=0.157\textwidth]{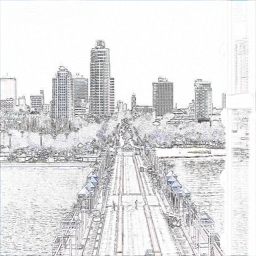}\\
\hline

\centering \adjustbox{angle=90}{ \texttt{GlassEffect}}  &  Add glass effect onto an image with random extent.  & 
\vspace{1mm} \includegraphics[width=0.157\textwidth]{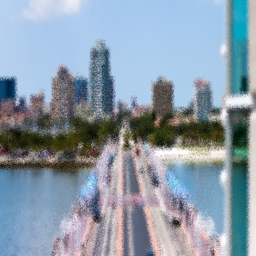} & 
\vspace{1mm} \includegraphics[width=0.157\textwidth]{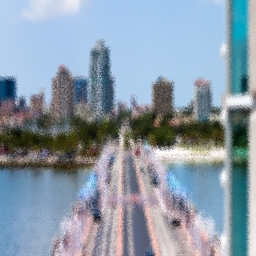} &
\vspace{1mm} \includegraphics[width=0.157\textwidth]{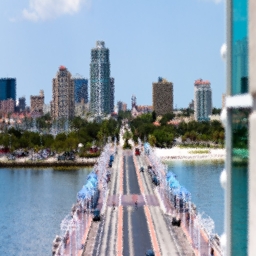} &
\vspace{1mm} \includegraphics[width=0.157\textwidth]{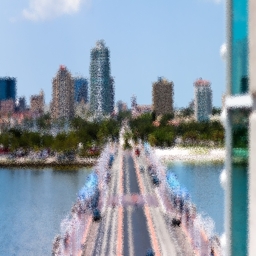}\\
\hline

\centering \adjustbox{angle=90}{ \texttt{OptDistort}}  &  Simulate the effect of viewing an image through a medium that randomly distorts the light.  & 
\vspace{1mm} \includegraphics[width=0.157\textwidth]{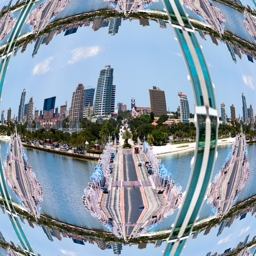} & 
\vspace{1mm} \includegraphics[width=0.157\textwidth]{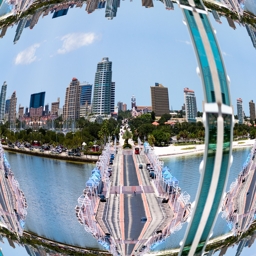} &
\vspace{1mm} \includegraphics[width=0.157\textwidth]{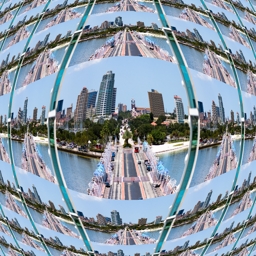} &
\vspace{1mm} \includegraphics[width=0.157\textwidth]{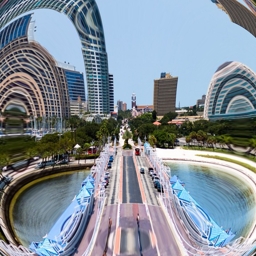}\\
\hline

\centering \adjustbox{angle=90}{ \texttt{Weather}}  &  Add the weather effect (such as rain, fog, clouds, frost, and snow) onto an image.  & 
\vspace{1mm} \includegraphics[width=0.157\textwidth]{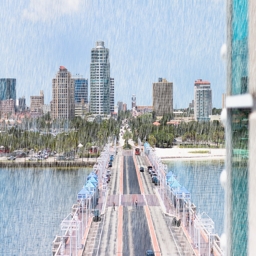} & 
\vspace{1mm} \includegraphics[width=0.157\textwidth]{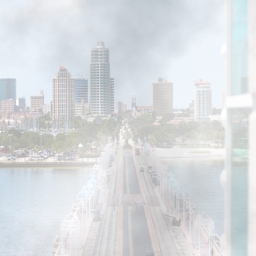} &
\vspace{1mm} \includegraphics[width=0.157\textwidth]{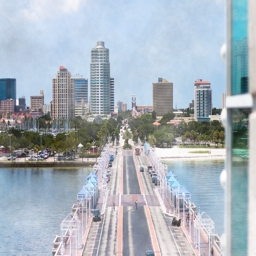} &
\vspace{1mm} \includegraphics[width=0.157\textwidth]{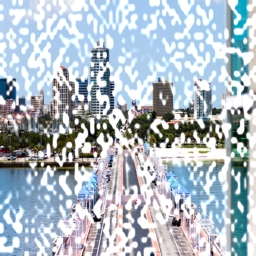}\\
\hline

\centering \adjustbox{angle=90}{ \texttt{SplitRotate}}  &  Randomly split and rotate an image.  & 
\vspace{1mm} \includegraphics[width=0.157\textwidth]{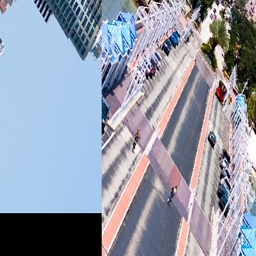} & 
\vspace{1mm} \includegraphics[width=0.157\textwidth]{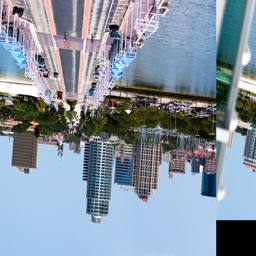} &
\vspace{1mm} \includegraphics[width=0.157\textwidth]{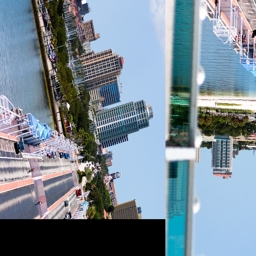} &
\vspace{1mm} \includegraphics[width=0.157\textwidth]{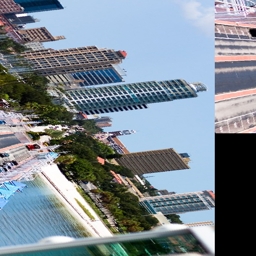}\\
\hline

\centering \adjustbox{angle=90}{ \texttt{Solarize}}  &  Invert all pixel values above a random threshold, creating a peculiar high-contrast effect.  & 
\vspace{1mm} \includegraphics[width=0.157\textwidth]{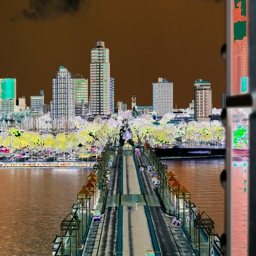} & 
\vspace{1mm} \includegraphics[width=0.157\textwidth]{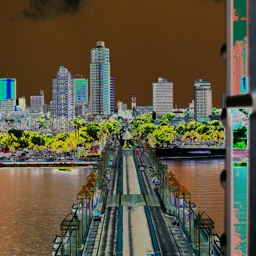} &
\vspace{1mm} \includegraphics[width=0.157\textwidth]{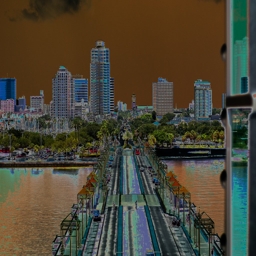} &
\vspace{1mm} \includegraphics[width=0.157\textwidth]{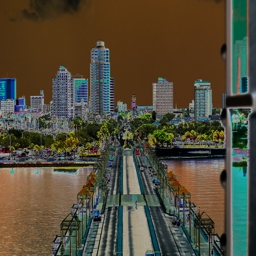}\\
\hline

\end{tabularx}}
\end{table*}

\begin{table*}[t]
\centering
\resizebox{0.97\textwidth}{!}{
\normalsize
 \begin{tabularx}{\textwidth}{|m{0.9cm}||m{2.5cm}|m{2.8cm}|m{2.8cm}|m{2.8cm}|m{2.8cm}|}
  \hline\thickhline
    
   \rowcolor{mygray}
  \multicolumn{1}{|c||}{Pattern} & \multicolumn{1}{c|}{Elaboration} & \multicolumn{1}{c|}{Demo 1} & \multicolumn{1}{c|}{Demo 2} & \multicolumn{1}{c|}{Demo 3} & \multicolumn{1}{c|}{Demo 4} \\
\hline\hline
\centering \adjustbox{angle=90}{ \texttt{Legofy}}  &  Randomly make an image look as if it is made out of $1 \times 1$  LEGO blocks.  & 
\vspace{1mm} \includegraphics[width=0.157\textwidth]{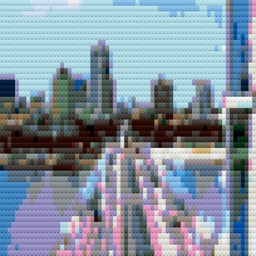} & 
\vspace{1mm} \includegraphics[width=0.157\textwidth]{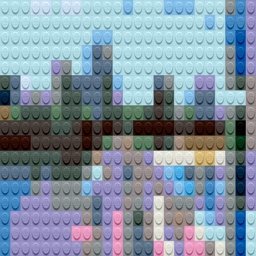} &
\vspace{1mm} \includegraphics[width=0.157\textwidth]{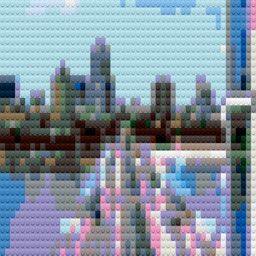} &
\vspace{1mm} \includegraphics[width=0.157\textwidth]{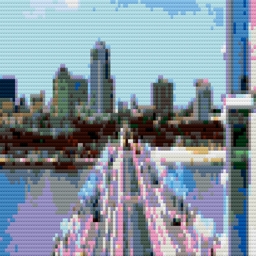}\\
\hline

\centering \adjustbox{angle=90}{ \texttt{Ink}}  &  Randomly make an image look like an ink painting.  & 
\vspace{1mm} \includegraphics[width=0.157\textwidth]{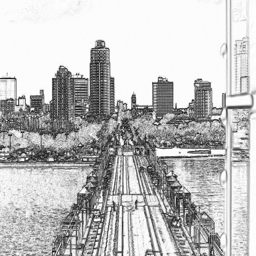} & 
\vspace{1mm} \includegraphics[width=0.157\textwidth]{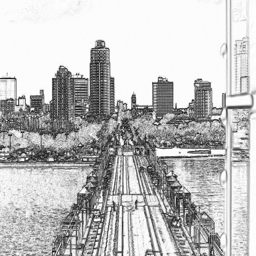} &
\vspace{1mm} \includegraphics[width=0.157\textwidth]{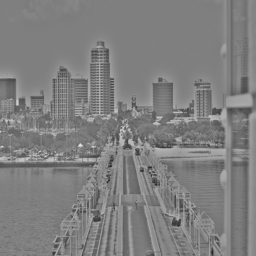} &
\vspace{1mm} \includegraphics[width=0.157\textwidth]{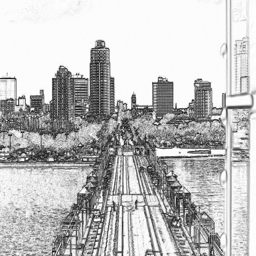}\\
\hline

\centering \adjustbox{angle=90}{ \texttt{Zooming}}  &  Randomly simulate the effect of zooming in or out.  & 
\vspace{1mm} \includegraphics[width=0.157\textwidth]{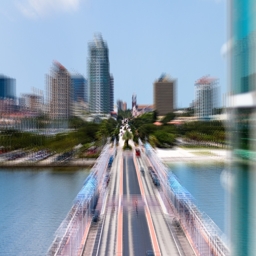} & 
\vspace{1mm} \includegraphics[width=0.157\textwidth]{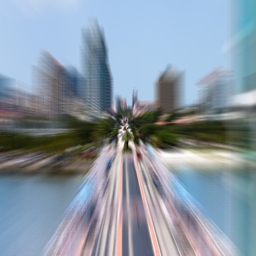} &
\vspace{1mm} \includegraphics[width=0.157\textwidth]{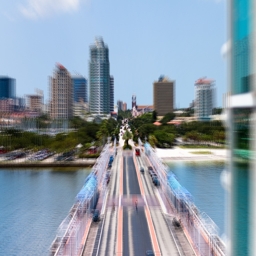} &
\vspace{1mm} \includegraphics[width=0.157\textwidth]{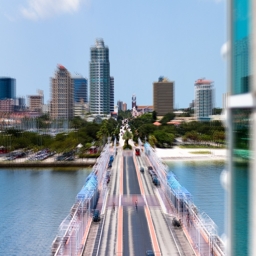}\\
\hline

\centering \adjustbox{angle=90}{ \texttt{CutPaste}}  &  Randomly change two parts in an image.  & 
\vspace{1mm} \includegraphics[width=0.157\textwidth]{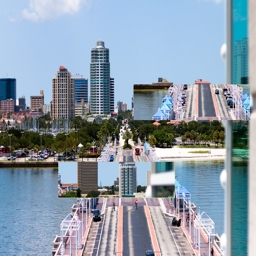} & 
\vspace{1mm} \includegraphics[width=0.157\textwidth]{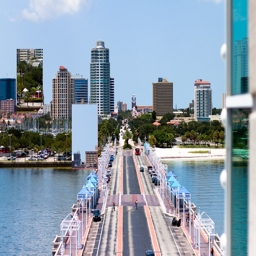} &
\vspace{1mm} \includegraphics[width=0.157\textwidth]{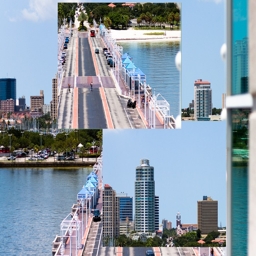} &
\vspace{1mm} \includegraphics[width=0.157\textwidth]{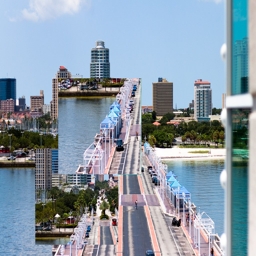}\\
\hline

\centering \adjustbox{angle=90}{ \texttt{Cartoonize}}  &  Transforms an image into a stylized, simplified version reminiscent of a cartoon.  & 
\vspace{1mm} \includegraphics[width=0.157\textwidth]{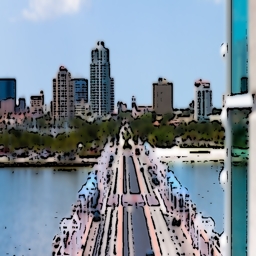} & 
\vspace{1mm} \includegraphics[width=0.157\textwidth]{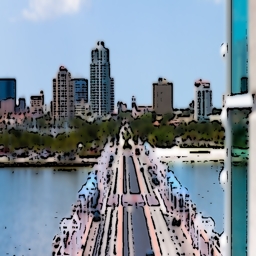} &
\vspace{1mm} \includegraphics[width=0.157\textwidth]{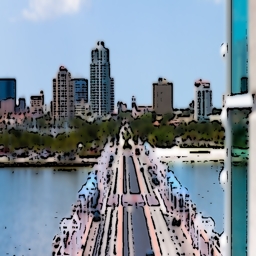} &
\vspace{1mm} \includegraphics[width=0.157\textwidth]{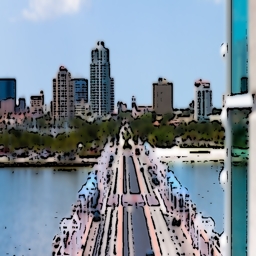}\\
\hline

\centering \adjustbox{angle=90}{ \texttt{FDA}}  &  Apply a random Fourier Domain Adaptation \textcolor{blue}{\citep{yang2020fda}} to an image.  & 
\vspace{1mm} \includegraphics[width=0.157\textwidth]{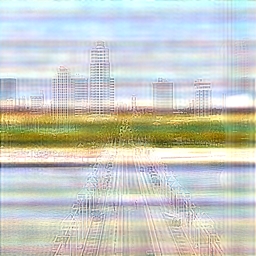} & 
\vspace{1mm} \includegraphics[width=0.157\textwidth]{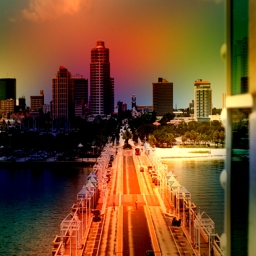} &
\vspace{1mm} \includegraphics[width=0.157\textwidth]{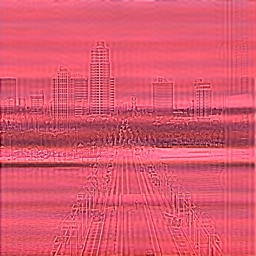} &
\vspace{1mm} \includegraphics[width=0.157\textwidth]{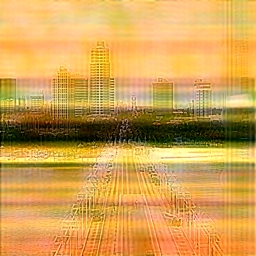}\\
\hline

\centering \adjustbox{angle=90}{ \texttt{\textcolor{blue}{OilPaint}}}  &  Randomly make an image look like an oil painting.  & 
\vspace{1mm} \includegraphics[width=0.157\textwidth]{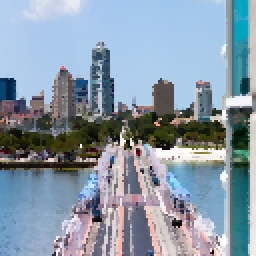} & 
\vspace{1mm} \includegraphics[width=0.157\textwidth]{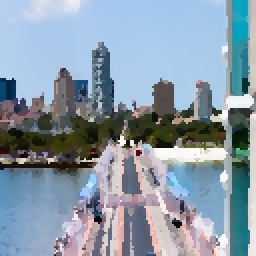} &
\vspace{1mm} \includegraphics[width=0.157\textwidth]{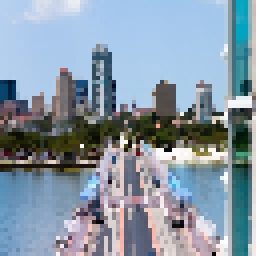} &
\vspace{1mm} \includegraphics[width=0.157\textwidth]{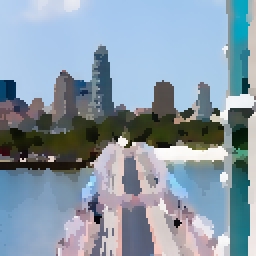}\\
\hline

\centering \adjustbox{angle=90}{ \texttt{Jigsaw}}  &  Randomly make an image look like an Jigsaw.  & 
\vspace{1mm} \includegraphics[width=0.157\textwidth]{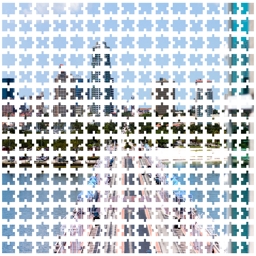} & 
\vspace{1mm} \includegraphics[width=0.157\textwidth]{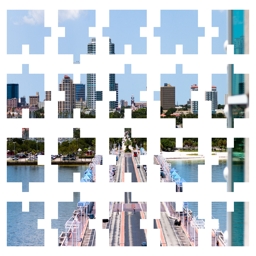} &
\vspace{1mm} \includegraphics[width=0.157\textwidth]{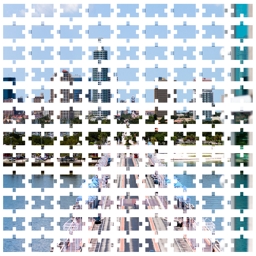} &
\vspace{1mm} \includegraphics[width=0.157\textwidth]{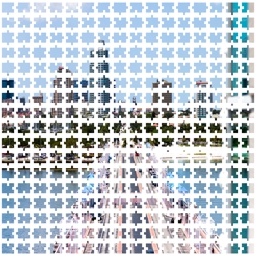}\\
\hline

\end{tabularx}}
\end{table*}

\begin{table*}[t]
\centering
\resizebox{0.97\textwidth}{!}{
\normalsize
  \begin{tabularx}{\textwidth}{|m{0.9cm}||m{2.5cm}|m{2.8cm}|m{2.8cm}|m{2.8cm}|m{2.8cm}|}
  \hline\thickhline
    
   \rowcolor{mygray}
  \multicolumn{1}{|c||}{Pattern} & \multicolumn{1}{c|}{Elaboration} & \multicolumn{1}{c|}{Demo 1} & \multicolumn{1}{c|}{Demo 2} & \multicolumn{1}{c|}{Demo 3} & \multicolumn{1}{c|}{Demo 4} \\
\hline\hline

\centering \adjustbox{angle=90}{ \texttt{FlRoCollage}}  &  A random combination of the flipped and rotated version of an image.  & 
\vspace{1mm} \includegraphics[width=0.157\textwidth]{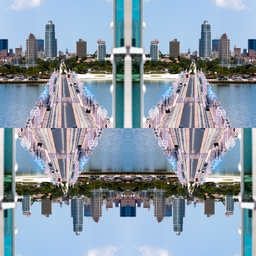} & 
\vspace{1mm} \includegraphics[width=0.157\textwidth]{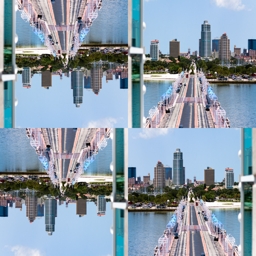} &
\vspace{1mm} \includegraphics[width=0.157\textwidth]{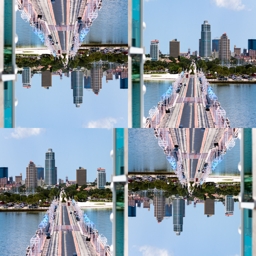} &
\vspace{1mm} \includegraphics[width=0.157\textwidth]{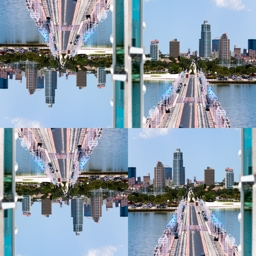}\\
\hline

\centering \adjustbox{angle=90}{ \texttt{Elastic}}  &  Simulate a jelly-like distortion of an image.  & 
\vspace{1mm} \includegraphics[width=0.157\textwidth]{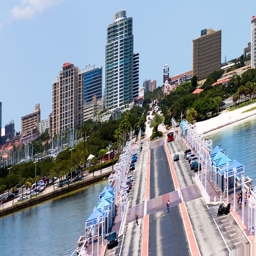} & 
\vspace{1mm} \includegraphics[width=0.157\textwidth]{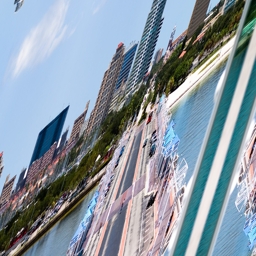} &
\vspace{1mm} \includegraphics[width=0.157\textwidth]{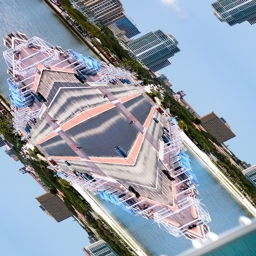} &
\vspace{1mm} \includegraphics[width=0.157\textwidth]{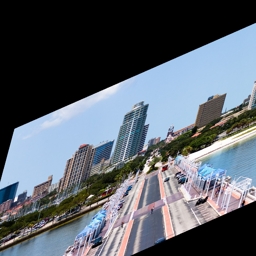}\\
\hline

\centering \adjustbox{angle=90}{ \texttt{PatchInter}}  &  Dividing two random images into patches and then interweave them.  & 
\vspace{1mm} \includegraphics[width=0.157\textwidth]{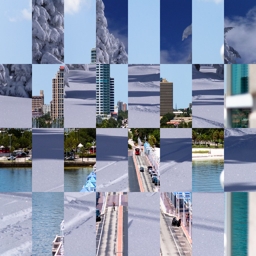} & 
\vspace{1mm} \includegraphics[width=0.157\textwidth]{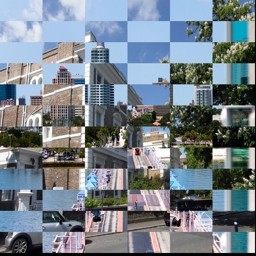} &
\vspace{1mm} \includegraphics[width=0.157\textwidth]{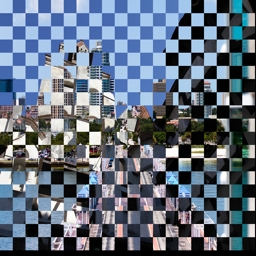} &
\vspace{1mm} \includegraphics[width=0.157\textwidth]{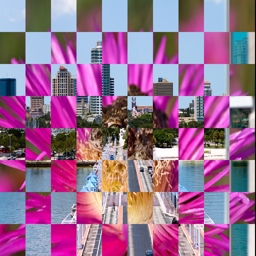}\\
\hline

\centering \adjustbox{angle=90}{ \texttt{FancyPCA}}  &  Random use PCA \textcolor{blue}{\citep{krizhevsky2012imagenet}} to alter the intensities of the RGB channels in an image.  & 
\vspace{1mm} \includegraphics[width=0.157\textwidth]{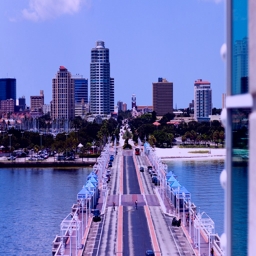} & 
\vspace{1mm} \includegraphics[width=0.157\textwidth]{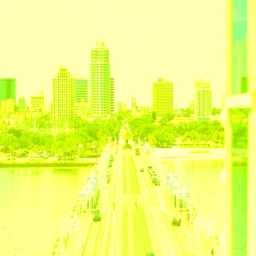} &
\vspace{1mm} \includegraphics[width=0.157\textwidth]{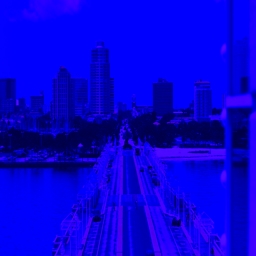} &
\vspace{1mm} \includegraphics[width=0.157\textwidth]{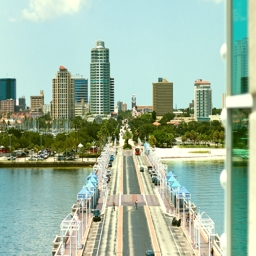}\\
\hline

\centering \adjustbox{angle=90}{ \texttt{GridDistort}}  &   Apply random non-linear distortions within a grid to an image.  & 
\vspace{1mm} \includegraphics[width=0.157\textwidth]{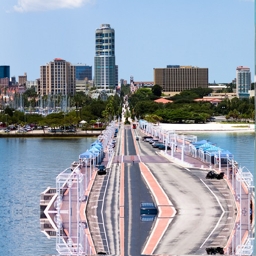} & 
\vspace{1mm} \includegraphics[width=0.157\textwidth]{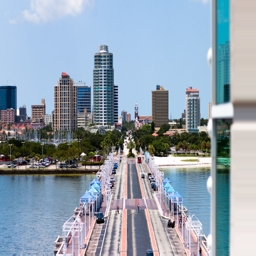} &
\vspace{1mm} \includegraphics[width=0.157\textwidth]{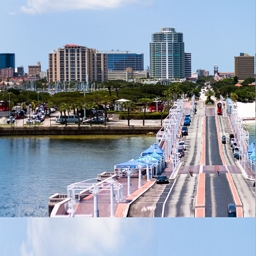} &
\vspace{1mm} \includegraphics[width=0.157\textwidth]{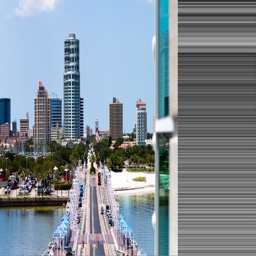}\\
\hline

\centering \adjustbox{angle=90}{ \texttt{HistMatch}}  &   Transform an image so that its histogram matches a random histogram.  & 
\vspace{1mm} \includegraphics[width=0.157\textwidth]{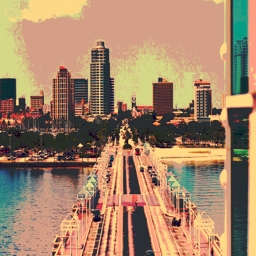} & 
\vspace{1mm} \includegraphics[width=0.157\textwidth]{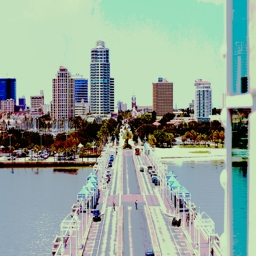} &
\vspace{1mm} \includegraphics[width=0.157\textwidth]{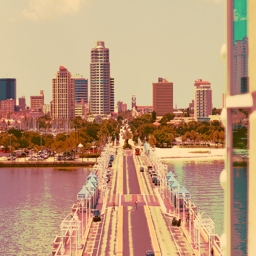} &
\vspace{1mm} \includegraphics[width=0.157\textwidth]{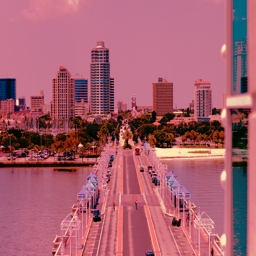}\\
\hline

\centering \adjustbox{angle=90}{ \texttt{ISONoise}}  &   Apply random camera sensor noise to an image.  & 
\vspace{1mm} \includegraphics[width=0.157\textwidth]{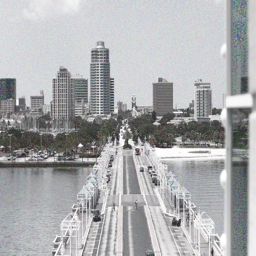} & 
\vspace{1mm} \includegraphics[width=0.157\textwidth]{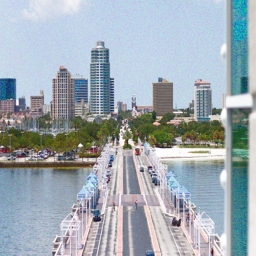} &
\vspace{1mm} \includegraphics[width=0.157\textwidth]{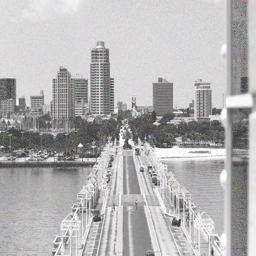} &
\vspace{1mm} \includegraphics[width=0.157\textwidth]{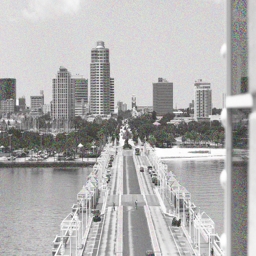}\\
\hline

\centering \adjustbox{angle=90}{ \texttt{Multiple}}  &   A random value, drawn from a distribution, is multiplied with the pixel values of an image.  & 
\vspace{1mm} \includegraphics[width=0.157\textwidth]{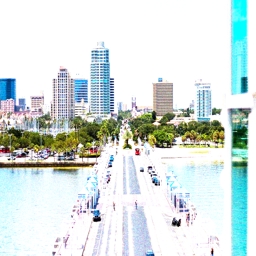} & 
\vspace{1mm} \includegraphics[width=0.157\textwidth]{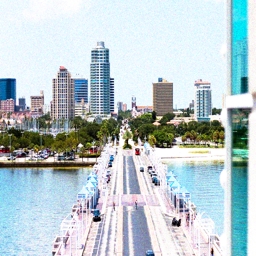} &
\vspace{1mm} \includegraphics[width=0.157\textwidth]{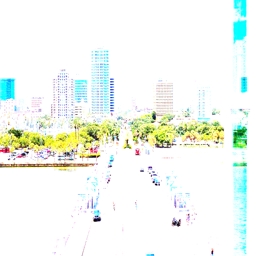} &
\vspace{1mm} \includegraphics[width=0.157\textwidth]{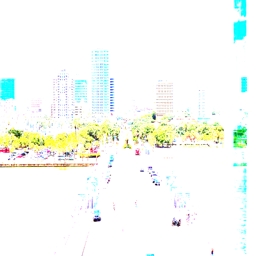}\\
\hline

\end{tabularx}}
\end{table*}

\begin{table*}[t]
\centering
\resizebox{0.97\textwidth}{!}{
\normalsize
  \begin{tabularx}{\textwidth}{|m{0.9cm}||m{2.5cm}|m{2.8cm}|m{2.8cm}|m{2.8cm}|m{2.8cm}|}
  \hline\thickhline
    
   \rowcolor{mygray}
  \multicolumn{1}{|c||}{Pattern} & \multicolumn{1}{c|}{Elaboration} & \multicolumn{1}{c|}{Demo 1} & \multicolumn{1}{c|}{Demo 2} & \multicolumn{1}{c|}{Demo 3} & \multicolumn{1}{c|}{Demo 4} \\
\hline\hline
\centering \adjustbox{angle=90}{ \texttt{Posterize}}  &   Randomly reduce the number of bits used to represent the color components of each pixel.  & 
\vspace{1mm} \includegraphics[width=0.157\textwidth]{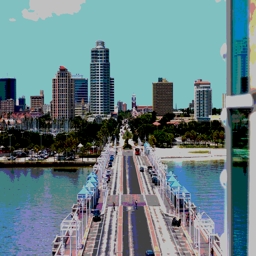} & 
\vspace{1mm} \includegraphics[width=0.157\textwidth]{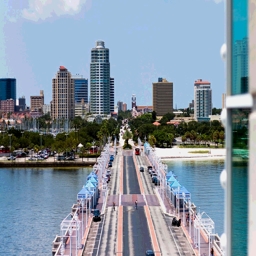} &
\vspace{1mm} \includegraphics[width=0.157\textwidth]{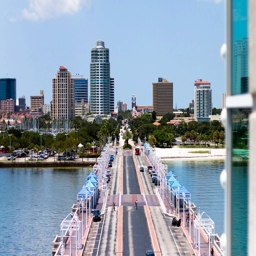} &
\vspace{1mm} \includegraphics[width=0.157\textwidth]{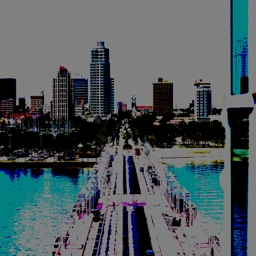}\\
\hline

\centering \adjustbox{angle=90}{ \texttt{Gamma}}  &   Alter the luminance values of an image by applying a power-law function.  & 
\vspace{1mm} \includegraphics[width=0.157\textwidth]{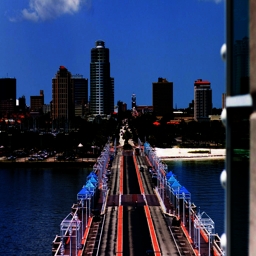} & 
\vspace{1mm} \includegraphics[width=0.157\textwidth]{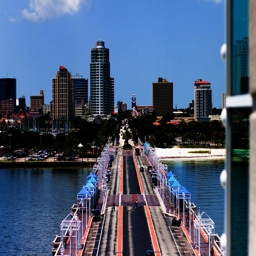} &
\vspace{1mm} \includegraphics[width=0.157\textwidth]{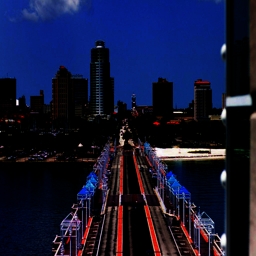} &
\vspace{1mm} \includegraphics[width=0.157\textwidth]{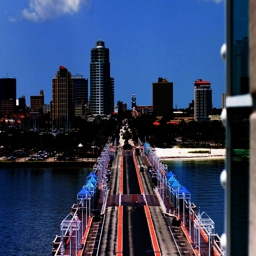}\\
\hline

\centering \adjustbox{angle=90}{ \texttt{Shadowy}}  &   Randomly add shadows onto an image.  & 
\vspace{1mm} \includegraphics[width=0.157\textwidth]{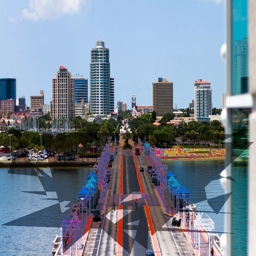} & 
\vspace{1mm} \includegraphics[width=0.157\textwidth]{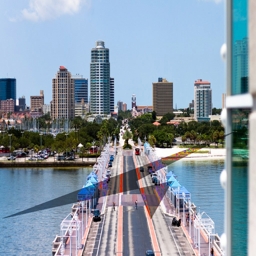} &
\vspace{1mm} \includegraphics[width=0.157\textwidth]{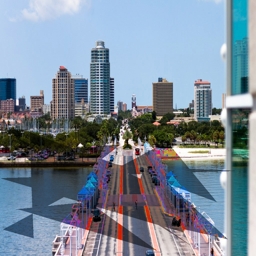} &
\vspace{1mm} \includegraphics[width=0.157\textwidth]{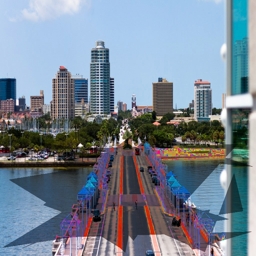}\\
\hline

\centering \adjustbox{angle=90}{ \texttt{Gravel}}  &   Randomly add gravels onto an image.  & 
\vspace{1mm} \includegraphics[width=0.157\textwidth]{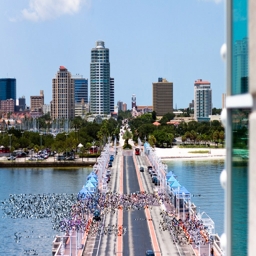} & 
\vspace{1mm} \includegraphics[width=0.157\textwidth]{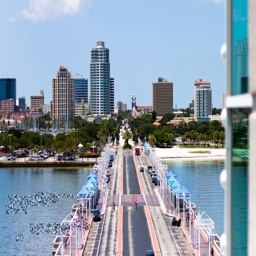} &
\vspace{1mm} \includegraphics[width=0.157\textwidth]{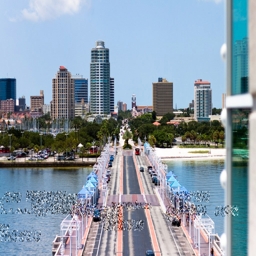} &
\vspace{1mm} \includegraphics[width=0.157\textwidth]{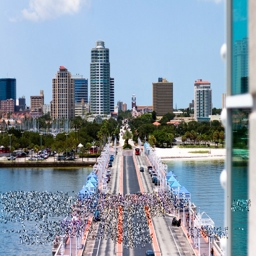}\\
\hline

\centering \adjustbox{angle=90}{ \texttt{SunFlare}}  &   Randomly add sun flares onto an image.  & 
\vspace{1mm} \includegraphics[width=0.157\textwidth]{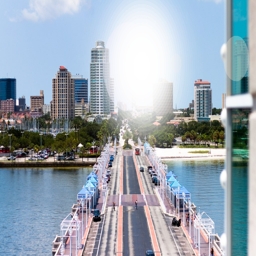} & 
\vspace{1mm} \includegraphics[width=0.157\textwidth]{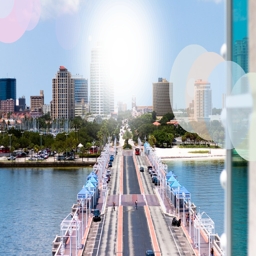} &
\vspace{1mm} \includegraphics[width=0.157\textwidth]{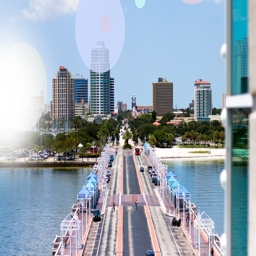} &
\vspace{1mm} \includegraphics[width=0.157\textwidth]{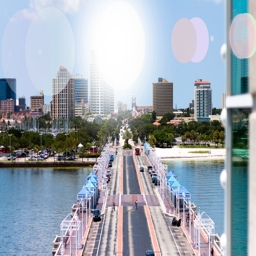}\\
\hline

\centering \adjustbox{angle=90}{ \texttt{SpeedUp}}  &   Randomly add speeding up effect onto an image.  & 
\vspace{1mm} \includegraphics[width=0.157\textwidth]{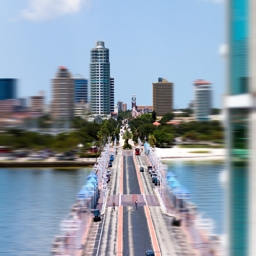} & 
\vspace{1mm} \includegraphics[width=0.157\textwidth]{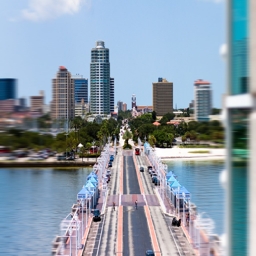} &
\vspace{1mm} \includegraphics[width=0.157\textwidth]{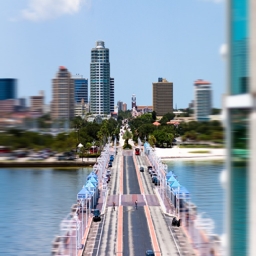} &
\vspace{1mm} \includegraphics[width=0.157\textwidth]{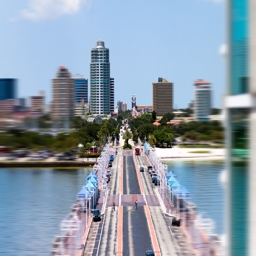}\\
\hline

\centering \adjustbox{angle=90}{ \texttt{Season}}  &   Randomly change the season of an image to spring, summer, autumn, or winter.  & 
\vspace{1mm} \includegraphics[width=0.157\textwidth]{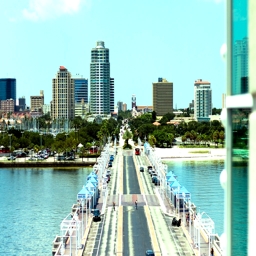} & 
\vspace{1mm} \includegraphics[width=0.157\textwidth]{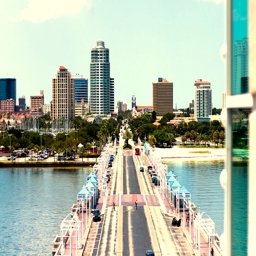} &
\vspace{1mm} \includegraphics[width=0.157\textwidth]{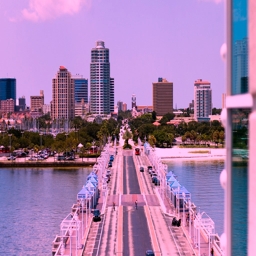} &
\vspace{1mm} \includegraphics[width=0.157\textwidth]{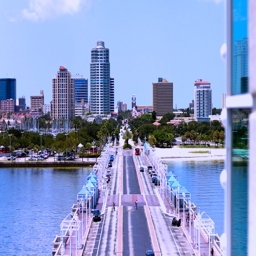}\\
\hline

\centering \adjustbox{angle=90}{ \texttt{CorrectExpo}}  &   Correct the exposure of an image.  & 
\vspace{1mm} \includegraphics[width=0.157\textwidth]{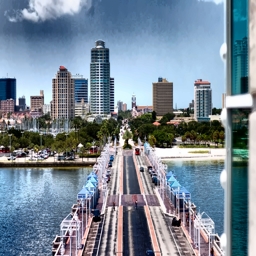} & 
\vspace{1mm} \includegraphics[width=0.157\textwidth]{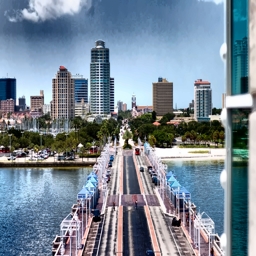} &
\vspace{1mm} \includegraphics[width=0.157\textwidth]{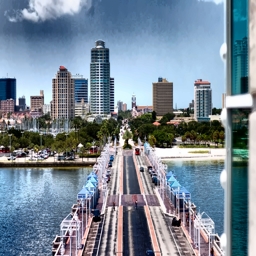} &
\vspace{1mm} \includegraphics[width=0.157\textwidth]{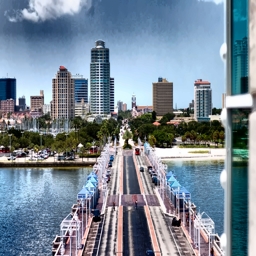}\\
\hline

\end{tabularx}}
\end{table*}

\begin{table*}[t]
\centering
\resizebox{0.97\textwidth}{!}{
\normalsize
 \begin{tabularx}{\textwidth}{|m{0.9cm}||m{2.5cm}|m{2.8cm}|m{2.8cm}|m{2.8cm}|m{2.8cm}|}
  \hline\thickhline
    
   \rowcolor{mygray}
  \multicolumn{1}{|c||}{Pattern} & \multicolumn{1}{c|}{Elaboration} & \multicolumn{1}{c|}{Demo 1} & \multicolumn{1}{c|}{Demo 2} & \multicolumn{1}{c|}{Demo 3} & \multicolumn{1}{c|}{Demo 4} \\
\hline\hline
\centering \adjustbox{angle=90}{ \textcolor{blue}{\texttt{Mosaic}}}  &   Randomly add mosaic effect on an image.  & 
\vspace{1mm} \includegraphics[width=0.157\textwidth]{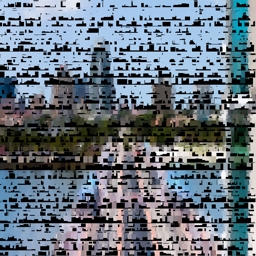} & 
\vspace{1mm} \includegraphics[width=0.157\textwidth]{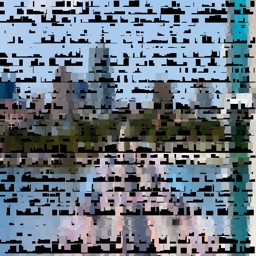} &
\vspace{1mm} \includegraphics[width=0.157\textwidth]{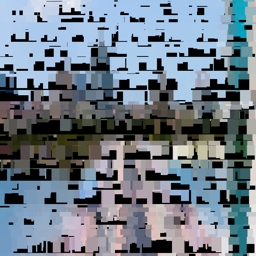} &
\vspace{1mm} \includegraphics[width=0.157\textwidth]{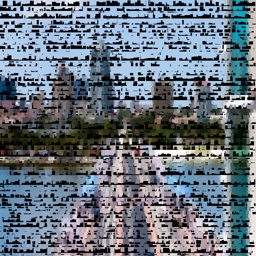}\\
\hline

\centering \adjustbox{angle=90}{ \texttt{PartsCollage}}  &   Randomly collage different parts of an image.  & 
\vspace{1mm} \includegraphics[width=0.157\textwidth]{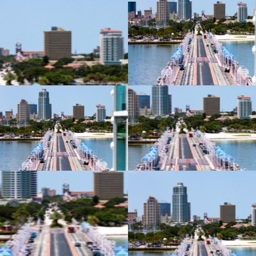} & 
\vspace{1mm} \includegraphics[width=0.157\textwidth]{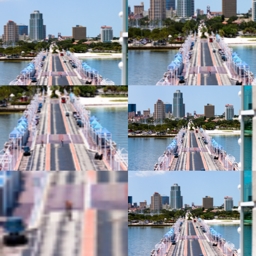} &
\vspace{1mm} \includegraphics[width=0.157\textwidth]{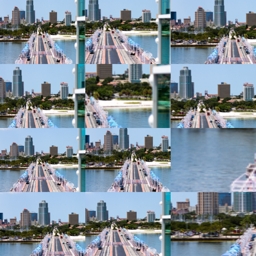} &
\vspace{1mm} \includegraphics[width=0.157\textwidth]{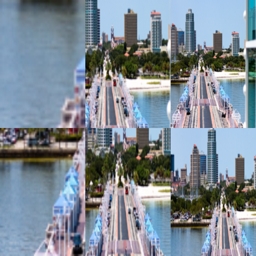}\\
\hline

\centering \adjustbox{angle=90}{ \texttt{RandCollage}}  &   Randomly collage $N \times M$ images.  & 
\vspace{1mm} \includegraphics[width=0.157\textwidth]{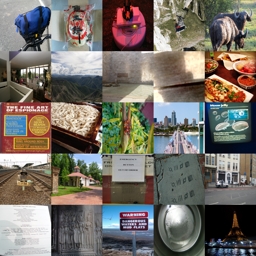} & 
\vspace{1mm} \includegraphics[width=0.157\textwidth]{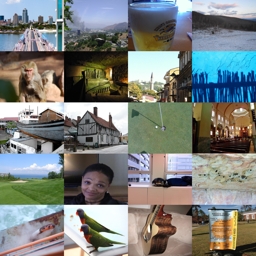} &
\vspace{1mm} \includegraphics[width=0.157\textwidth]{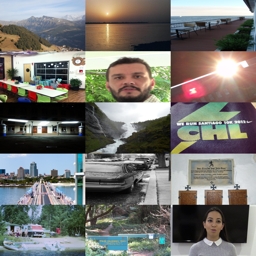} &
\vspace{1mm} \includegraphics[width=0.157\textwidth]{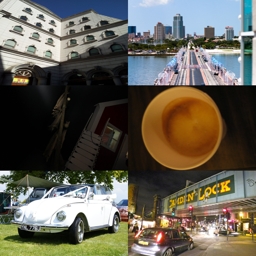}\\
\hline

\centering \adjustbox{angle=90}{ \texttt{Spatter}}  &   Randomly create a spatter effect on an image.  & 
\vspace{1mm} \includegraphics[width=0.157\textwidth]{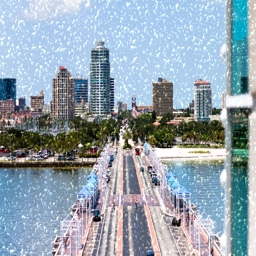} & 
\vspace{1mm} \includegraphics[width=0.157\textwidth]{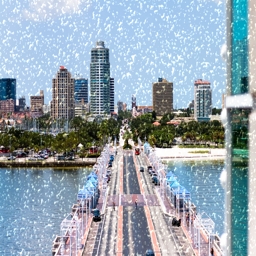} &
\vspace{1mm} \includegraphics[width=0.157\textwidth]{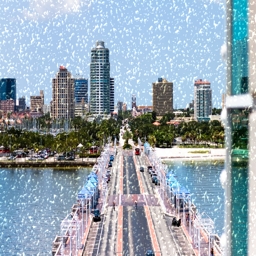} &
\vspace{1mm} \includegraphics[width=0.157\textwidth]{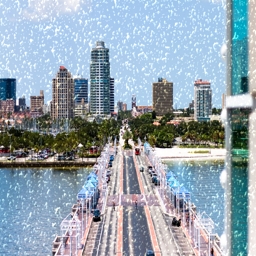}\\
\hline

\centering \adjustbox{angle=90}{ \texttt{SuperPixel}}  &   Random perceptual group pixels in an image \textcolor{blue}{\citep{neubert2014compact}}.  & 
\vspace{1mm} \includegraphics[width=0.157\textwidth]{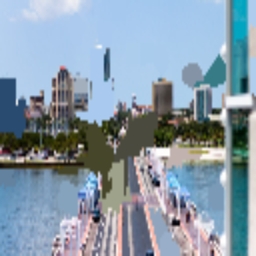} & 
\vspace{1mm} \includegraphics[width=0.157\textwidth]{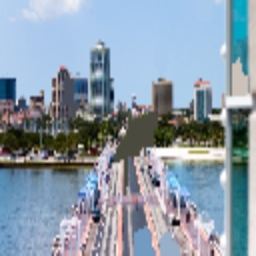} &
\vspace{1mm} \includegraphics[width=0.157\textwidth]{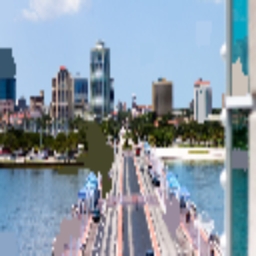} &
\vspace{1mm} \includegraphics[width=0.157\textwidth]{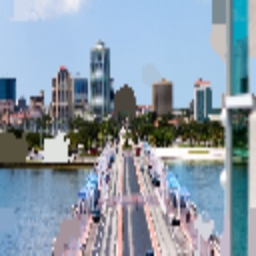}\\
\hline

\centering \adjustbox{angle=90}{ \texttt{Heatmap}}  &   Randomly apply a color map to an image.  & 
\vspace{1mm} \includegraphics[width=0.157\textwidth]{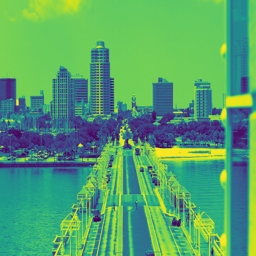} & 
\vspace{1mm} \includegraphics[width=0.157\textwidth]{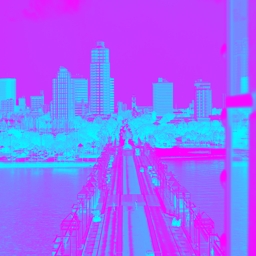} &
\vspace{1mm} \includegraphics[width=0.157\textwidth]{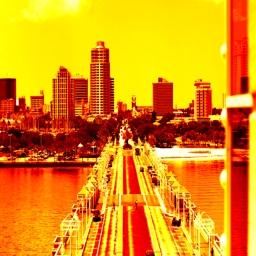} &
\vspace{1mm} \includegraphics[width=0.157\textwidth]{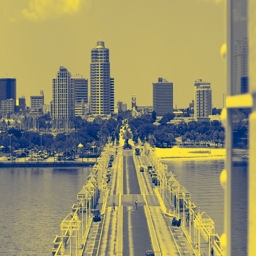}\\
\hline

\centering \adjustbox{angle=90}{ \texttt{PolarWarp}}  &   Randomly apply other patterns in a polar-transformed space.  & 
\vspace{1mm} \includegraphics[width=0.157\textwidth]{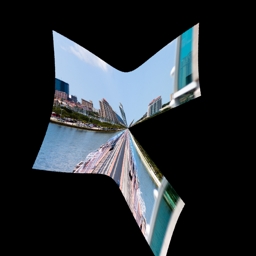} & 
\vspace{1mm} \includegraphics[width=0.157\textwidth]{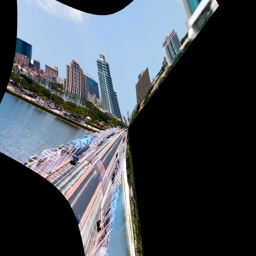} &
\vspace{1mm} \includegraphics[width=0.157\textwidth]{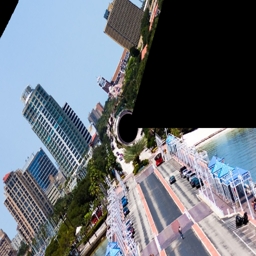} &
\vspace{1mm} \includegraphics[width=0.157\textwidth]{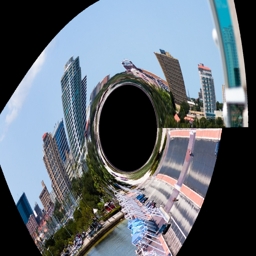}\\
\hline

\centering \adjustbox{angle=90}{ \texttt{DropChannel}}  &   Randomly drop a channel of an image.  & 
\vspace{1mm} \includegraphics[width=0.157\textwidth]{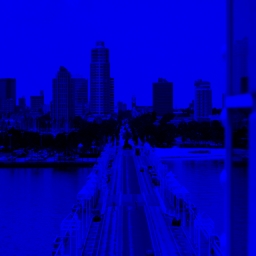} & 
\vspace{1mm} \includegraphics[width=0.157\textwidth]{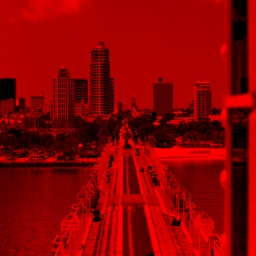} &
\vspace{1mm} \includegraphics[width=0.157\textwidth]{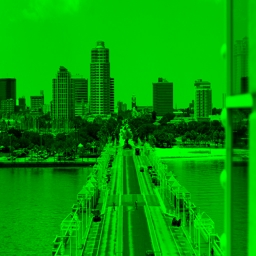} &
\vspace{1mm} \includegraphics[width=0.157\textwidth]{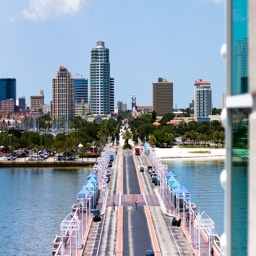}\\
\hline
\end{tabularx}}
\end{table*}

\begin{table*}[t]
\centering
\resizebox{0.97\textwidth}{!}{
\normalsize
  \begin{tabularx}{\textwidth}{|m{0.9cm}||m{2.5cm}|m{2.8cm}|m{2.8cm}|m{2.8cm}|m{2.8cm}|}
  \hline\thickhline
    
   \rowcolor{mygray}
  \multicolumn{1}{|c||}{Pattern} & \multicolumn{1}{c|}{Elaboration} & \multicolumn{1}{c|}{Demo 1} & \multicolumn{1}{c|}{Demo 2} & \multicolumn{1}{c|}{Demo 3} & \multicolumn{1}{c|}{Demo 4} \\
\hline\hline
\centering \adjustbox{angle=90}{ \texttt{ColorQuant}}  &   Randomly reduce the number of distinct colors used in an image.  & 
\vspace{1mm} \includegraphics[width=0.157\textwidth]{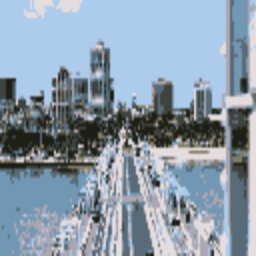} & 
\vspace{1mm} \includegraphics[width=0.157\textwidth]{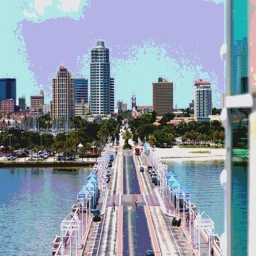} &
\vspace{1mm} \includegraphics[width=0.157\textwidth]{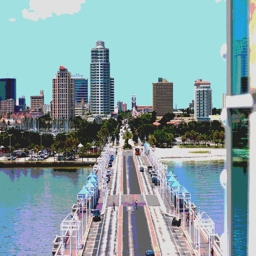} &
\vspace{1mm} \includegraphics[width=0.157\textwidth]{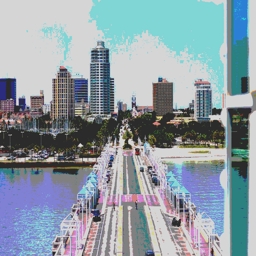}\\
\hline

\centering \adjustbox{angle=90}{ \texttt{Emboss}}  &   Replace each pixel of an image with a highlight or shadow to give a three-dimensional impression.  & 
\vspace{1mm} \includegraphics[width=0.157\textwidth]{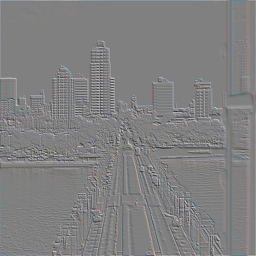} & 
\vspace{1mm} \includegraphics[width=0.157\textwidth]{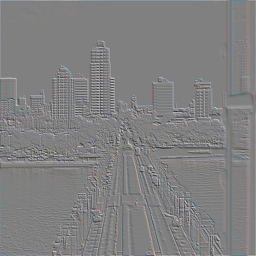} &
\vspace{1mm} \includegraphics[width=0.157\textwidth]{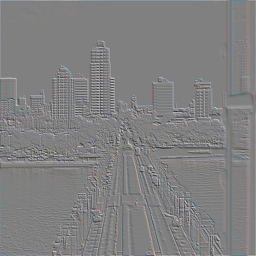} &
\vspace{1mm} \includegraphics[width=0.157\textwidth]{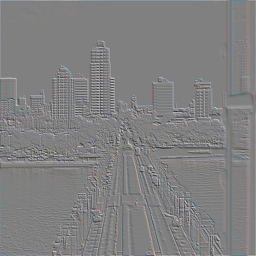}\\
\hline

\centering \adjustbox{angle=90}{ \textcolor{blue}{\texttt{Voronoi}}}&   Randomly segment an image based on the `closeness' of pixels to certain features or points.  & 
\vspace{1mm} \includegraphics[width=0.157\textwidth]{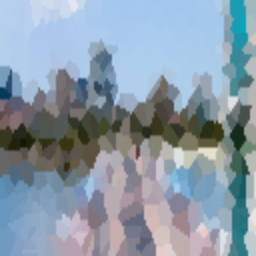} & 
\vspace{1mm} \includegraphics[width=0.157\textwidth]{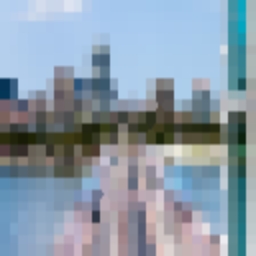} &
\vspace{1mm} \includegraphics[width=0.157\textwidth]{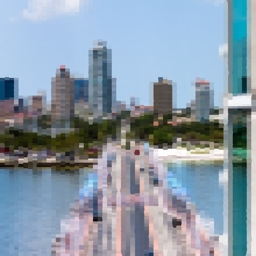} &
\vspace{1mm} \includegraphics[width=0.157\textwidth]{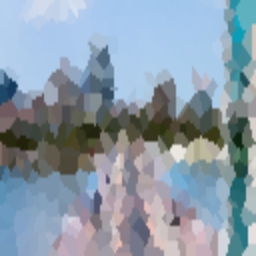}\\
\hline

\centering \adjustbox{angle=90}{ \texttt{ColorSpace}}  &   Randomly change an image with RGB space to HSV, HLS, LAB, YCrCb, LUV, XYZ, or space.  & 
\vspace{1mm} \includegraphics[width=0.157\textwidth]{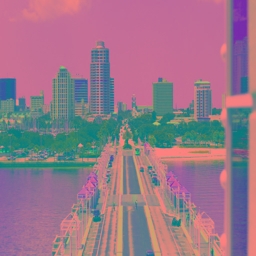} & 
\vspace{1mm} \includegraphics[width=0.157\textwidth]{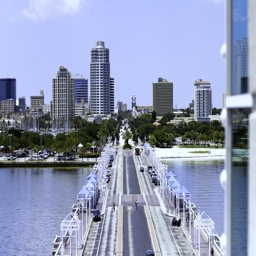} &
\vspace{1mm} \includegraphics[width=0.157\textwidth]{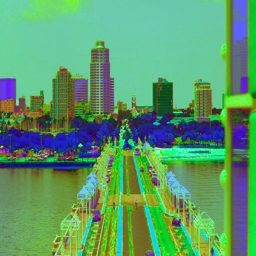} &
\vspace{1mm} \includegraphics[width=0.157\textwidth]{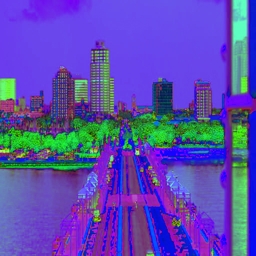}\\
\hline

\centering \adjustbox{angle=90}{ \texttt{OldSchool}}  &   Randomly change an image into old school style.  & 
\vspace{1mm} \includegraphics[width=0.157\textwidth]{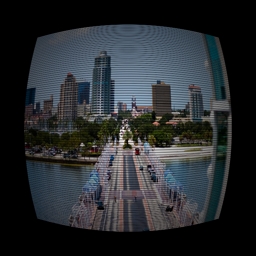} & 
\vspace{1mm} \includegraphics[width=0.157\textwidth]{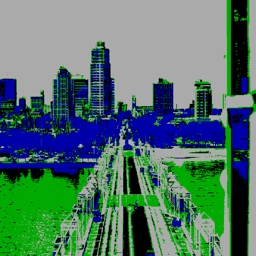} &
\vspace{1mm} \includegraphics[width=0.157\textwidth]{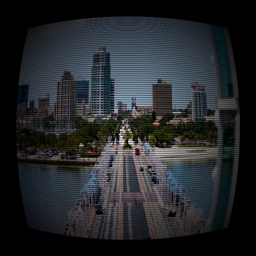} &
\vspace{1mm} \includegraphics[width=0.157\textwidth]{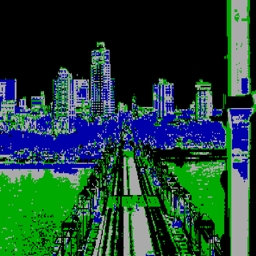}\\
\hline

\centering \adjustbox{angle=90}{ \texttt{Ascii}}  &   Randomly use colored ASCII texts to represent an image.  & 
\vspace{1mm} \includegraphics[width=0.157\textwidth]{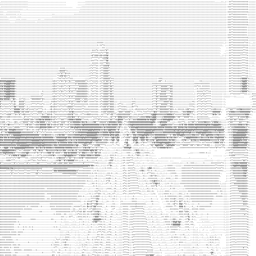} & 
\vspace{1mm} \includegraphics[width=0.157\textwidth]{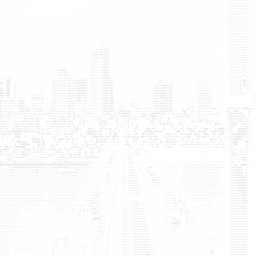} &
\vspace{1mm} \includegraphics[width=0.157\textwidth]{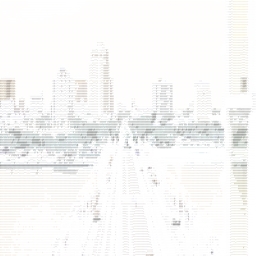} &
\vspace{1mm} \includegraphics[width=0.157\textwidth]{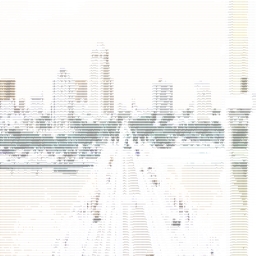}\\
\hline

\centering \adjustbox{angle=90}{ \texttt{PixelMelt}}  &   Randomly simulate the effect of "melting" pixels down the image.  & 
\vspace{1mm} \includegraphics[width=0.157\textwidth]{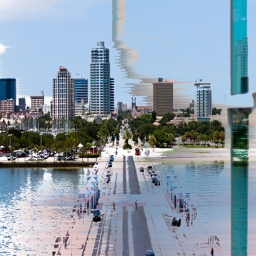} & 
\vspace{1mm} \includegraphics[width=0.157\textwidth]{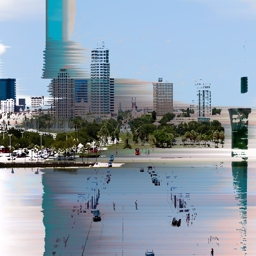} &
\vspace{1mm} \includegraphics[width=0.157\textwidth]{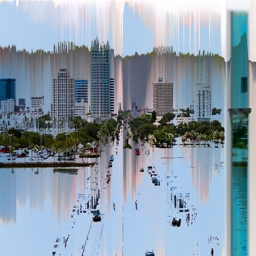} &
\vspace{1mm} \includegraphics[width=0.157\textwidth]{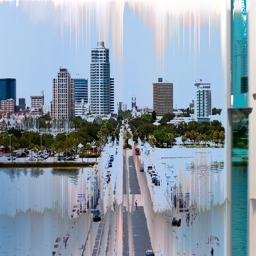}\\
\hline

\centering \adjustbox{angle=90}{ \textcolor{blue}{\texttt{Stylize}}} &   Randomly replace the style of an image with another one's \textcolor{blue}{\citep{zhang2017multistyle}}.  & 
\vspace{1mm} \includegraphics[width=0.157\textwidth]{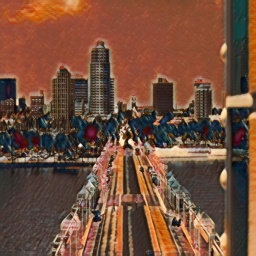} & 
\vspace{1mm} \includegraphics[width=0.157\textwidth]{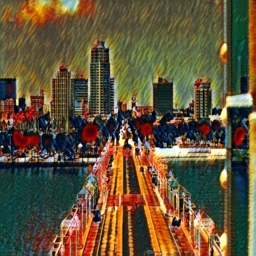} &
\vspace{1mm} \includegraphics[width=0.157\textwidth]{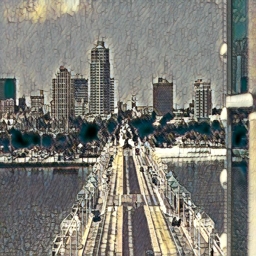} &
\vspace{1mm} \includegraphics[width=0.157\textwidth]{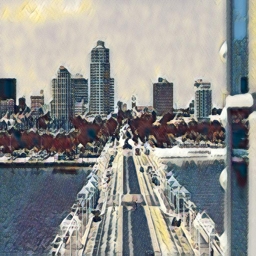}\\
\hline

\end{tabularx}}
\end{table*}

\begin{table*}[t]
\centering
\resizebox{0.97\textwidth}{!}{
\normalsize
  \begin{tabularx}{\textwidth}{|m{0.9cm}||m{2.5cm}|m{2.8cm}|m{2.8cm}|m{2.8cm}|m{2.8cm}|}
  \hline\thickhline
    
   \rowcolor{mygray}
  \multicolumn{1}{|c||}{Pattern} & \multicolumn{1}{c|}{Elaboration} & \multicolumn{1}{c|}{Demo 1} & \multicolumn{1}{c|}{Demo 2} & \multicolumn{1}{c|}{Demo 3} & \multicolumn{1}{c|}{Demo 4} \\
\hline\hline
\centering \adjustbox{angle=90}{ \texttt{Animation}}  &   Randomly transfer an image to animation style \textcolor{blue}{\citep{chen2020animegan}}.  & 
\vspace{1mm} \includegraphics[width=0.157\textwidth]{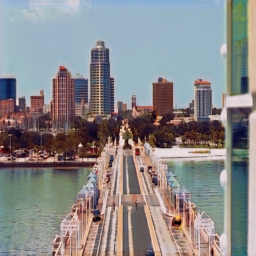} & 
\vspace{1mm} \includegraphics[width=0.157\textwidth]{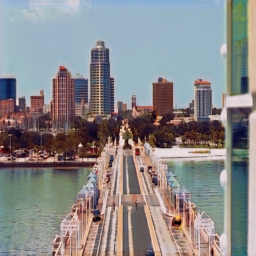} &
\vspace{1mm} \includegraphics[width=0.157\textwidth]{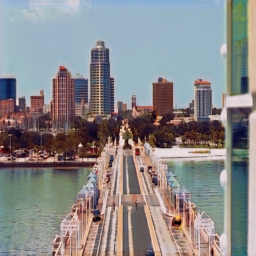} &
\vspace{1mm} \includegraphics[width=0.157\textwidth]{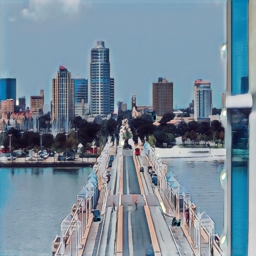}\\
\hline

\centering \adjustbox{angle=90}{ \textcolor{blue}{\texttt{AdvAttack}}}  &   Randomly perform adversarial attacks to an image \textcolor{blue}{\citep{kim2020torchattacks}}.  & 
\vspace{1mm} \includegraphics[width=0.157\textwidth]{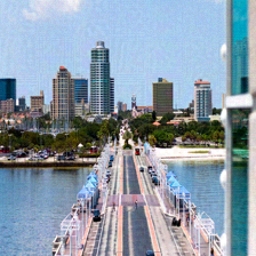} & 
\vspace{1mm} \includegraphics[width=0.157\textwidth]{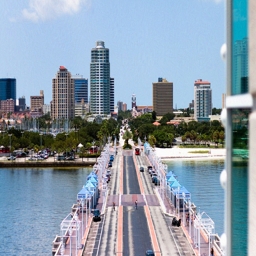} &
\vspace{1mm} \includegraphics[width=0.157\textwidth]{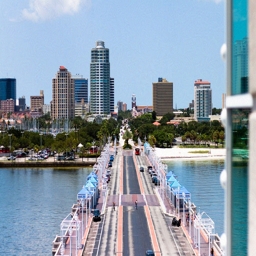} &
\vspace{1mm} \includegraphics[width=0.157\textwidth]{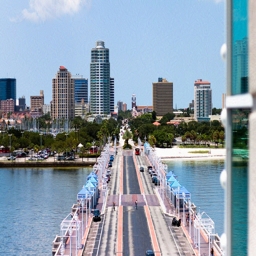}\\
\hline

\centering \adjustbox{angle=90}{ \texttt{CLAHE}}  &   Apply Contrast Limited Adaptive Histogram Equalization \textcolor{blue}{\citep{reza2004realization}} to equalize an image.  & 
\vspace{1mm} \includegraphics[width=0.157\textwidth]{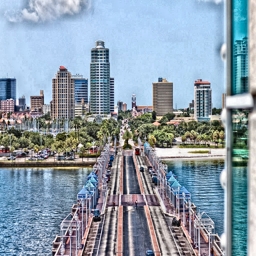} & 
\vspace{1mm} \includegraphics[width=0.157\textwidth]{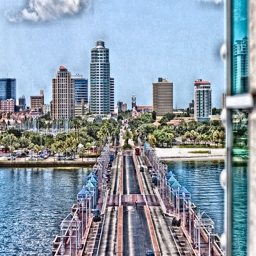} &
\vspace{1mm} \includegraphics[width=0.157\textwidth]{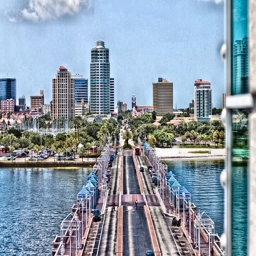} &
\vspace{1mm} \includegraphics[width=0.157\textwidth]{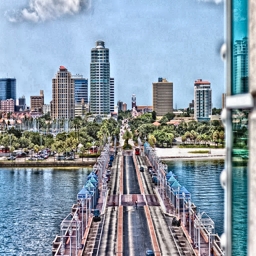}\\
\hline

\centering \adjustbox{angle=90}{ \texttt{Screenshot}}  &   Emboss an image into a random screen shot.  & 
\vspace{1mm} \includegraphics[width=0.157\textwidth]{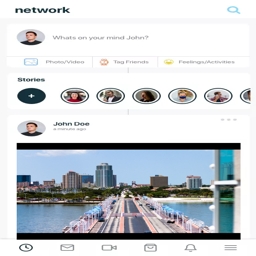} & 
\vspace{1mm} \includegraphics[width=0.157\textwidth]{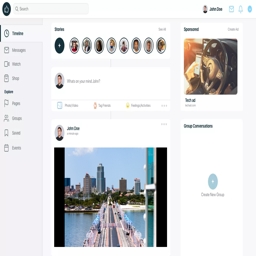} &
\vspace{1mm} \includegraphics[width=0.157\textwidth]{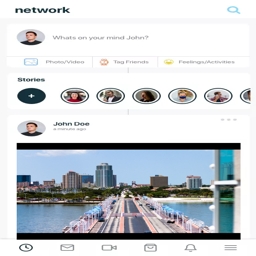} &
\vspace{1mm} \includegraphics[width=0.157\textwidth]{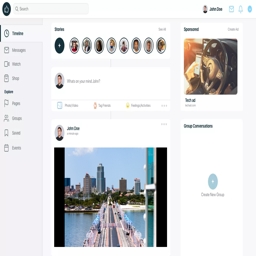}\\
\hline

\centering \adjustbox{angle=90}{ \texttt{DisLocation}}  &   Random the location of patches in an image.  & 
\vspace{1mm} \includegraphics[width=0.157\textwidth]{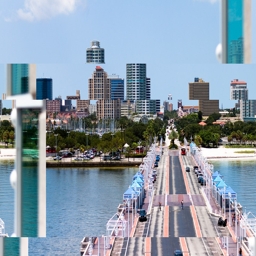} & 
\vspace{1mm} \includegraphics[width=0.157\textwidth]{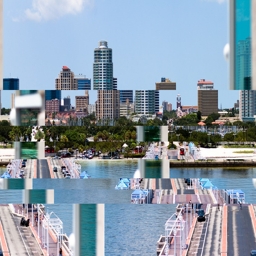} &
\vspace{1mm} \includegraphics[width=0.157\textwidth]{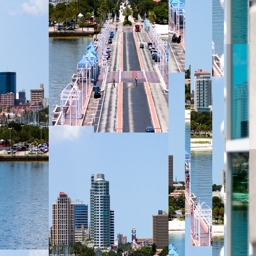} &
\vspace{1mm} \includegraphics[width=0.157\textwidth]{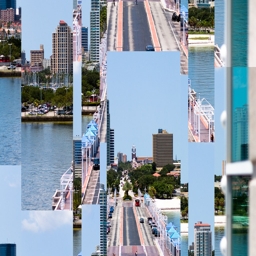}\\
\hline

\centering \adjustbox{angle=90}{ \texttt{DropArea}}  &   Apply DropOut \textcolor{blue}{\citep{srivastava2014dropout}} or DropBlock \textcolor{blue}{\citep{ghiasi2018dropblock}}.  & 
\vspace{1mm} \includegraphics[width=0.157\textwidth]{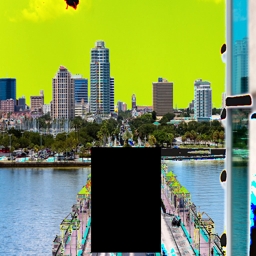} & 
\vspace{1mm} \includegraphics[width=0.157\textwidth]{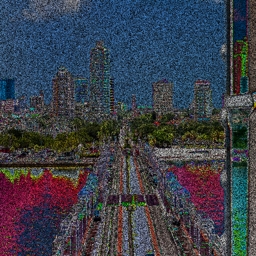} &
\vspace{1mm} \includegraphics[width=0.157\textwidth]{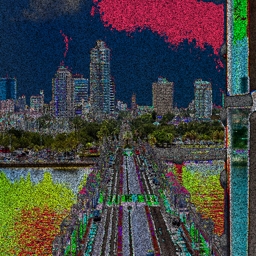} &
\vspace{1mm} \includegraphics[width=0.157\textwidth]{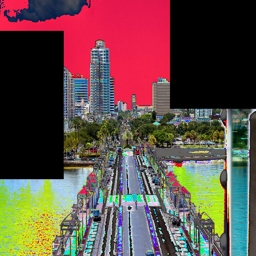}\\
\hline

\centering \adjustbox{angle=90}{ \textcolor{blue}{\texttt{WaveBlock}}}  &   Apply WaveBlock \textcolor{blue}{\citep{wang2022attentive}} on the image level.  & 
\vspace{1mm} \includegraphics[width=0.157\textwidth]{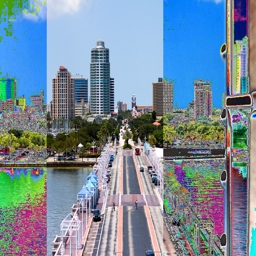} & 
\vspace{1mm} \includegraphics[width=0.157\textwidth]{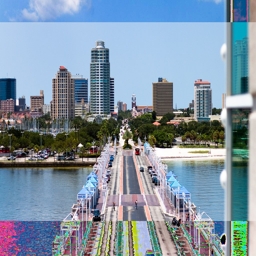} &
\vspace{1mm} \includegraphics[width=0.157\textwidth]{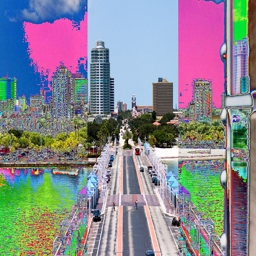} &
\vspace{1mm} \includegraphics[width=0.157\textwidth]{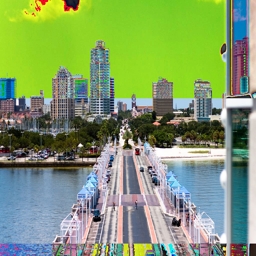}\\
\hline

\centering \adjustbox{angle=90}{ \texttt{Shape}}  &   Randomly crop an area with circle, heart, polygon, or star shape from an image.  & 
\vspace{1mm} \includegraphics[width=0.157\textwidth]{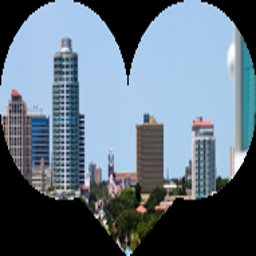} & 
\vspace{1mm} \includegraphics[width=0.157\textwidth]{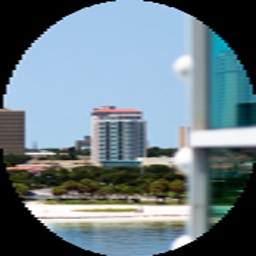} &
\vspace{1mm} \includegraphics[width=0.157\textwidth]{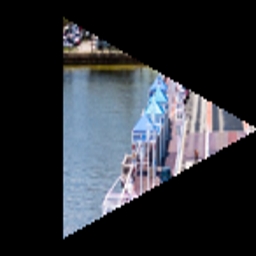} &
\vspace{1mm} \includegraphics[width=0.157\textwidth]{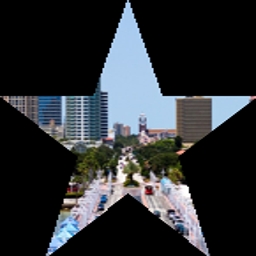}\\
\hline

\end{tabularx}}
\end{table*}

\begin{table*}[t]
\centering
\resizebox{0.97\textwidth}{!}{
\normalsize
 \begin{tabularx}{\textwidth}{|m{0.9cm}||m{2.5cm}|m{2.8cm}|m{2.8cm}|m{2.8cm}|m{2.8cm}|}
  \hline\thickhline
    
   \rowcolor{mygray}
  \multicolumn{1}{|c||}{Pattern} & \multicolumn{1}{c|}{Elaboration} & \multicolumn{1}{c|}{Demo 1} & \multicolumn{1}{c|}{Demo 2} & \multicolumn{1}{c|}{Demo 3} & \multicolumn{1}{c|}{Demo 4} \\
\hline\hline
\centering \adjustbox{angle=90}{ \texttt{VanGoghize}}  &   Change the style of an image to a painting from Vincent van Gogh randomly.  & 
\vspace{1mm} \includegraphics[width=0.16\textwidth]{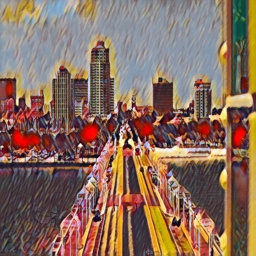} & 
\vspace{1mm} \includegraphics[width=0.157\textwidth]{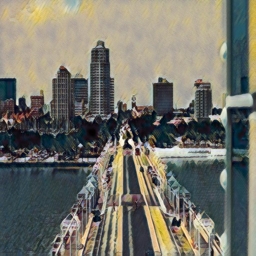} &
\vspace{1mm} \includegraphics[width=0.157\textwidth]{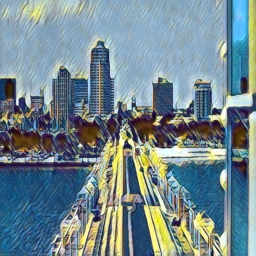} &
\vspace{1mm} \includegraphics[width=0.157\textwidth]{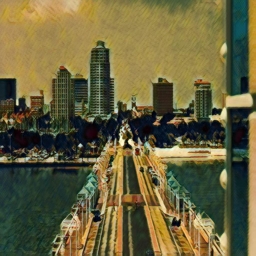}\\
\hline

\centering \adjustbox{angle=90}{ \texttt{Moire}}  &   Add random Moire patterns onto one image.  & 
\vspace{1mm} \includegraphics[width=0.157\textwidth]{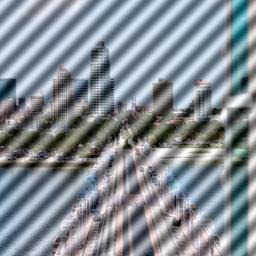} & 
\vspace{1mm} \includegraphics[width=0.157\textwidth]{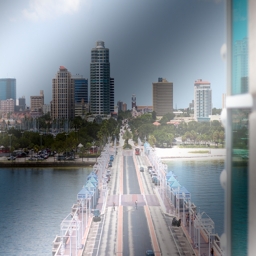} &
\vspace{1mm} \includegraphics[width=0.157\textwidth]{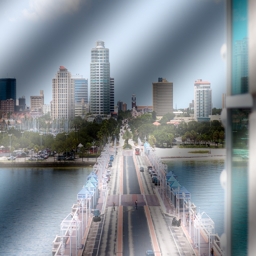} &
\vspace{1mm} \includegraphics[width=0.157\textwidth]{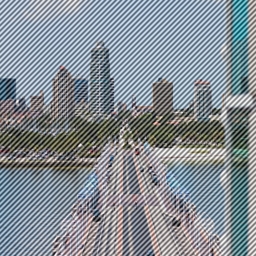}\\
\hline

\centering \adjustbox{angle=90}{ \texttt{Wavelet}}  &   Apply Wavelet \textcolor{blue}{\citep{torrence1998practical}} with random frequency to an image.  & 
\vspace{1mm} \includegraphics[width=0.157\textwidth]{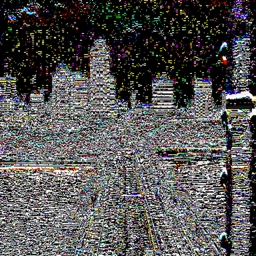} & 
\vspace{1mm} \includegraphics[width=0.157\textwidth]{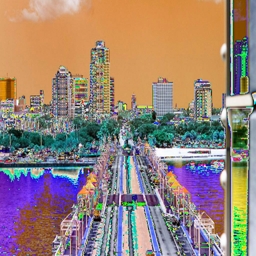} &
\vspace{1mm} \includegraphics[width=0.157\textwidth]{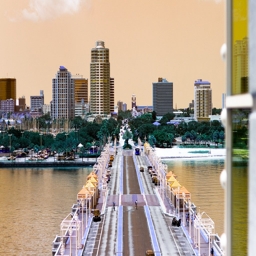} &
\vspace{1mm} \includegraphics[width=0.157\textwidth]{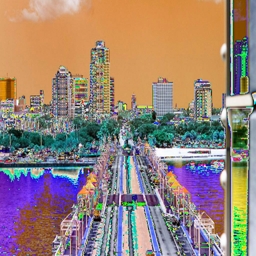}\\
\hline

\centering \adjustbox{angle=90}{ \texttt{Sepia}}  &   
Change an image to Sepia style.  & 
\vspace{1mm} \includegraphics[width=0.157\textwidth]{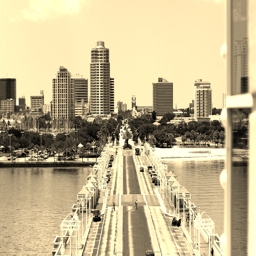} & 
\vspace{1mm} \includegraphics[width=0.157\textwidth]{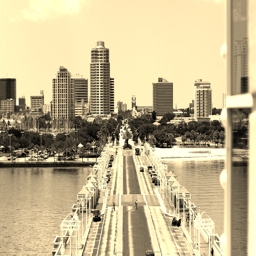} &
\vspace{1mm} \includegraphics[width=0.157\textwidth]{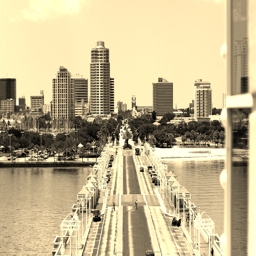} &
\vspace{1mm} \includegraphics[width=0.157\textwidth]{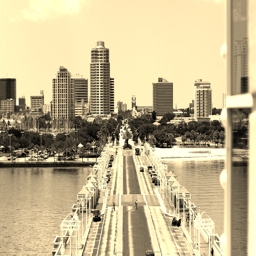}\\
\hline

\centering \adjustbox{angle=90}{ \texttt{ToneCurve}}  &   
Randomly change the bright and dark areas of the image by manipulating its tone curve.  & 
\vspace{1mm} \includegraphics[width=0.157\textwidth]{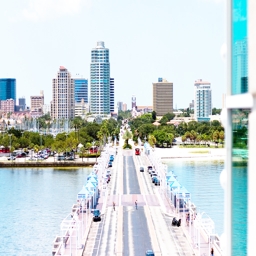} & 
\vspace{1mm} \includegraphics[width=0.157\textwidth]{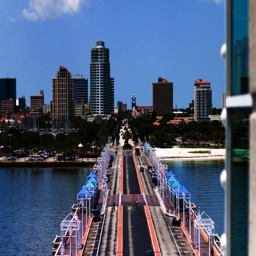} &
\vspace{1mm} \includegraphics[width=0.157\textwidth]{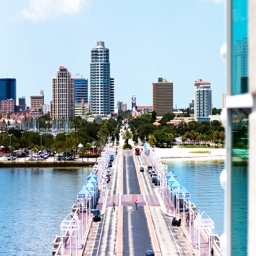} &
\vspace{1mm} \includegraphics[width=0.157\textwidth]{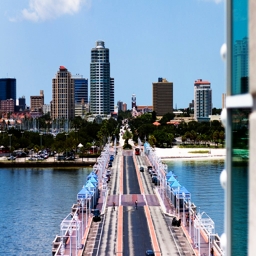}\\
\hline

\centering \adjustbox{angle=90}{ \texttt{Erasing}}  &   
Erase a random area of an image \textcolor{blue}{\citep{zhong2020random}}.  & 
\vspace{1mm} \includegraphics[width=0.157\textwidth]{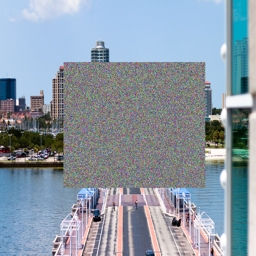} & 
\vspace{1mm} \includegraphics[width=0.157\textwidth]{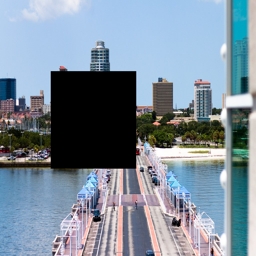} &
\vspace{1mm} \includegraphics[width=0.157\textwidth]{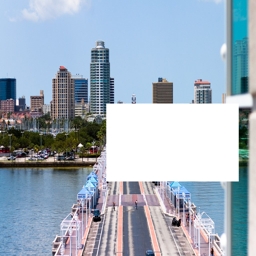} &
\vspace{1mm} \includegraphics[width=0.157\textwidth]{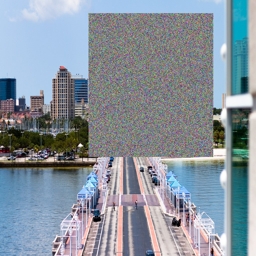}\\
\hline

\centering \adjustbox{angle=90}{ \textcolor{blue}{\texttt{AutoAug}}}  &   
Apply AutoAugment \textcolor{blue}{\citep{cubuk2018autoaugment}} to an image.  & 
\vspace{1mm} \includegraphics[width=0.157\textwidth]{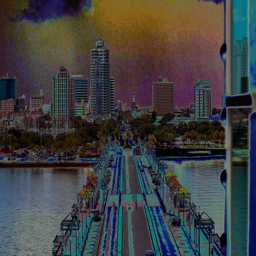} & 
\vspace{1mm} \includegraphics[width=0.157\textwidth]{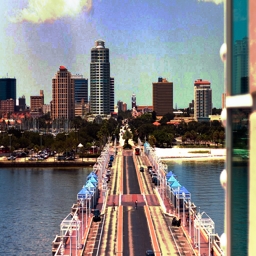} &
\vspace{1mm} \includegraphics[width=0.157\textwidth]{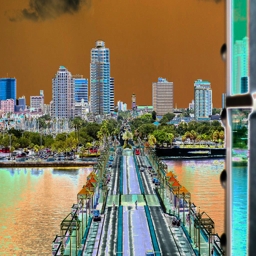} &
\vspace{1mm} \includegraphics[width=0.157\textwidth]{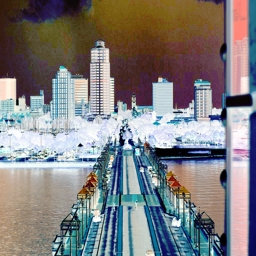}\\
\hline

\centering \adjustbox{angle=90}{ \textcolor{blue}{\texttt{AutoSeg}}}  &   
Use the Segment Anything \textcolor{blue}{\citep{kirillov2023segany}} model to preform segmentation.  & 
\vspace{1mm} \includegraphics[width=0.157\textwidth]{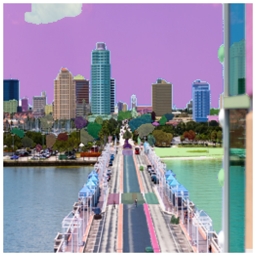} & 
\vspace{1mm} \includegraphics[width=0.157\textwidth]{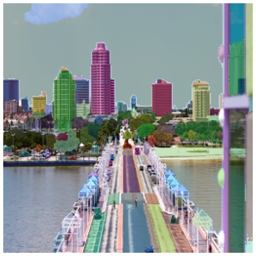} &
\vspace{1mm} \includegraphics[width=0.157\textwidth]{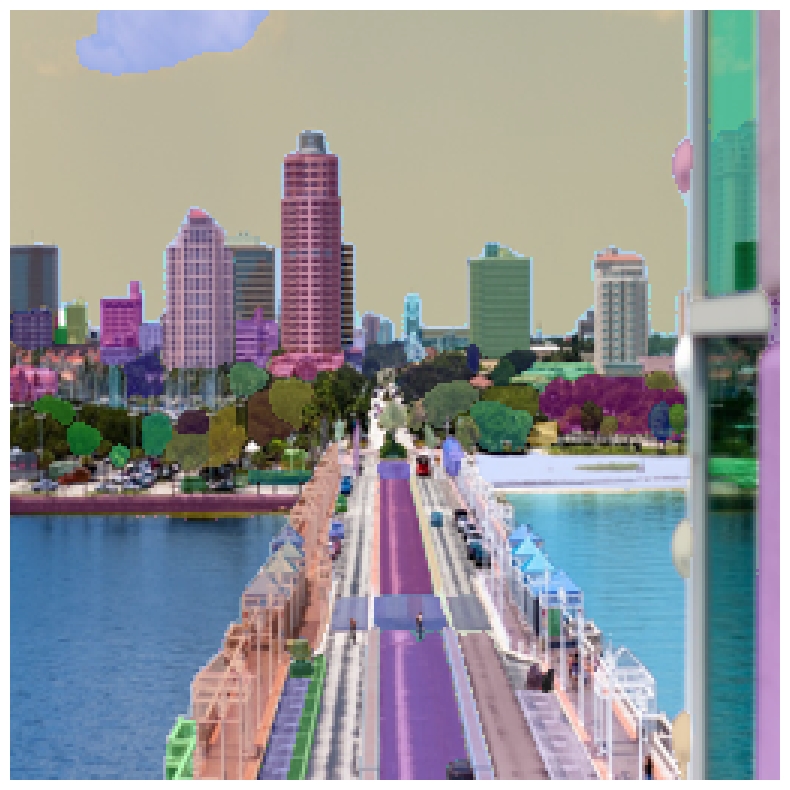} &
\vspace{1mm} \includegraphics[width=0.157\textwidth]{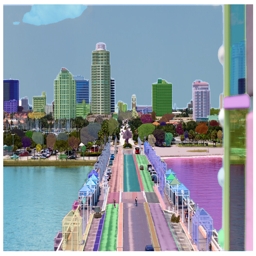}\\
\hline

\end{tabularx}}
\end{table*}

\begin{table*}[t]
\centering
\resizebox{0.97\textwidth}{!}{
\normalsize
 \begin{tabularx}{\textwidth}{|m{0.9cm}||m{2.5cm}|m{2.8cm}|m{2.8cm}|m{2.8cm}|m{2.8cm}|}
  \hline\thickhline
    
   \rowcolor{mygray}
  \multicolumn{1}{|c||}{Pattern} & \multicolumn{1}{c|}{Elaboration} & \multicolumn{1}{c|}{Demo 1} & \multicolumn{1}{c|}{Demo 2} & \multicolumn{1}{c|}{Demo 3} & \multicolumn{1}{c|}{Demo 4} \\
\hline\hline
\centering \adjustbox{angle=90}{ \texttt{FGRemove}}  &   Remove the foreground of an image.  & 
\vspace{1mm} \includegraphics[width=0.157\textwidth]{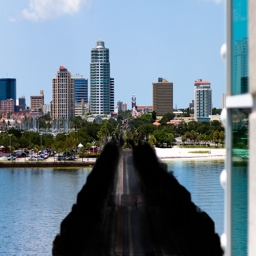} & 
\vspace{1mm} \includegraphics[width=0.157\textwidth]{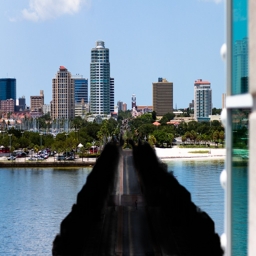} &
\vspace{1mm} \includegraphics[width=0.157\textwidth]{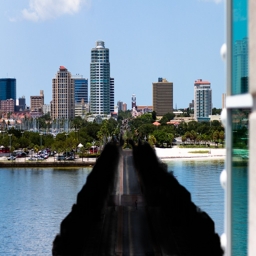} &
\vspace{1mm} \includegraphics[width=0.157\textwidth]{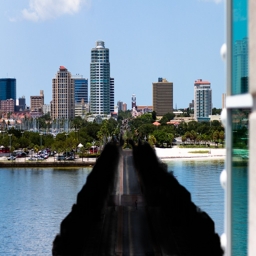}\\
\hline

\centering \adjustbox{angle=90}{ \texttt{ErodeDilate}}  &   Randomly add the erosion or dilation effect onto an image.  & 
\vspace{1mm} \includegraphics[width=0.157\textwidth]{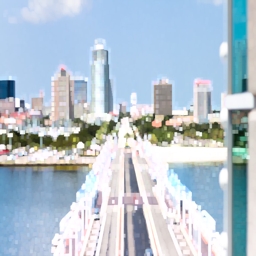} & 
\vspace{1mm} \includegraphics[width=0.157\textwidth]{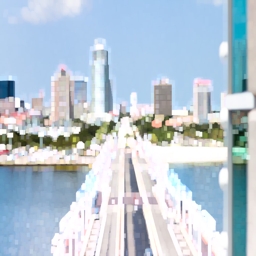} &
\vspace{1mm} \includegraphics[width=0.157\textwidth]{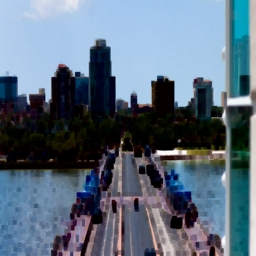} &
\vspace{1mm} \includegraphics[width=0.157\textwidth]{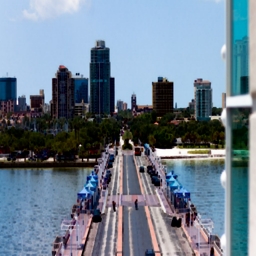}\\
\hline

\centering \adjustbox{angle=90}{ \texttt{Fleet}}  &   Transform an image into one with a sense of time or the accumulation of years.  & 
\vspace{1mm} \includegraphics[width=0.157\textwidth]{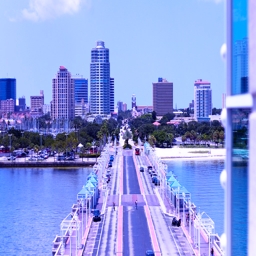} & 
\vspace{1mm} \includegraphics[width=0.157\textwidth]{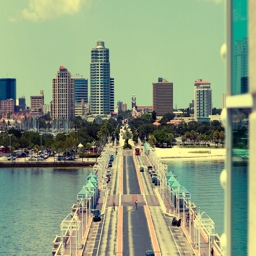} &
\vspace{1mm} \includegraphics[width=0.157\textwidth]{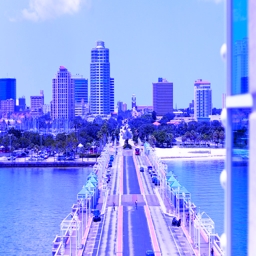} &
\vspace{1mm} \includegraphics[width=0.157\textwidth]{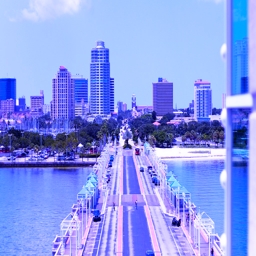}\\
\hline

\centering \adjustbox{angle=90}{ \texttt{WaterWave}}  &   Add random water wave effect onto an image.  & 
\vspace{1mm} \includegraphics[width=0.157\textwidth]{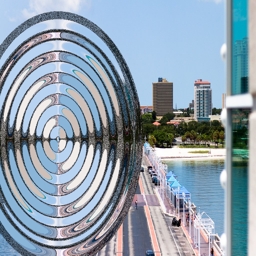} & 
\vspace{1mm} \includegraphics[width=0.157\textwidth]{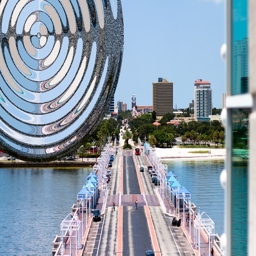} &
\vspace{1mm} \includegraphics[width=0.157\textwidth]{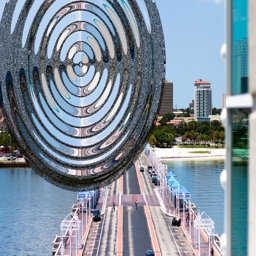} &
\vspace{1mm} \includegraphics[width=0.157\textwidth]{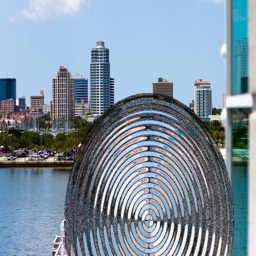}\\
\hline

\centering \adjustbox{angle=90}{ \texttt{Binary}}  &   Use different methods to binarize an image.  & 
\vspace{1mm} \includegraphics[width=0.157\textwidth]{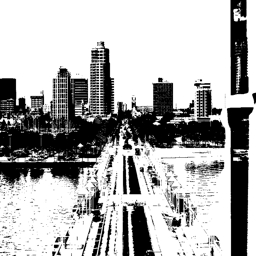} & 
\vspace{1mm} \includegraphics[width=0.157\textwidth]{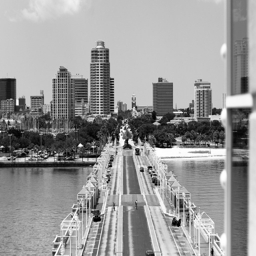} &
\vspace{1mm} \includegraphics[width=0.157\textwidth]{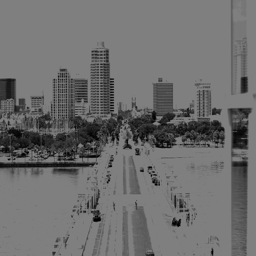} &
\vspace{1mm} \includegraphics[width=0.157\textwidth]{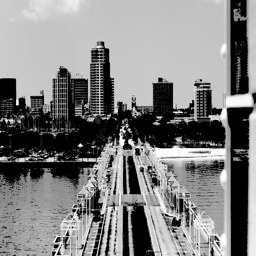}\\
\hline

\centering \adjustbox{angle=90}{ \texttt{Dehaze}}  &   Remove the haze from an image using dark channel prior \textcolor{blue}{\citep{he2010single}}.  & 
\vspace{1mm} \includegraphics[width=0.157\textwidth]{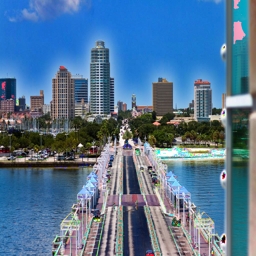} & 
\vspace{1mm} \includegraphics[width=0.157\textwidth]{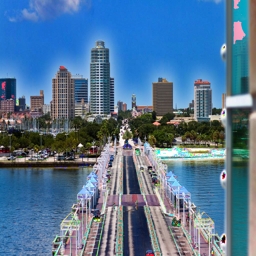} &
\vspace{1mm} \includegraphics[width=0.157\textwidth]{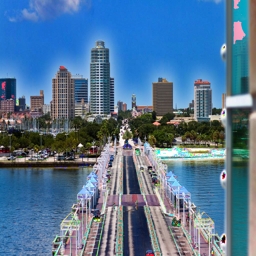} &
\vspace{1mm} \includegraphics[width=0.157\textwidth]{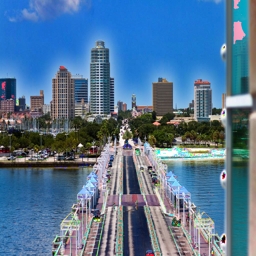}\\
\hline

\centering \adjustbox{angle=90}{ \textcolor{blue}{\texttt{Pyramid}}}  &   Stack images to create a pyramid architecture.  & 
\vspace{1mm} \includegraphics[width=0.157\textwidth]{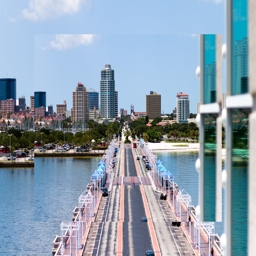} & 
\vspace{1mm} \includegraphics[width=0.157\textwidth]{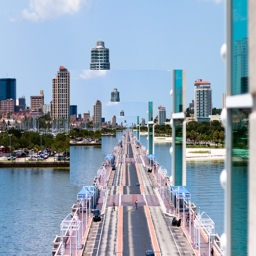} &
\vspace{1mm} \includegraphics[width=0.157\textwidth]{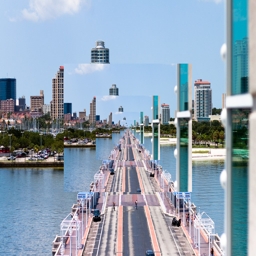} &
\vspace{1mm} \includegraphics[width=0.157\textwidth]{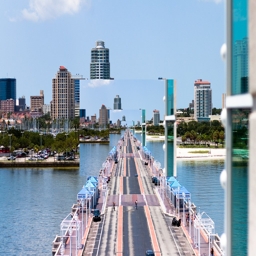}\\
\hline

\centering \adjustbox{angle=90}{ \textcolor{blue}{\texttt{Swirl}}}  &   Create a random swirl effect on an image.  & 
\vspace{1mm} \includegraphics[width=0.157\textwidth]{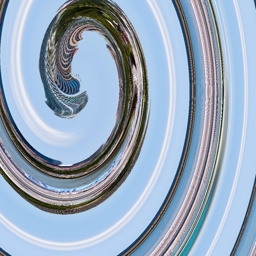} & 
\vspace{1mm} \includegraphics[width=0.157\textwidth]{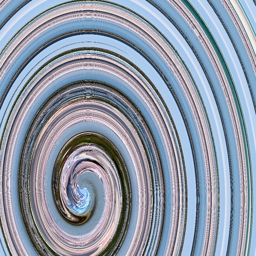} &
\vspace{1mm} \includegraphics[width=0.157\textwidth]{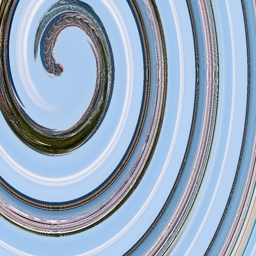} &
\vspace{1mm} \includegraphics[width=0.157\textwidth]{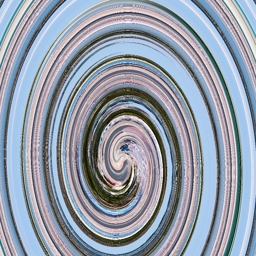}\\
\hline

\end{tabularx}}
\end{table*}

\clearpage

\section{Data Sheet for \dsnameM~}
\label{sec:datasheet}

\dscolor{\textbf{For what purpose was the dataset created?} Was there a specific task in mind? Was there a specific gap that needed to be filled? Please provide a description.}
\\
This paper explores in-context learning for image copy detection (ICD), i.e., prompting an ICD model to identify replicated images with new tampering patterns without the need for additional training.
Unlike the standard updating approach, our in-context ICD eliminates the need for fine-tuning, making it more efficient. 
To accommodate the ``seen → unseen” generalization scenario, we construct the first large-scale pattern dataset named AnyPattern, which has the largest number of tamper patterns (90 for training and 10 for testing) among all the existing ones.
\\ \\
\dscolor{\textbf{Who created the dataset (e.g., which team, research group) and on
behalf of which entity (e.g., company, institution, organization)?}} \\
The dataset was created by Wenhao Wang (University of Technology Sydney), Yifan Sun (Baidu Inc.), Zhentao Tan (Baidu Inc.), and Yi Yang (Zhejiang University).
\\ \\
\dscolor{\textbf{Who funded the creation of the dataset?} If there is an associated grant, please provide the name of the grantor and the grant name and number.} \\
Funded in part by Faculty of Engineering and Information Technology Scholarship, University of Technology Sydney.
\\ \\
\dscolor{\textbf{Any other comments?}} \\
None.\\

\noindent\fbox{\begin{minipage}{\linewidth}
  \vspace{3pt}
  \centering
  \large{\textbf{\dscolor{Composition}}}
  \vspace{3pt}
\end{minipage}}
\\

\noindent
\dscolor{\textbf{What do the instances that comprise the dataset represent (e.g., documents, photos, people, countries)?} Are there multiple types of instances (e.g., movies, users, and ratings; people and interactions between them; nodes and edges)? Please provide a description.} \\
Each instance represents an edit copy of an original image or the original image itself.
\\ \\
\dscolor{\textbf{How many instances are there in total (of each type, if appropriate)?}} \\
There are 10,000,000 training images, 1,000,000 reference images, 1,200 image-replica pairs, and 25,000 queries.
\\ \\
\dscolor{\textbf{Does the dataset contain all possible instances or is it a sample (not necessarily random) of instances from a larger set?} If the dataset is
a sample, then what is the larger set? Is the sample representative of the
larger set (e.g., geographic coverage)? If so, please describe how this
representativeness was validated/verified. If it is not representative of the larger set, please describe why not (e.g., to cover a more diverse range of instances, because instances were withheld or unavailable).} \\
The dataset contains all possible instances. 
\\ \\
\dscolor{\textbf{What data does each instance consist of?} “Raw” data (e.g., unprocessed text or images)or features? In either case, please provide a description.} \\
Each instance consists of an image.
\\ \\
\dscolor{\textbf{Is there a label or target associated with each instance?} If so, please provide a description.} \\
Yes, each edited copy has a pointer to its original image.
\\ \\
\dscolor{\textbf{Is any information missing from individual instances?} If so, please provide a description, explaining why this information is missing (e.g., because it was unavailable). This does not include intentionally removed information, but might include, e.g., redacted text.} \\
Everything is included. No data is missing.
\\ \\
\dscolor{\textbf{Are relationships between individual instances made explicit (e.g.,
users’ movie ratings, social network links)?} If so, please describe
how these relationships are made explicit.} \\
Not applicable.
\\ \\
\dscolor{\textbf{Are there recommended data splits (e.g., training, development/validation, testing)?} If so, please provide a description of these splits, explaining the rationale behind them.} \\
Yes. We follow the general retrieval task to split training, reference, and queries.
\\ \\
\dscolor{\textbf{Are there any errors, sources of noise, or redundancies in the
dataset?} If so, please provide a description.} \\
No. The patterns are all generated by code, and thus there is no error.
\\ \\
\dscolor{\textbf{Is the dataset self-contained, or does it link to or otherwise rely on external resources (e.g., websites, tweets, other datasets)?}} \\
The dataset is entirely self-contained.
\\ \\
\dscolor{\textbf{Does the dataset contain data that might be considered confidential
(e.g., data that is protected by legal privilege or by doctor–patient
confidentiality, data that includes the content of individuals’ nonpublic communications)?} If so, please provide a description.
Unknown to the authors of the datasheet.} \\
No.
\\ \\
\dscolor{\textbf{Does the dataset contain data that, if viewed directly, might be offensive, insulting, threatening, or might otherwise cause anxiety?} If so, please describe why.} \\
No.
\\ \\
\dscolor{\textbf{Does the dataset identify any subpopulations (e.g., by age, gender)?} If so, please describe how these subpopulations are identified and provide a description of their respective distributions within the dataset.} \\
No.
\\ \\
\dscolor{\textbf{Is it possible to identify individuals (i.e., one or more natural persons), either directly or indirectly (i.e., in combination with other
data) from the dataset?} If so, please describe how.} \\
No.
\\
\dscolor{\textbf{Any other comments?}} \\
None. \\

\noindent\fbox{\begin{minipage}{\linewidth}
  \vspace{3pt}
  \centering
  \large{\textbf{\dscolor{Collection}}}
  \vspace{3pt}
\end{minipage}}
\\

\noindent
\dscolor{\textbf{How was the data associated with each instance acquired?} Was
the data directly observable (e.g., raw text, movie ratings), reported by
subjects (e.g., survey responses), or indirectly inferred/derived from other
data (e.g., part-of-speech tags, model-based guesses for age or language)?
If the data was reported by subjects or indirectly inferred/derived from
other data, was the data validated/verified? If so, please describe how.} \\
The data was auto-generated by the code of each pattern. We release the code at: \url{https://github.com/WangWenhao0716/AnyPattern}.
\\ \\
\dscolor{\textbf{What mechanisms or procedures were used to collect the data
(e.g., hardware apparatuses or sensors, manual human curation,
software programs, software APIs)?} How were these mechanisms or
procedures validated?} \\
Not applicable.
\\ \\
\dscolor{\textbf{If the dataset is a sample from a larger set, what was the sampling
strategy (e.g., deterministic, probabilistic with specific sampling
probabilities)?}} \\
AnyPattern does not sample from a larger set.
\\ \\
\dscolor{\textbf{Who was involved in the data collection process (e.g., students,
crowdworkers, contractors) and how were they compensated (e.g.,
how much were crowdworkers paid)?}} \\
No crowdworkers are needed in the data collection process. 
\\ \\
\dscolor{\textbf{Over what timeframe was the data collected? Does this timeframe
match the creation timeframe of the data associated with the instances
(e.g., recent crawl of old news articles)?} If not, please describe the timeframe in which the data associated with the instances was created.} \\
Not applicable.
\\ \\
\dscolor{\textbf{Were any ethical review processes conducted (e.g., by an institutional review board)?} If so, please provide a description of these review processes, including the outcomes, as well as a link or other access point to any supporting documentation.} \\
There were no ethical review processes conducted.
\\ \\
\dscolor{\textbf{Did you collect the data from the individuals in question directly,
or obtain it via third parties or other sources (e.g., websites)?}} \\
The data was auto-generated by the code of each pattern. We release the code at: \url{https://github.com/WangWenhao0716/AnyPattern}.
\\ \\
\dscolor{\textbf{Were the individuals in question notified about the data collection?} If so, please describe (or show with screenshots or other information) how notice was provided, and provide a link or other access point to, or otherwise reproduce, the exact language of the notification itself.}\\
Not applicable.

\noindent{}\dscolor{\textbf{Did the individuals in question consent to the collection and use
of their data?} If so, please describe (or show with screenshots or other
information) how consent was requested and provided, and provide a
link or other access point to, or otherwise reproduce, the exact language
to which the individuals consented.} \\
Not applicable.

\dscolor{\textbf{If consent was obtained, were the consenting individuals provided with a mechanism to revoke their consent in the future or for certain uses?} If so, please provide a description, as well as a link or other access point to the mechanism (if appropriate).} \\
Not applicable.
\\ \\
\dscolor{\textbf{Has an analysis of the potential impact of the dataset and its use
on data subjects (e.g., a data protection impact analysis) been conducted?} If so, please provide a description of this analysis, including
the outcomes, as well as a link or other access point to any supporting
documentation.} \\
Not applicable.
\\ \\
\dscolor{\textbf{Any other comments?}} \\
None.
\\ \\
\noindent\fbox{\begin{minipage}{\linewidth}
  \vspace{3pt}
  \centering
  \large{\textbf{\dscolor{Preprocessing}}}
  \vspace{3pt}
\end{minipage}}
\\

\noindent{}\dscolor{\textbf{Was any preprocessing/cleaning/labeling of the data done (e.g.,
discretization or bucketing, tokenization, part-of-speech tagging,
SIFT feature extraction, removal of instances, processing of missing values)?} If so, please provide a description. If not, you may skip the
remaining questions in this section.} \\
Not applicable.
\\ \\
\dscolor{\textbf{Was the “raw” data saved in addition to the preprocessed/cleaned/labeled
data (e.g., to support unanticipated future uses)?} If so, please provide a link or other access point to the “raw” data.} \\
Yes, raw data is saved.
\\ \\
\dscolor{\textbf{Is the software that was used to preprocess/clean/label the data
available?} If so, please provide a link or other access point.} \\
Not applicable.
\\ \\
\dscolor{\textbf{Any other comments?}}\\
None.
\\

\noindent\fbox{\begin{minipage}{\linewidth}
  \vspace{3pt}
  \centering
  \large{\textbf{\dscolor{Uses}}}
  \vspace{3pt}
\end{minipage}}
\\

\noindent
\dscolor{\textbf{Has the dataset been used for any tasks already?} If so, please provide a description.}\\
No.
\\\\
\dscolor{\textbf{Is there a repository that links to any or all papers or systems that use the dataset?} If so, please provide a link or other access point.}\\
No.
\\\\
\dscolor{\textbf{What (other) tasks could the dataset be used for?}}\\
This dataset is specifically designed for in-context image copy detection.
\\\\
\dscolor{\textbf{Is there anything about the composition of the dataset or the way it was collected and preprocessed/cleaned/labeled that might impact future uses? } For example, is there anything that a dataset consumer might need to know to avoid uses that could result in unfair treatment of individuals or groups (e.g., stereotyping, quality of service issues) or other risks or harms (e.g., legal risks, financial harms)? If so, please provide a description. Is there anything a dataset consumer could do to mitigate these risks or harms?}\\
There is minimal risk for harm: the data were already public.
No.
\\\\
\dscolor{\textbf{Are there tasks for which the dataset should not be used?} If so, please provide a description.}\\
All tasks that utilize this dataset should follow the MIT License.
\\\\
\dscolor{\textbf{Any other comments?}}\\ \
None.
\\

\noindent\fbox{\begin{minipage}{\linewidth}
  \vspace{3pt}
  \centering
  \large{\textbf{\dscolor{Distribution}}}
  \vspace{3pt}
\end{minipage}}
\\

\noindent
\dscolor{\textbf{Will the dataset be distributed to third parties outside of the entity (e.g., company, institution, organization) on behalf of which the dataset was created?} If so, please provide a description.}\\
Yes, the dataset is publicly available on the internet.
\\ \\
\dscolor{\textbf{How will the dataset will be distributed (e.g., tarball on website, API, GitHub)?} Does the dataset have a digital object identifier (DOI)?}\\
The dataset is distributed on the project website: \url{https://anypattern.github.io/}.
The dataset shares the same DOI as this paper.
\\ \\
\dscolor{\textbf{When will the dataset be distributed?}}\\
The dataset is released in April, 2024.
\\\\
\dscolor{\textbf{Will the dataset be distributed under a copyright or other intellectual property (IP) license, and/or under applicable terms of use (ToU)?} If so, please describe this license and/or ToU, and provide a link or other access point to, or otherwise reproduce, any relevant licensing terms or ToU, as well as any fees associated with these restrictions.}\\
No.
\\\\
\dscolor{\textbf{Have any third parties imposed IP-based or other restrictions on the data associated with the instances?} If so, please describe these restrictions, and provide a link or other access point to, or otherwise reproduce, any relevant licensing terms, as well as any fees associated with these restrictions.}\\
No.
\\\\
\dscolor{\textbf{Do any export controls or other regulatory restrictions apply to the dataset or to individual instances?} If so, please describe these restrictions, and provide a link or other access point to, or otherwise reproduce, any supporting documentation.}\\
No.
\\\\
\dscolor{\textbf{Any other comments?}}\\
None.
\\

\noindent\fbox{\begin{minipage}{\linewidth}
  \vspace{3pt}
  \centering
  \large{\textbf{\dscolor{Maintenance}}}
  \vspace{3pt}
\end{minipage}}
\\\\
\dscolor{\textbf{Who will be supporting/hosting/maintaining the dataset?}}\\
The authors of this paper will be supporting and maintaining the dataset.
\\\\
\dscolor{\textbf{How can the owner/curator/manager of the dataset be contacted (e.g., email address)?}}\\
The contact information of the curators of the dataset is listed on the project website: \url{https://anypattern.github.io/}.
\\\\
\dscolor{\textbf{Is there an erratum?} If so, please provide a link or other access point.}\\
There is no erratum for our initial release.
Errata will be documented in future releases on the dataset website.
\\\\
\dscolor{\textbf{Will the dataset be updated (e.g., to correct labeling errors, add new instances, delete instances)?} If so, please describe how often, by whom, and how updates will be communicated to dataset consumers (e.g., mailing list, GitHub)?}\\
Yes, we will monitor cases when users can report harmful images and creators can remove their videos.
We may include more patterns in the future.
\\\\
\dscolor{\textbf{If the dataset relates to people, are there applicable limits on the retention of the data associated with the instances (e.g., were the individuals in question told that their data would be retained for a fixed period of time and then deleted)?} If so, please describe these limits and explain how they will be enforced.}\\
No, this dataset is not related to people.
\\\\
\dscolor{\textbf{Will older versions of the dataset continue to be supported/hosted/maintained? }If so, please describe how. If not, please describe how its obsolescence will be communicated to dataset consumers.}\\
We will continue to support older versions of the dataset.
\\\\
\dscolor{\textbf{If others want to extend/augment/build on/contribute to the dataset, is there a mechanism for them to do so?} If so, please provide a description. Will these contributions be validated/verified? If so, please describe how. If not, why not? Is there a process for communicating/distributing these contributions to dataset consumers? If so, please provide a description.}\\
Anyone can extend/augment/build on/contribute to \dsnameM~.
Potential collaborators can contact the dataset authors.
\\\\
\dscolor{\textbf{Any other comments?}}\\
None.

\end{appendices}
\bibliographystyle{spbasic} 
\bibliography{main}

\end{document}